%% file: 0_main.tex
\documentclass[runningheads]{llncs}

\usepackage[mobile]{eccv}

\usepackage{eccvabbrv}

\usepackage{graphicx}
\usepackage{booktabs}

\usepackage[accsupp]{axessibility}  

\usepackage{hyperref}

\usepackage{orcidlink}

\input{preamble}
\input{commands}

\begin{document}

\title{Discover-then-Name: Task-Agnostic Concept Bottlenecks via Automated Concept Discovery} 

\titlerunning{Task-Agnostic Concept Bottlenecks via Automated Concept Discovery}

\author{
Sukrut Rao$^{*,}$\inst{1,2}\orcidlink{0000-0001-8896-7619}
\and
Sweta Mahajan$^{*,}$\inst{1,2}\orcidlink{0009-0004-8811-8771}
\and
Moritz Böhle\inst{1}\orcidlink{0000-0002-5479-3769}
\and
Bernt Schiele\inst{1}\orcidlink{0000-0001-9683-5237}
}

\authorrunning{S.~Rao et al.}

\institute{
Max Planck Institute for Informatics, Saarland Informatics Campus, Saarbrücken
\and
RTG Neuroexplicit Models, Saarbrücken
\email{\{sukrut.rao,sweta.mahajan,mboehle,schiele\}@mpi-inf.mpg.de}
}

\maketitle

\begin{abstract}
Concept Bottleneck Models (CBMs) have recently been proposed to address the `black-box' problem of deep neural networks, by first mapping images to a human-understandable concept space and then linearly combining concepts for classification. Such models typically require first coming up with a set of concepts relevant to the task and then aligning the representations of a feature extractor to map to these concepts. However, even with powerful foundational feature extractors like CLIP, there are no guarantees that the specified concepts are detectable.
In this work, we leverage recent advances in mechanistic interpretability and propose
a novel CBM approach --- called Discover-then-Name-CBM (\ours) --- that inverts the typical paradigm: instead of pre-selecting concepts based on the downstream classification task, we use sparse autoencoders to first \emph{discover} concepts learnt by the model, and then \emph{name} them and train linear probes for classification. Our concept extraction strategy is \emph{efficient}, since it is agnostic to the downstream task, and uses concepts \emph{already known} to the model. We perform a comprehensive evaluation across multiple datasets and CLIP architectures and show that our method yields semantically meaningful concepts, assigns appropriate names to them that make them easy to interpret, and yields performant and interpretable CBMs. Code available at \href{https://github.com/neuroexplicit-saar/Discover-then-Name}{https://github.com/neuroexplicit-saar/discover-then-name}.
\end{abstract}

\def\thefootnote{*}\footnotetext{Equal contribution.}
\def\thefootnote{\arabic{footnote}}

\input{1_introduction}

\input{2_related}
\input{3_method}
\input{4_experiments}

\input{5_results}
\input{6_conclusion}

\section*{Acknowledgements}
Funded in part by the Deutsche Forschungsgemeinschaft (DFG, German Research Foundation) - GRK 2853/1 ``Neuroexplicit Models of Language, Vision, and Action'' - project number 471607914.

%
%
\bibliographystyle{splncs04}
\bibliography{references}

\clearpage

\input{supplement/0_main}

\bibliographystyleS{splncs04}
\bibliographyS{supp_references}

\end{document}

%% file: preamble.tex
\usepackage{hhline}
\usepackage{changepage}
\usepackage{enumitem}
\usepackage[resetlabels,labeled]{multibib}
\newcites{S}{Supplementary References}
\usepackage{bm}
\usepackage{color}
\definecolor{forestgreen}{RGB}{0, 150, 0}

%% file: 1_introduction.tex
\section{Introduction}
\label{sec:introduction}

\input{figures_tex/teaser}

Deep neural networks have been immensely successful for a variety of tasks, yet their `black-box' nature poses a risk for their use in safety-critical applications. While attribution methods \cite{selvaraju2017grad,ribeiro2016should,bach2015pixel} have popularly been used to explain such models \emph{post-hoc}, they have been shown to often provide explanations unfaithful to the model \cite{adebayo2018sanity,adebayo2021post,rao2022towards}. To address this, \emph{inherently interpretable} models have been proposed \cite{chen2019looks,koh2020concept,boehle2022bcos} that constrain the model
to yield more faithful and human-understandable explanations in the form of heatmaps, concepts, or prototypes.

Concept Bottleneck Models (CBMs) \cite{koh2020concept,yuksekgonul2022post,oikarinen2023label} are a class of inherently interpretable models that express their prediction as a linear combination of simpler but human-interpretable concepts detected from the input features. While typically constrained by the need of a labelled attribute dataset for training \cite{koh2020concept}, recent CBMs leverage large-language models (LLMs) such as GPT-3\cite{brown2020language} to generate class-specific concepts and vision-language models (VLMs) such as \clip\cite{radford2021learning} to learn the mapping from inputs to concepts in an attribute-label-free manner \cite{oikarinen2023label,yang2023language,menon2022visual,panousis2023sparse}, and have been shown to be performant even on large datasets such as \imagenet\cite{deng2009imagenet}. However, such methods still require querying LLMs based on the classification task, and it is unclear if the concepts one \emph{wants} the model to detect \emph{can} be detected at all; in fact, recent works have suggested that while plausible, explanations from such CBMs may not be faithful \cite{margeloiu2021concept,roth2023waffling}.

To address this, in this work, we invert the typical CBM paradigm, and aim to \emph{discover} concepts the model \emph{knows}, name them, and then perform classification (\cref{fig:teaser}). We specifically use \clip feature extractors to leverage vision-language alignment for automated naming of concepts. While raw features of a network are typically uninterpretable \cite{elhage2022toy}, sparse autoencoders (SAEs) have been shown to be a promising tool in the context of language models wherein they disentangle learned representations into a sparse set of human-understandable concepts. This is achieved by decomposing the representations into a sparse linear combination of a set of learned dictionary vectors \cite{bricken2023monosemanticity,cunningham2023sparse}.
We extend this to vision and find it to be similarly promising, and surprisingly, find that the dictionary vectors appear to align well with text embeddings of concepts they represent in \clip space, thus making their corresponding concepts nameable (\cref{fig:taskgnostic}). Finally, we use this latent concept space as a concept bottleneck, and show that, once learnt, it can be frozen and used `as is' to train classifiers to construct performant CBMs for a variety of downstream classification tasks. Our approach is also computationally efficient since it learns concept bottlenecks in a \emph{task-agnostic} manner, eliminating the need to make queries to external LLMs to find task-relevant concepts.

In summary, \textbf{our contributions are} $\bullet$ We propose \ours, a novel CBM that leverages sparse autoencoders (SAEs) to \emph{discover} concepts learnt by CLIP. We find that SAEs lend themselves well to our simple and intuitive approach for automated concept discovery.
$\bullet$  We propose a novel approach to automatically \emph{name} the discovered concepts, by mapping concepts to text with embeddings most similar to the corresponding dictionary vectors of the learned concept. We find that this often yields names semantically consistent to the images activating the concept (\cref{fig:taskgnostic}).
$\bullet$  We show that, once discovered and named, 
the learnt concept mapping can be used to train concept bottleneck models (CBMs) out-of-the-box for a variety of downstream classification tasks. Specifically, we discover concepts using \cctm\cite{sharma2018conceptual} in a task-agnostic fashion, and then construct CBMs for a variety of downstream datasets: \imagenet\cite{deng2009imagenet}, \places\cite{zhou2017places}, \cifart\cite{krizhevsky2009learning}, and \cifarh\cite{krizhevsky2009learning}. Importantly, this  task-agnostic concept discovery approach yields both performant (\cref{tab:accuracy}) and interpretable (\cref{fig:localexp,fig:cbmcompare}) classifiers.

%% file: figures_tex/teaser.tex
\begin{figure}
    \centering
    \includegraphics[trim=0 6cm 0 2cm, clip, width=.85\linewidth]{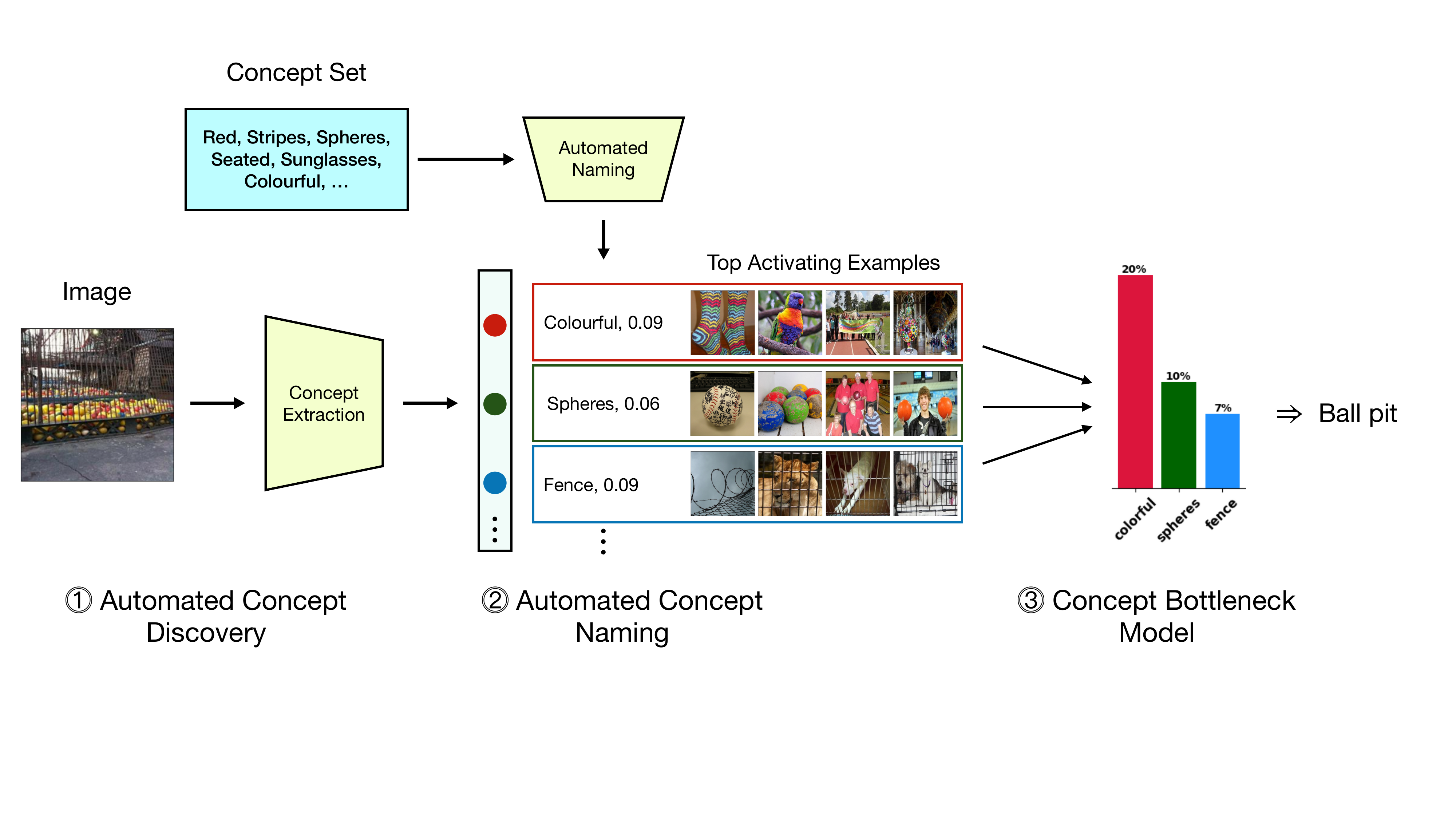}
    \caption{\textbf{Automated concept extraction and naming to construct task-agnostic concept bottlenecks.} Our approach consists of three steps: \textbf{(1)} we use a sparse autoencoder to extract disentangled concepts from \clip feature extractors, \textbf{(2)} automatically name extracted concepts by matching the dictionary vectors with the closest text embedding in \clip space from a concept set of texts, and \textbf{(3)} use this named concept extractor layer as a concept bottleneck to create concept bottleneck models for classification on different datasets. In the example shown, the concepts `colorful', `spheres', and `fence' are extracted from the image with high strengths, resulting in a prediction of `ball pit'. For details, see \cref{fig:method} and \cref{sec:method}.}
    \label{fig:teaser}
\end{figure}

%% file: 2_related.tex
\section{Related Work}
\label{sec:related}

\myparagraph{Concept-based Explanations} (\eg\cite{kim2018interpretability,achtibat2023attribution,koh2020concept,oikarinen2022clip}) aim to express a model's decision via human-understandable concepts. Unlike popularly used post-hoc attribution heatmaps (\eg\cite{selvaraju2017grad,sundararajan2017axiomatic,shrikumar2017learning,lundberg2017unified,ribeiro2016should,bach2015pixel}) that only inform \emph{which regions} in the input is influential for the decision, such methods attempt to also answer \emph{what} high-level concepts are important for the model\cite{achtibat2023attribution}. In our work, we propose a pipeline for automatically extracting and naming such concepts from \clip and using them to build interpretable models.

\myparagraph{Concept Discovery} \cite{fel2023craft,fel2023holistic,omahony2023disentangling,ghorbani2019towards,zhang2021invertible,graziani2023concept,bau2017network,oikarinen2022clip,panousis2023discover,bricken2023monosemanticity,cunningham2023sparse} methods have been proposed to better understand models by discovering and extracting semantically meaningful concepts learnt by them. They typically focus on explaining the function of neurons in a model \cite{oikarinen2022clip,panousis2023discover,bau2017network,fel2023craft}, or on discovering features present in an input, and have been shown to be useful for diagnosing model failures \cite{fel2023craft}. However, these methods assign concepts to individual neurons, which may often be polysemantic and not decodable to human-understandable concepts \cite{elhage2022toy}. 
Recently, \cite{bricken2023monosemanticity} showed that sparse autoencoders (SAEs) can be effective to address the polysemanticity and superposition problem in deep networks \cite{elhage2022toy} and extract mono-semantic concepts from language models (\cf\cite{cunningham2023sparse}). We extend the setup of \cite{bricken2023monosemanticity} to vision and use sparse autoencoders to automatically extract concepts learnt by \clip.

\myparagraph{Explanations using Language} \cite{dani2023devil,oikarinen2022clip,panousis2023discover,hernandez2021natural,zhao2023automated,chen2023interpreting,hendricks2016generating,tewel2022zerocap,moayeri2023text} have become popular to express a model's learnt representations \cite{dani2023devil,oikarinen2022clip,panousis2023discover} in an easily human-interpretable manner. To decode concepts to language, such methods typically use a large-language model (LLM) such as GPT-3\cite{brown2020language} and learn a mapping from vision features to the LLM input. More recently, \cite{panousis2023discover,oikarinen2022clip} leverage \clip, by aligning vision features of the model being explained to the representation space of \clip and finding the closest aligned texts from a large concept set. Similar to \emph{Concept Discovery} methods, this line of work assigns concepts to individual neurons as well. In contrast, we use sparse autoencoders (SAEs) \cite{bricken2023monosemanticity} to first disentangle the representation space into more human-interpretable concepts, and then name each concept. In particular, we encode the CLIP features to a high dimensional latent space, which we then pass through the SAE decoder to reconstruct back the \clip features. Surprisingly, when using \clip feature extractors, we find that the dictionary vectors of the SAE decoder can directly be decoded into text by finding most similar text embeddings in \clip space, without needing to use LLMs as used by \cite{dani2023devil,tewel2022zerocap,moayeri2023text}, and are effective in yielding semantically meaningful and human-interpretable concepts (\cref{fig:taskgnostic,fig:conceptaccuracy_userstudy}).

\myparagraph{Concept Bottleneck Models (CBMs)} \cite{koh2020concept,oikarinen2023label,panousis2023sparse,xu2024energy,yuksekgonul2022post,kim2023probabilistic,chauhan2023interactive,panousis2023hierarchical,alukaev2023cross,kazmierczak2023clip,jeyakumar2022automatic,zhou2018interpretable,sawada2022concept,yang2023language,menon2022visual} are a recently popular class of inherently interpretable models (\eg\cite{boehle2022bcos,chen2019looks}) that use a concept bottleneck layer (CBL) to extract named concepts and then learn a (typically sparse) linear classifier that predicts by combining such concepts, yielding highly interpretable explanations. While such methods typically require a labelled concept dataset \cite{koh2020concept} to learn the concept bottleneck, recent works leverage LLMs such as GPT-3\cite{brown2020language} and VLMs such as \clip\cite{radford2021learning} to learn such bottlenecks without needing the concept labels\cite{oikarinen2023label,panousis2023sparse,panousis2023hierarchical,yang2023language}, making them scale to large datasets such as \imagenet\cite{deng2009imagenet} in a performant manner. Given a classification task, such methods first query an LLM for concepts relevant to the task, and use a VLM to learn a concept bottleneck where each neuron aligns to one of the desired concepts. However, it is unclear if the feature extractor can truly recognize all such concepts when specified a priori, since they may often be non-visual \cite{roth2023waffling,yang2023language} and their faithfulness has also been called into question \cite{roth2023waffling,margeloiu2021concept}. Further, the CBL needs to be trained separately for each classification dataset. 
In contrast, we flip the paradigm and first extract concepts that are detected by the model to train the concept bottleneck, using a dataset \emph{independent} of the downstream classification task. We then \emph{fix} the concept bottleneck and train linear classifiers for several datasets and show that this yields highly performant and interpretable models. Similar to us, \cite{jeyakumar2022automatic} also first discover concepts before constructing a CBM; however, in contrast our method does not require any external text annotations for the images and the concept discovery can even be done using a dataset different from that of the downstream task.

\myparagraph{Explaining CLIP.} Several approaches have been proposed to specifically explain and understand \clip \cite{radford2021learning} models \cite{tewel2022zerocap,panousis2023sparse,bhalla2024interpreting}. Similar to \cite{bhalla2024interpreting}, we disentangle CLIP features into human interpretable concepts. However, in contrast to \cite{bhalla2024interpreting}, we do not optimize for a sparse concept representation per image using a predefined concept set, and instead first apply a general concept discovery framework \cite{bricken2023monosemanticity} for extracting human understandable concepts and then name them post hoc in a task-agnostic manner to construct CBMs. Concurrent to our work, \cite{fry2024towards} also uses sparse autoencoders to extract and disentangle concepts learnt by CLIP vision encoders. However, in contrast to \cite{fry2024towards} that then analyzes properties of the trained SAEs similar to \cite{bricken2023monosemanticity}, our work uses their dictionary vectors to automatically name the extracted concepts and then construct performant task-agnostic CBMs on downstream datasets.

%% file: 3_method.tex
\section{Constructing CBMs via Automated Concept Discovery}
\label{sec:method}
\input{figures_tex/method}
In this section, we describe our approach which consists of three stages:
discovering the concepts the CLIP model has learnt via a sparse autoencoder (\cref{sec:method:extract}),
naming those concepts in natural language by leveraging the \clip text embeddings from a large vocabulary, (\cref{sec:method:name}), and, lastly, training an interpretable concept bottleneck model (CBM) based on the discovered concepts (\cref{sec:method:cbm}).

\subsection{Extracting Concepts Learned by the Model}
\label{sec:method:extract}

To discover the concepts learned by the model, we adapt the sparse autoencoder (SAE) approach as described by \cite{bricken2023monosemanticity}.
Specifically, we aim to discover concepts by representing the \clip features in a high-dimensional, but very sparsely activating space.
For language models, this has been shown to yield representations in which individual neurons (dimensions) are more easily interpretable \cite{bricken2023monosemanticity}.

\myparagraph{The Sparse Autoencoders (SAEs)} proposed by \cite{bricken2023monosemanticity} consist of a linear encoder $f(\cdot)$ with weights $\vec W_{E}\myin\mathbb R^{d\times h}$, a ReLU non-linearity $\phi$, and a linear decoder $g(\cdot)$ with weights $\vec W_D\myin\mathbb R^{h\times d}$. For a given input $\vec a$, the SAE computes:
\begin{align}
\label{eq:SAE}
    \text{SAE}(\vec a) = (g \circ \phi \circ f) (\vec a) = \vec W_D^T\;\phi\left(\vec W_E^T \vec a\right)\;.
\end{align}
Importantly, the hidden representation $f(\vec a)$ is of significantly higher dimensionality than the CLIP embedding space (\ie $h\gg d$), but optimised to activate only very sparsely. Specifically, the SAE is trained with an $L_2$ reconstruction loss, as well as an $L_1$ sparsity regularisation:
\begin{align}
\label{eq:sae_loss}
\mathcal L_\text{SAE}(\vec a) = \lVert\text{SAE}(\vec a) - \vec a\rVert^2_2 + \lambda_1 \lVert \phi(f(\vec a))\rVert_1
\end{align}
 with $\lambda_1$ a hyperparameter.
To discover a diverse set of concepts for usage in downstream tasks, we train the SAE on a large dataset $\Dext$; given the reconstruction objective, no labels for this dataset are required.

Note that sparsity does of course not \textit{guarantee} that individual neurons in the hidden representation of the SAE align with human-interpretable concepts. However, similar to \cite{bricken2023monosemanticity}, in our experiments we find that this is often the case, and, as we discuss in the next section, can often even be automatically named.

\myparagraph{Why SAEs?}
While SAEs are certainly not the only option for concept discovery in DNNs, recent work on language models suggests that they might be particularly well suited to discover interpretable concepts, see \cite{bricken2023monosemanticity,cunningham2023sparse}, and exhibit certain properties that lend themselves well for automatically naming visual concepts.
Specifically, as we will see in the next section, by reconstructing the original feature space, we are able to leverage the dictionary vectors of the reconstruction matrix $\vec W_D$ for assigning names to individual concepts.
Moreover, in contrast to dimensionality reduction techniques (e.g. PCA), SAEs are able to represent more features than there are neurons, which was shown to be advantageous to address the problem of polysemanticity \cite{bricken2023monosemanticity}.

\subsection{Automated Concept Naming}
\label{sec:method:name}

Once we trained the SAE,
we aim to automatically name the individual feature dimensions in the hidden representation of the SAE.
For this, we propose using a large vocabulary of English words, say $\vocab\myeq\{v_1,v_2,\dots\}$, which we embed via the \clip text encoder $\mathcal T$ to obtain word embeddings $\mathcal E\myeq\{\vec e_1, \vec e_2, \dots\}$.

To name the SAE's hidden features, we propose to leverage the fact that each of the SAE neurons $c$ is assigned a specific dictionary vector $\vec p_c$, corresponding to a column of the decoder weight matrix:
\begin{align}
\label{eq:dicvec}
    \vec p_c = [\vec W_D]_c \in \mathbb R^{d}\;. 
\end{align}
If the SAE indeed succeeds to decompose image representations given by \clip into individual concepts, we expect the $\vec p_c$ to resemble the embeddings of particular words that \clip has learnt to expect in a corresponding image caption. 

Hence, to name the `concept' neuron $c$ of the SAE, we propose to assign it the word $s_c$ of the closest text embedding in $\mathcal E$:
\begin{equation}
\label{eq:name}
    s_c = \arg\min_{v \in \vocab} \;\left[\cos\left(\vec p_c,\clipdec(v)\right) \right]\;.  
\end{equation}

Note that this setting
is equivalent to using the SAE to reconstruct a CLIP feature when only the concept to be named is present.
As CLIP was trained to optimise cosine similarities between text and image embeddings, using the cosine similarity to assign names to concept nodes is a natural choice in this context.

\subsection{Constructing Concept Bottleneck Models}
\label{sec:method:cbm}
Thus far, we trained an SAE to obtain sparse representations (\cref{sec:method:extract}), and named individual `neurons' by leveraging the similarity between dictionary vectors $\vec p_c$ to word embeddings obtained via \clip's text encoder $\mathcal T$ (\cref{sec:method:name}).

Such a sparse decomposition into named `concepts' constitutes the ideal starting point for constructing interpretable {Concept Bottleneck Models} (CBMs) \cite{koh2020concept,yuksekgonul2022post,oikarinen2023label}: 
for a given \textit{labelled} dataset $\mathcal D_\text{probe}$, we can now train a linear transformation $h(\cdot)$ on the SAE's \textit{sparse concept activations},
yielding
our CBM
$t(\cdot)$:
\begin{align}
    t(\vec x_i) = (\underbrace{h}_\text{Probe} \circ \;\underbrace{\phi \circ f}_\text{SAE}  \; \circ \underbrace{\mathcal I}_\text{CLIP}) (\vec x_i)\;.
\end{align}
Here, $\vec x_i$ denotes an image from the probe dataset.
The probe is trained using the cross-entropy loss, and to increase the interpretability of the resulting CBM classifier, we additionally apply a sparsity loss to the probe weights: 
\begin{align}
\mathcal L_\text{probe}(\vec x_i) = \text{CE}\left(t(\vec x_i), y_i\right)+ \lambda_2 \lVert\omega\rVert_1
\end{align}
where, $\lambda_2$ is a hyperparameter, $y_i$ the ground truth label of $\vec x_i$ in the probe dataset, and $\omega$ denotes the parameters of the linear probe.

Importantly, note that the feature extractor, the dataset used for concept discovery, and the vocabulary used for naming can be freely chosen. As such, our approach is likely to benefit from advances in any of these directions.

%% file: figures_tex/method.tex
\begin{figure}[t]
    \centering
    \includegraphics[width=.85\linewidth]{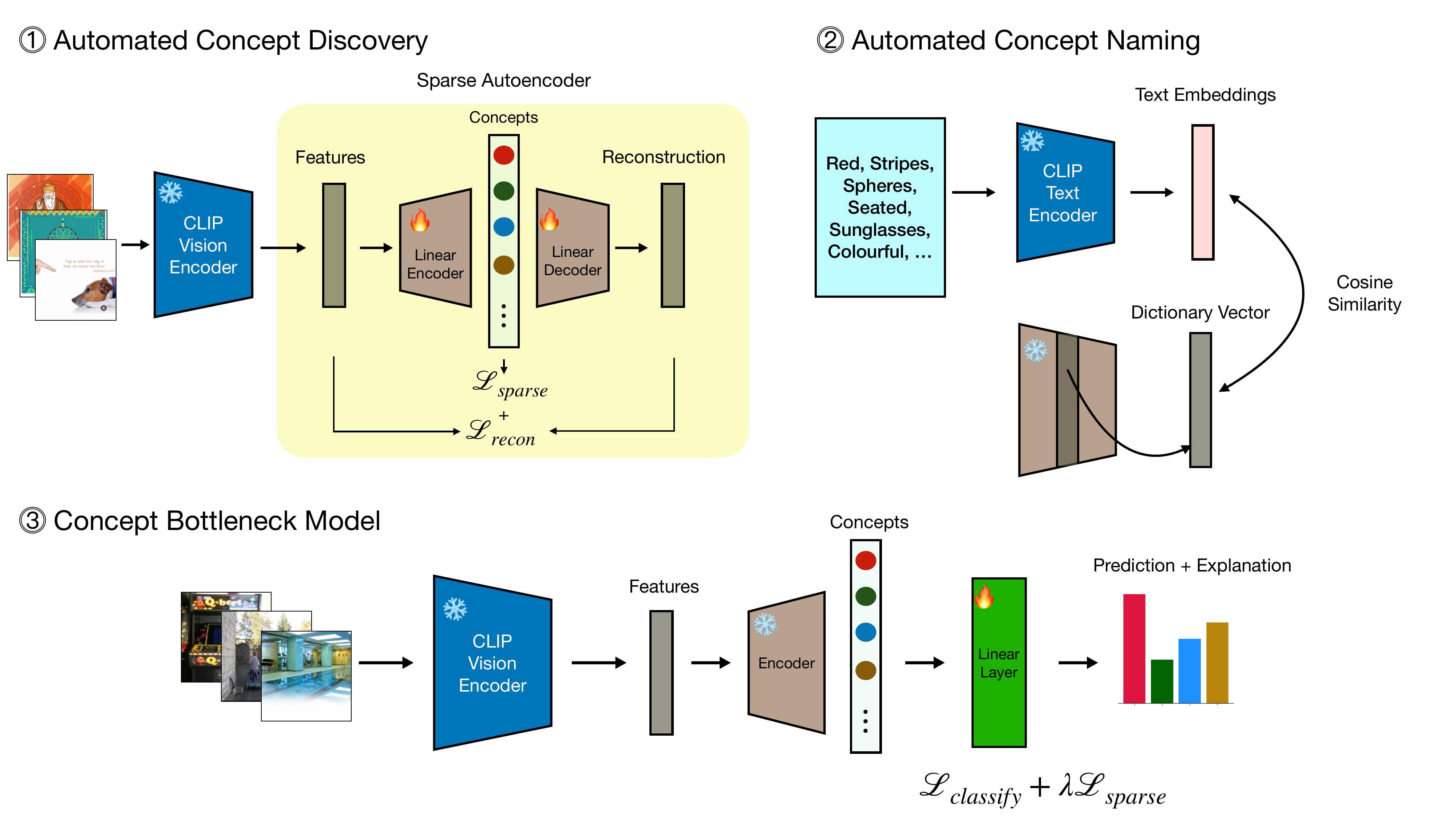}
    \caption{\textbf{Overview.} Our approach consists of three steps. \textbf{(1)} We train a sparse autoencoder to extract disentangled concepts from a \clip vision backbone. The autencoder is trained on a large dataset $\Dext$ to reconstruct 
    \clip features using a linear combination of encoded concepts, which are optimized to be sparse using $L_1$ sparsity. The weights of the decoder can be interpreted as dictionary vectors whose linear sum with concept strengths reconstructs the original feature (\cref{sec:method:extract}). \textbf{(2)} We use a large concept set of texts $\vocab$ to name each extracted concept, by finding the text from the set whose embedding has the highest cosine similarity to concept's dictionary vector (\cref{sec:method:name}). \textbf{(3)} We use the extracted and named concepts as a concept bottleneck layer, and train linear classifiers to construct inherently interpretable concept bottleneck models across downstream datasets $\Dclass$ using the same bottleneck layer (\cref{sec:method:cbm}).
    }
    \label{fig:method}
\end{figure}

%% file: 4_experiments.tex
\section{Evaluation of Concept Discovery and Naming}
\label{sec:discoverynaming}

In this section, we evaluate the effectiveness of using SAEs to discover and name concepts in CLIP vision encoders; see \cref{sec:results} for an evaluation of CBMs built on the SAEs. In \cref{sec:discoverynaming:taskagnosticity}, we first evaluate the accuracy and task agnosticity of the discovered concepts qualitatively and quantitatively, in \cref{sec:discoverynaming:supernodes}, we discuss the impact of the vocabulary $\vocab$ towards the granularity of concept names, and in \cref{sec:discoverynaming:metaclustering}, we evaluate how well semantically similar concepts group together.

\myparagraph{Setup.} We use a \clip\cite{radford2021learning} \resnetf\cite{he2016deep} vision encoder for extracting features, and use the corresponding text encoder for labelling the extracted concepts. For additional results using \clip\vitb and \vitl \cite{dosovitskiy2021an}, see \cref{supp:sec:qualitative,supp:sec:quantitative}. 
To extract concepts, we follow a setup similar to \cite{bricken2023monosemanticity} 
and train SAEs using the CC3M dataset~\cite{sharma2018conceptual}. Following \cite{oikarinen2022clip}, we use the set of 20k most frequent English words as the vocabulary $\vocab$ (\cref{eq:name}). For details, see \cref{supp:sec:implementation:training}.

\subsection{Task-Agnosticity and Accuracy of Concepts}
\label{sec:discoverynaming:taskagnosticity}

In this section, we qualitatively and quantitatively evaluate the extracted and named concepts for semantic consistency and accuracy.

\myparagraph{Qualitative.}
To showcase the promise of our proposed approach, in \cref{fig:taskgnostic} we visualize the top activating images across four datasets for various concepts that were discovered and named as described in \cref{sec:method:extract,sec:method:name}. For this, we select concepts $c$ from the vocabulary with a high cosine similarity between $\vec p_c$ and $\clipdec(s_c)$, see \cref{eq:dicvec,eq:name}.
In particular, we show examples for various low-level concepts (turquoise, pink, striped), object and scene-specific concepts (sunglasses, fog, silhouette), as well as higher-level concepts (asleep, smiling), and find that the visualized concepts not only exhibit a high level of semantic consistency, but also that the automatically chosen names for the concepts accurately reflect the common feature in the images, despite coming from very different datasets.
This highlights the promise of the SAE for disentangling representations into human interpretable concepts as well as of the proposed strategy for naming those concepts. Interestingly, as expected, we find that the accuracy of the ascribed names correlates with the cosine similarity between the text embedding $\clipdec(s_c)$ and the dictionary vector $\vec p_c$ (\cf\cref{eq:dicvec}), as we discuss next (see also \cref{fig:conceptaccuracy_userstudy}). This indicates that our naming strategy could be significantly improved with a larger vocabulary, as we also discuss in \cref{sec:discoverynaming:supernodes}.

\input{figures_tex/task_agnosticity}

\myparagraph{Quantitative.}
To not only rely on the visual assessment of a few selected samples, we perform quantitative evaluations to assess the concept consistency and naming accuracy. This is generally challenging as only a few datasets include concept labels, and, even if they do, might describe different concepts in the image than those that were extracted by our task-agnostic approach.
To address this, we perform a user study to evaluate concept accuracy. Specifically, we sort concepts based on how well their dictionary vector is aligned to the text embeddings of the name assigned to them (\cref{sec:method:name}), and sample concepts with high, intermediate, and low alignments. We then extract the top activating images for each concept from three datasets (\imagenet, \places, \cctm), and for each concept, we ask two questions: (1) how semantically consistent the concept is, \ie if the top activating images map to some human interpretable concept, and (2) how accurate the assigned name is, if so. To evaluate if our SAE yields more disentangled concepts, we also compare with the neurons from the CLIP image features, named using CLIP-Dissect \cite{oikarinen2022clip}, as a baseline. For full details, see \cref{supp:sec:implementation:userstudy}. In \cref{fig:conceptaccuracy_userstudy} (left), we report the distribution of consistency scores both for our discovered concepts and the CLIP baseline each for the high, intermediate, and low aligned concepts, 
 and find that our approach provides significantly more human interpretable concepts. Interestingly, for both sets of concepts, the consistency decreases as the alignment with the text embedding decreases, suggesting that some concepts are not human interpretable.
In \cref{fig:conceptaccuracy_userstudy} (right), we evaluate the concept consistency against name accuracy, and find that our assigned names score highly in terms of accurately representing the concept (top right) as compared to the baseline. Note that some concepts, despite being consistent, are not named accurately, which could also be because of limitations in the vocabulary used; for more discussion, see \cref{sec:discoverynaming:supernodes}.

In addition to the human evaluation, we also perform a small quantitative evaluation using the SUNAttributes dataset \cite{patterson2014sun}, following \cite{panousis2023sparse}. We use its labelled attributes as the vocabulary for naming the nodes in the SAE (\cref{sec:method:name})  to match discovered concepts to ground truth labels. To account for concepts outside the labelled attribute set, we filter out nodes where the cosine similarity between the dictionary vector and the assigned text embedding is below a threshold, and merge concepts assigned to the same name. As a baseline, we compare against images obtained using CLIP retrieval from the ground truth attributes. We obtain a Jaccard index of 18.3, as compared to 22.0 for the CLIP retrieval baseline (for comparison, \cite{panousis2023sparse} report a Jaccard index of 15.7 under a similar setting) despite not optimizing the SAE to learn dataset-specific concepts.

\input{figures_tex/concept_accuracy}

\subsection{Impact of Vocabulary on Concept Name Granularity}
\label{sec:discoverynaming:supernodes}

As seen in \cref{sec:discoverynaming:taskagnosticity} and \cref{fig:conceptaccuracy_userstudy}, some of the SAE nodes may not map to human interpretable concepts (\cref{fig:conceptaccuracy_userstudy}, left), or may not be named appropriately (\cref{fig:conceptaccuracy_userstudy}, right). The latter could be a result of limitations in the vocabulary: it being finite and only consisting of single words, it is possible that even concepts that the SAE discovers cannot be named accurately.

To explore this, in \cref{fig:supernodes_qualitative} we visualize examples of concept pairs that are originally assigned the same name (e.g. right: `bridges'), but visually correspond to distinct modalities of the concept. We find that a more fine-grained name is assigned to the concept when added to the vocabulary $\vocab$ (e.g. `arch bridge', `suspension bridge'). Conversely, removing the assigned name `bridge' from the vocabulary leads to worse names being assigned (e.g. `prague', `lisbon'; interestingly, note that the cities contain a prominent arch and suspension bridge, respectively). This suggests that the granularity and size of the vocabulary can significantly affect the name accuracy, and can also serve as a tool for practitioners to control  the granularity of assigned names depending on the use case.

\input{figures_tex/supernodes_qualitative}

\subsection{Clustering Concept Vectors}
\label{sec:discoverynaming:metaclustering}
To further measure semantic consistency, we also evaluate how well semantically related concepts cluster together in the latent concept space. To do this, we perform K-Means clustering on the concept representations across all images in the \places dataset, and visualize a random selection of clusters. For each cluster, we compute the cluster centroid and then visualize the strongest concepts. We find that semantically similar concepts and their associated images cluster together in concept space (\eg farming related concepts and images in the right), showing that our concept-based (latent) representation does indeed result in semantically meaningful and nameable similarities.

\input{figures_tex/meta_clusters}

%% file: figures_tex/task_agnosticity.tex
\begin{figure}[t]
    \centering
    \begin{subfigure}[c]{.85\textwidth}
    \includegraphics[width=\linewidth]{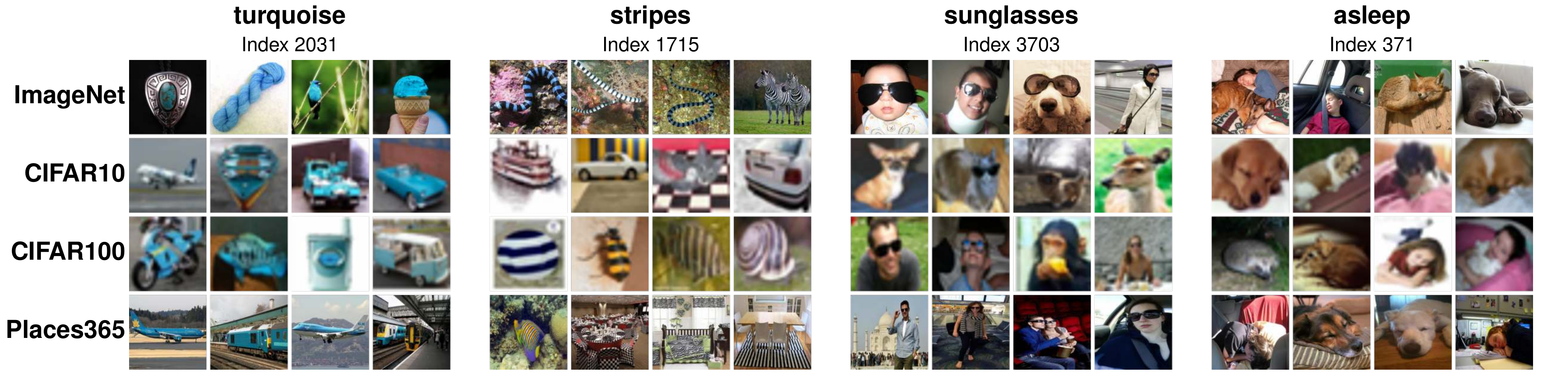}
    \end{subfigure}\\
    \begin{subfigure}[c]{.85\textwidth}
    \includegraphics[width=\linewidth]{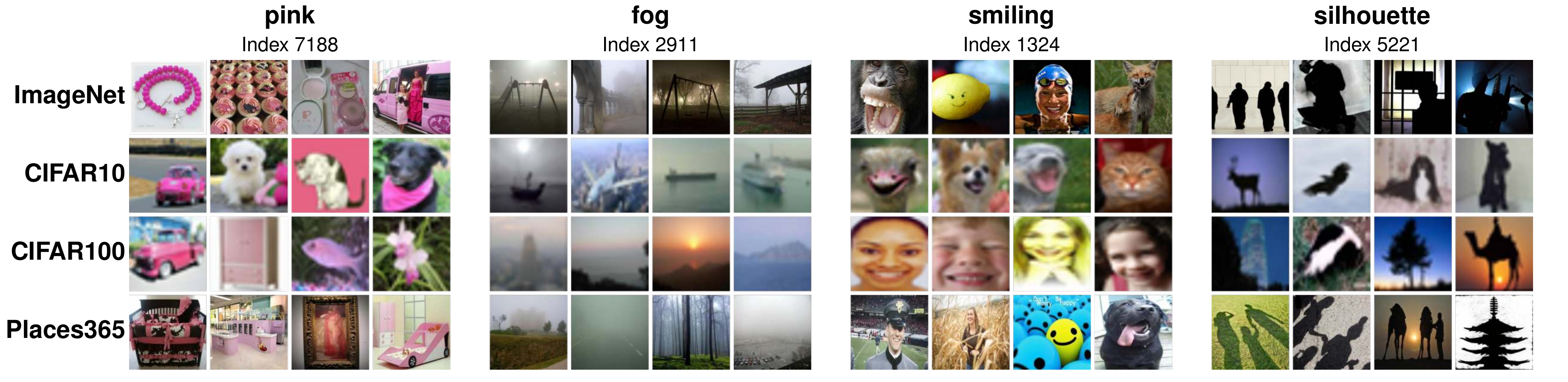}
    \end{subfigure}
    \caption{\textbf{Task-agnosticity of concept extraction.} We show examples of named concepts (blocks) and top images activating them from four datasets (rows). We find that the images activating the concept are highly consistent with the concept name across datasets (\eg the `asleep' concept yields images across different species), despite not using these datasets for extraction and naming, showing the robustness of our approach.
    }
    \label{fig:taskgnostic}
\end{figure}

%% file: figures_tex/concept_accuracy.tex
\begin{figure}[t!]
    \centering
    \includegraphics[width=.95\linewidth]{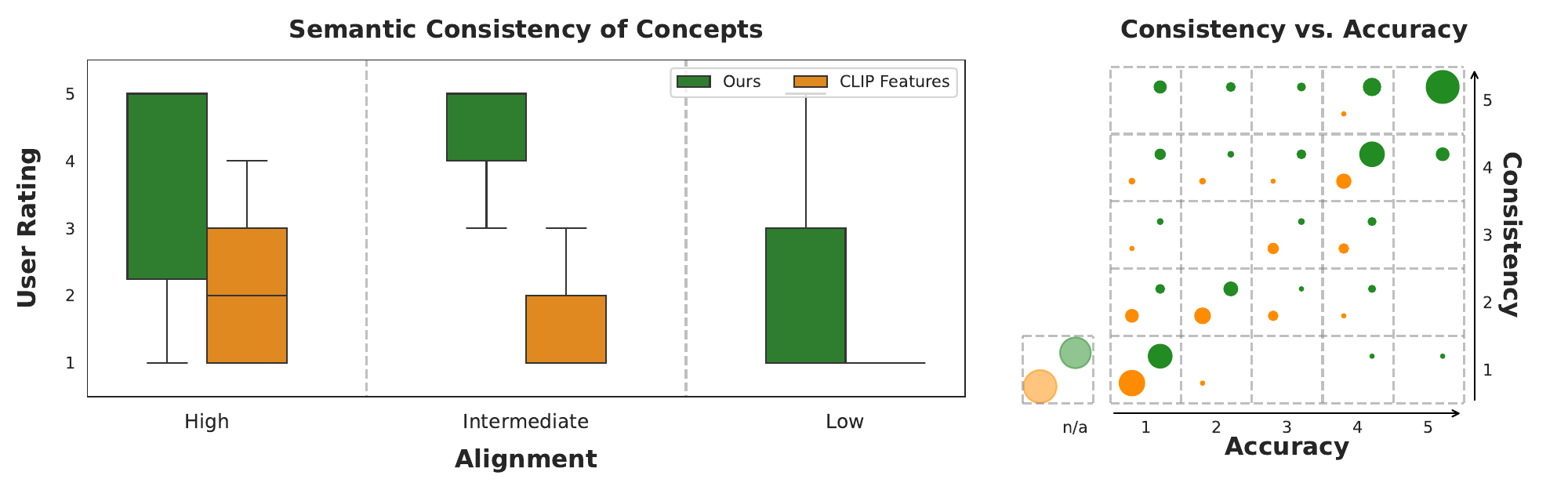}
    \caption{\textbf{User study on concept accuracy.} \emph{Left:} We evaluate the semantic consistency
    of concepts for nodes with 
    high, intermediate, and low
    alignment with the text embeddings of the name assigned to them, both for nodes from our SAE (green) and the CLIP features (orange). We find that the concepts from the SAE are significantly more semantically consistent than CLIP features, and the consistency increases with
    alignment. The poor performance of the `low alignment' group suggests that some nodes do not correspond to a consistent human interpretable concept. \emph{Right:} We plot the scores for semantic consistency against name accuracy from human evaluators, both for nodes from our SAE (green) and the CLIP features (orange). We find that compared to the baseline, our SAE nodes are generally more consistent and accurately named.}
\label{fig:conceptaccuracy_userstudy}
\end{figure}

%% file: figures_tex/supernodes_qualitative.tex
\begin{figure}[t]
    \centering
    \begin{subfigure}[c]{.495\textwidth}
    \includegraphics[width=\linewidth]{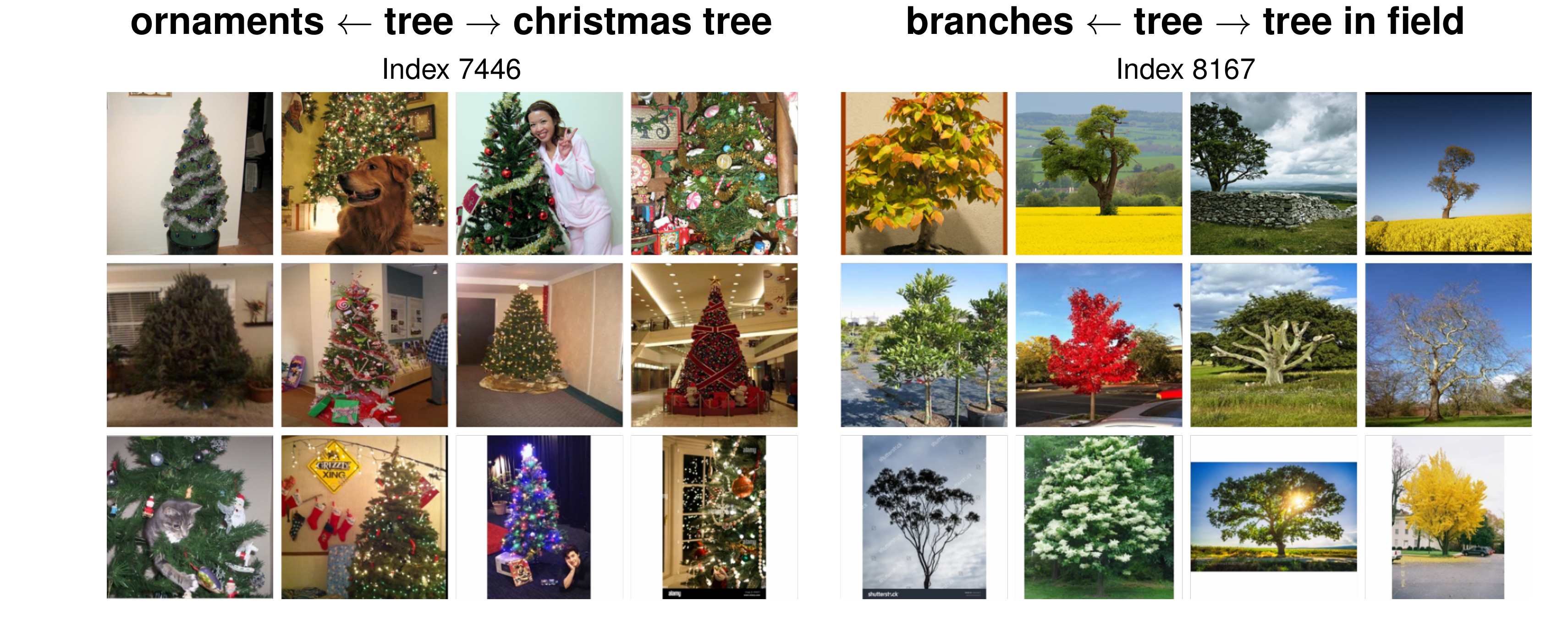}
    \end{subfigure}
    \hfill
    \begin{subfigure}[c]{.495\textwidth}
    \includegraphics[width=\linewidth]{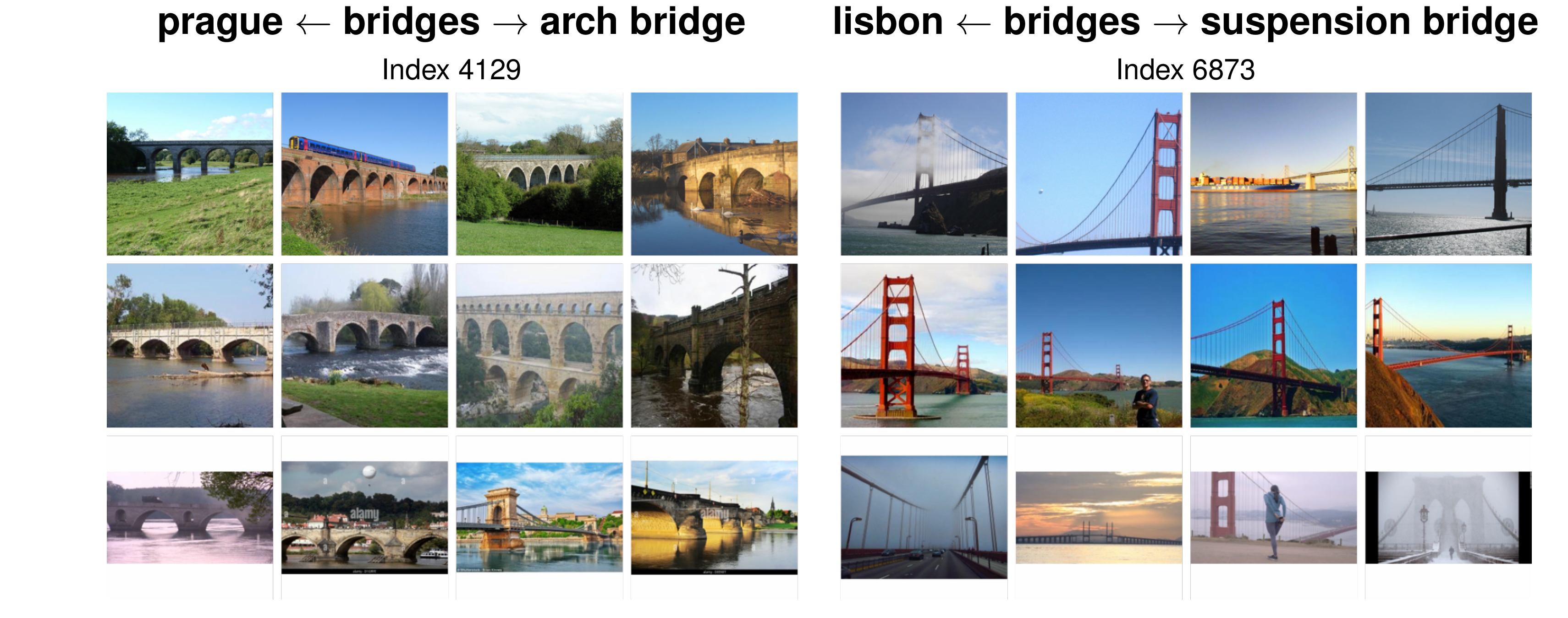}
    \end{subfigure}
    \caption{\textbf{Impact of vocabulary.} We show examples of pairs of concepts that, despite being assigned to the same coarse grained name (e.g. left: `tree'), correspond to distinct fine-grained concepts. Better names that can distinguishing such concepts are assigned if added to the vocabulary (e.g. `christmas tree' for the first concept, and `tree in field' for the second). On the other hand, removing the assigned name from the vocabulary leads to worse names being assigned (e.g. `ornaments' and `branches'), which shows that the granularity of the vocabulary can impact name accuracy.}
    \label{fig:supernodes_qualitative}
\end{figure}

%% file: figures_tex/meta_clusters.tex
\begin{figure}
    \centering
    \begin{subfigure}[c]{.95\textwidth}
    \includegraphics[trim=0 0 0 2cm, clip, width=\linewidth]{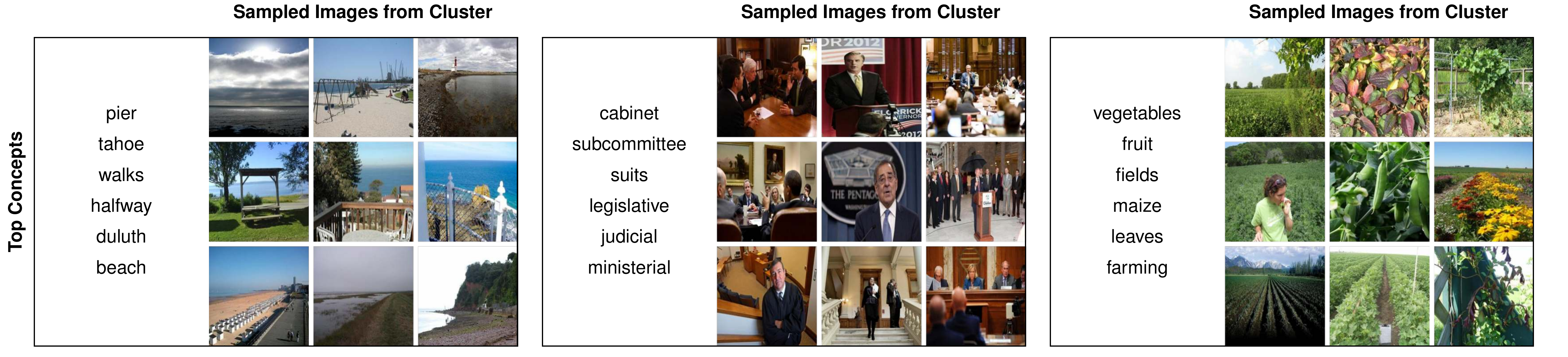}
    \end{subfigure}
    \caption{\textbf{Extracting meaningful clusters from concept strength vectors.}
    We perform K-Means clustering over concept activation vectors on the \places dataset to evaluate the semantic consistency of these latent representations. We show a random subset of clusters: each block represents a cluster, and we show top concepts from the cluster centroid and randomly selected images assigned to the cluster. We find that highly semantically consistent clusters of concepts emerge (\eg right: concepts and images from classes related to farming are grouped together).}
    \label{fig:metaclusters}
\end{figure}

%% file: 5_results.tex
\section{Evaluation of DN-CBM}
\label{sec:results}

We now present results on the concept bottleneck models (DN-CBM) (\cref{sec:method:cbm}) built on the discovered and named concepts (Secs.\ \ref{sec:method:extract}, \ref{sec:method:name}), evaluating accuracy (\cref{sec:results:accuracy}), interpretability (\cref{sec:results:interpretability}), and effectiveness of interventions (\cref{sec:results:interventions}).

\myparagraph{Setup.}
Similar to prior work \cite{oikarinen2023label,panousis2023sparse}, we train linear 
classifiers on top of the extracted concepts on four datasets---\imagenet\cite{deng2009imagenet}, \cifart\cite{krizhevsky2009learning}, \cifarh\cite{krizhevsky2009learning}, and \places\cite{zhou2017places}---and evaluate them for accuracy and interpretability. 
We train with various hyperparameters and pick the configurations based on performance on a heldout set. We compare our CBMs with recently proposed label-free approaches: \lfcbm\cite{oikarinen2023label}, \labo\cite{yang2023language}, \dclip\cite{menon2022visual} and \cdm\cite{panousis2023sparse}, and also report the linear probe and zero-shot performance of the \clip model we use as a backbone for reference. We use the respective concept sets of each baseline method, and for a fair comparison, 
the same feature extractor across methods.

\subsection{Classification Performance}
\label{sec:results:accuracy}

\input{tables/table_accuracy}
In \cref{tab:accuracy}, we show the classification performance of our \dncbm on four datasets and two feature extractors and compare them with the baselines. We find that \dncbm is highly performant across datasets and backbones. Despite being task-agnostic, \dncbm almost always outperforms the baselines, which use concept sets optimized for the downstream task, showing the generality of our approach. The highest gains are with \places (\ie 52.70$\to$53.53~pp.\ on \resnetf and 52.58$\to$55.11~pp.\ on \vitb), which is a scene-classification dataset rich in a wide variety of objects, which correspond to coarser, higher level concepts than \eg body parts of animals as in \imagenet or \cifart, and are likely more well-represented in our concept space trained on \cctm.

\subsection{Interpretability of DN-CBM}
\label{sec:results:interpretability}

\myparagraph{Local Explanations (Image-Level).}
\input{figures_tex/local_explanations}
\input{figures_tex/cbm_comparison}
In \cref{fig:localexp}, we show qualitative examples of \emph{local} explanations from our \dncbm, \ie, explanations of individual decisions. For each image, we show the most contributing concepts
along with their contribution strengths. We find that the concepts used are intuitive and class-relevant, thus aiding interpretability. The concepts used are also diverse, and include objects in the scenes (\eg `rocks' for `swimming hole', top-right), similar features (\eg `toaster' given the corroded surface for `junkyard', bottom-left), and high-level concepts (\eg `abandoned' for `junkyard', bottom-left). Interestingly, we also find concepts associated with the class (\eg `alps' or `everest' for `glacier', top-left), which shows that the model's decision is also based on what a scene \emph{looks like}, akin to ProtoPNets \cite{chen2019looks}. Finally, we observe that the concepts for predicting the same class change
based on the contents of the image, \eg in the bottom row, we find that despite both images depicting a junkyard where the most influential concept is `corrosion', 
the second highest concepts are `car' and `tractor' respectively, reflecting the image contents.

In \cref{fig:cbmcompare}, we also compare explanations from \dncbm with baselines (\lfcbm, \cdm) on the same images from \places. 
Interestingly, we find that our approach yields similarly convincing explanations as prior state-of-the-art CBM models, despite the fact that it does not use a task-specific vocabulary and extracts the concepts on a separate dataset (CC3M).

\myparagraph{Global Explanations (Class-level).}
\input{figures_tex/global_explanations}
In \cref{fig:globalexp}, we show qualitative examples of \emph{global} explanations from our \dncbm, \ie, explanations of which concepts contribute the most to a class \emph{as a whole}. To do this, for each class, we compute the average contribution of all concepts for images from that class, and visualize the set of top concepts. Qualitatively, we find this set to be semantically consistent with what is contained in each class.

\subsection{Effectiveness of Concept Interventions}
\label{sec:results:interventions}

In addition to understanding model decisions, explanations have also been used
to debug models \cite{koh2020concept} and fix models' reasoning \cite{ross2017right,petryk2022gals,rao2023studying}. Specifically, concept bottleneck models allow human interventions on individual concepts to control the models' reliance on them. We assess the effectiveness of our DN-CBM with interventions by training on the Waterbirds-100 \cite{sagawa2020distributionally,petryk2022gals} dataset. This contains images of Landbirds and Waterbirds, with landbirds (waterbirds) on land (water) backgrounds during training, but without any such correlation in the test set. Following \cite{petryk2022gals,rao2022towards} we evaluate if intervening to (1) only keep bird related concepts, and (2) only remove such concepts increases (respectively, decreases) the performance on the worse group classification. To do this, we train a DN-CBM model that uses only five concepts for each class. For full details, see \cref{supp:sec:implementation:interventions}.

In \cref{tab:waterbirds}, we report the accuracy before and after
the two interventions. We find that
keeping only bird related concepts significantly improves the overall and worst group (Landbird on Water, Waterbird on Land) accuracies, with only a small drop in the other groups. Similarly, removing only such concepts leads to a large drop in accuracies, showing the effectiveness of interventions.

\input{tables/table_waterbirds}

%% file: tables/table_accuracy.tex
\begin{table}
    \centering
    \caption{\textbf{Performance of our CBM in comparison to prior work.} We report the classification accuracy ($\%$) of our CBM and baselines using \clip\resnetf and \vitb feature extractors (\vitl in \cref{supp:sec:quantitative})
    on \imagenet, \places, \cifart, and \cifarh. We find that our CBM performs competitively and often outperforms prior work, despite using a common set of concepts across datasets. `*' indicates results reported for the respective baselines, and zero-shot performance is as reported by \cite{radford2021learning}.
    }
    \begin{tabular}{l@{\hskip4pt}c@{\hskip1pt}c@{\hskip6pt}c@{\hskip6pt}c@{\hskip6pt}c@{\hskip14pt}c@{\hskip6pt}c@{\hskip6pt}c@{\hskip6pt}c}\toprule
     \multirow{2}{*}{\footnotesize
     Model} & \multirow{2}{*}{\scriptsize
     \begin{tabular}{c}\scriptsize Task\\\scriptsize  Agnostic\end{tabular}} 
     &
    \multicolumn{4}{c}{\bf \clip\resnetf\phantom{shift}}&\multicolumn{4}{c}{\bf \clip \vitb\phantom{shift}}\\
    & & \scriptsize{IMN} & \scriptsize{Places} & \scriptsize{Cif10} & \scriptsize{Cif100} & \scriptsize{IMN} & \scriptsize{Places} & \scriptsize{Cif10} & \scriptsize{Cif100} \\
    \midrule
    Linear Probe & - & 73.3* & 53.4 & 88.7* & 70.3* & 80.2* & 55.1 & 96.2* & 83.1* \\
    Zero Shot & - & 59.6* & 38.7 & 75.6* & 41.6* & 68.6* & 41.2 & 91.6* & 68.7* \\
    \midrule 
    \lfcbm\cite{oikarinen2023label} & \xmark & 67.5\phantom{*} & 49.0\phantom{*} & 86.4* & 65.1* & 75.4\phantom{*} & 50.6\phantom{*} & 94.6\phantom{*} &  77.4\phantom{*} \\
    
    \labo\cite{yang2023language} & \xmark & 68.9\phantom{*} & - & \textbf{87.9}* & \textbf{69.1}* & 78.9\phantom{*} & - & 95.7\phantom{*} & 81.2\phantom{*} \\
    
    \cdm\cite{panousis2023sparse} & \xmark & 72.2* & 52.7* & 86.5* & 67.6* & 79.3* & 52.6* & 95.3* & 80.5* \\
    
    \dclip\cite{menon2022visual} & \xmark & 59.6\phantom{*} & 37.9\phantom{*} & - & - & 68.0* & 40.3* & - & -  \\
    \midrule
    \dncbm (Ours) & \cmark & \textbf{72.9}\phantom{*} & \textbf{53.5}\phantom{*} & 87.6\phantom{*}  & 67.5\phantom{*} &  \textbf{79.5}\phantom{*} & \textbf{55.1}\phantom{*} & \textbf{96.0}\phantom{*} & \textbf{82.1}\phantom{*} \\
    \bottomrule
    \end{tabular}
\label{tab:accuracy}
\end{table}

%% file: figures_tex/local_explanations.tex
\begin{figure}[t]
    \centering
    \begin{subfigure}[c]{.4\textwidth}    \includegraphics[width=\linewidth]{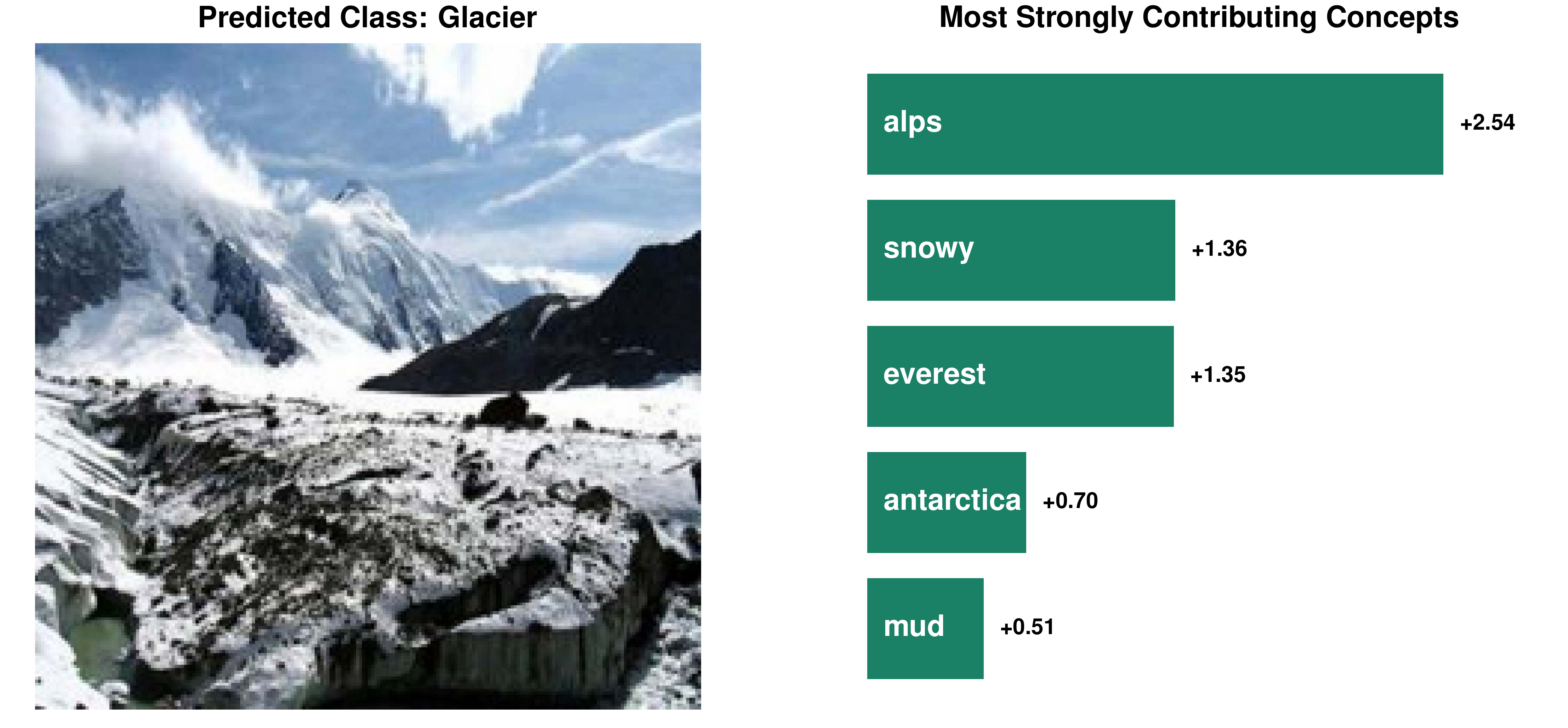}
    \end{subfigure}\hfill
    \begin{subfigure}[c]{.4\textwidth}    \includegraphics[width=\linewidth]{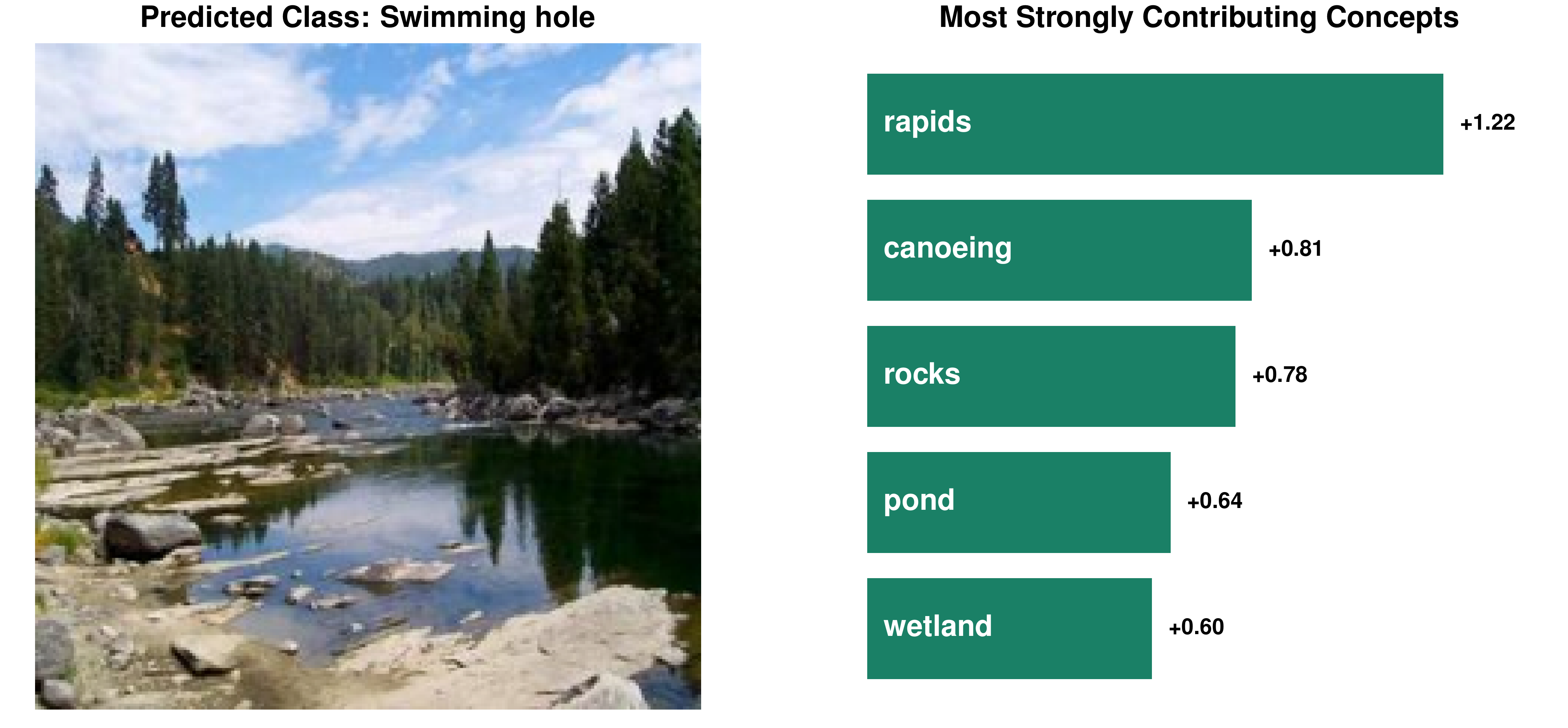}
    \end{subfigure}\\
    \begin{subfigure}[c]{.4\textwidth}    \includegraphics[width=\linewidth]{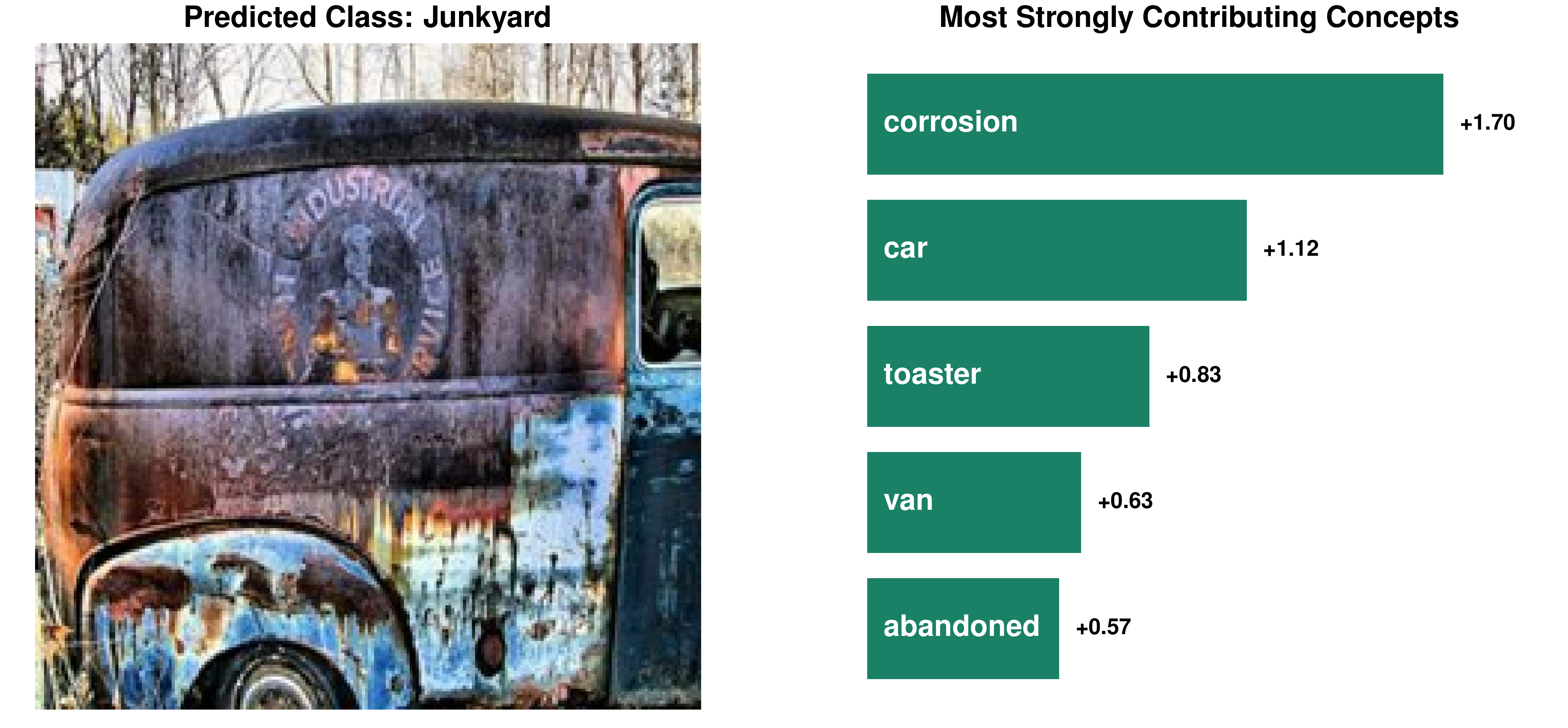}
    \end{subfigure}\hfill
    \begin{subfigure}[c]{.4\textwidth}    \includegraphics[width=\linewidth]{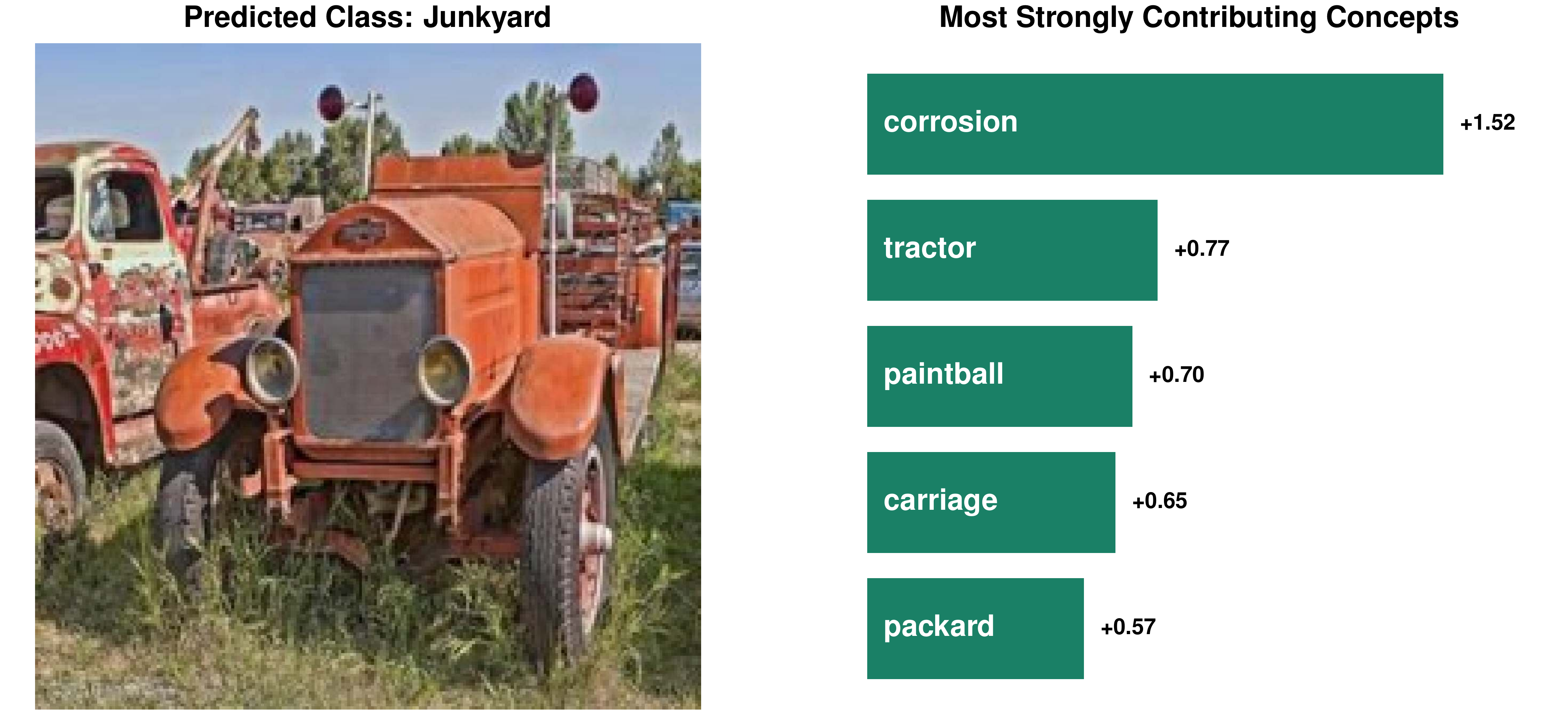}
    \end{subfigure}\\
    \caption{\textbf{Explaining decisions using our CBM.} We show examples of images from the \places dataset along with the top concepts contributing to the decision. We find that our CBM classifies based on a diverse set of concepts present in the image, including objects, similar features, higher level concepts, and things associated with the class (e.g. similar locations), thus aiding interpretability.
    }
    \label{fig:localexp}
\end{figure}

%% file: figures_tex/cbm_comparison.tex
\begin{figure}[t!]
    \centering
    \begin{subfigure}[c]{\textwidth}
    \centering
    \includegraphics[ width=0.9\linewidth]{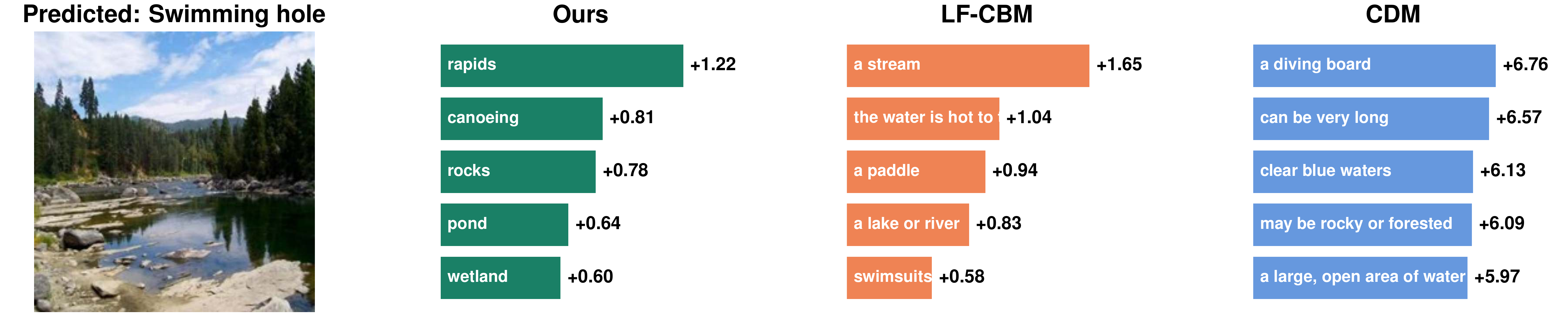}
    \end{subfigure}
    \begin{subfigure}[c]{\textwidth}
    \centering
    \includegraphics[width=0.9\linewidth]{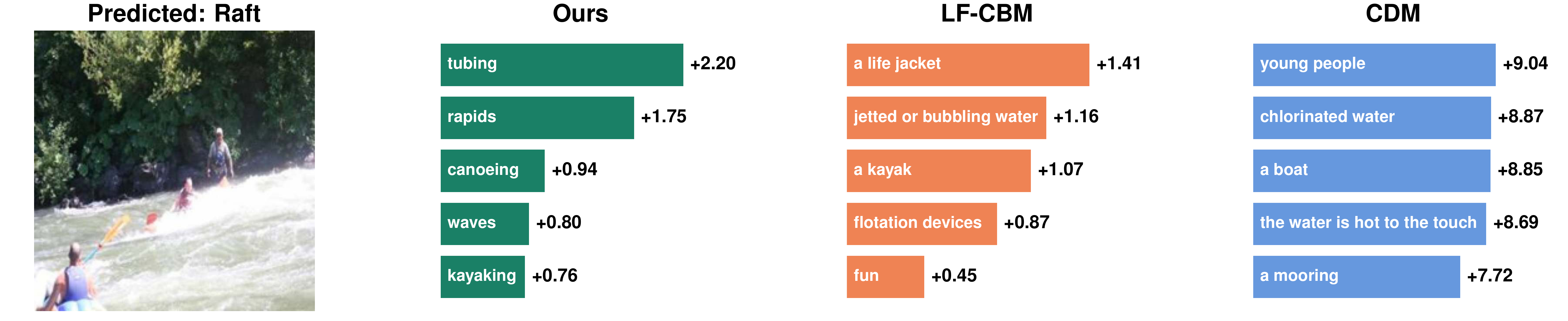}
    \end{subfigure}
    \caption{\textbf{Comparing interpretability across CBMs.} We show an example from the \places dataset with explanations consisting of top contributing concepts using our CBM, \lfcbm\cite{oikarinen2023label}, and \cdm\cite{panousis2023sparse}. We find that our approach yields similar explanations despite not querying LLMs for concepts specific to the task and instead using a single task-agnostic concept bottleneck layer that is named post hoc. 
    }
    \label{fig:cbmcompare}
\end{figure}

%% file: figures_tex/global_explanations.tex
\begin{figure}[t!]
    \centering
    \begin{subfigure}[c]{.95\textwidth}
    \includegraphics[width=\linewidth]{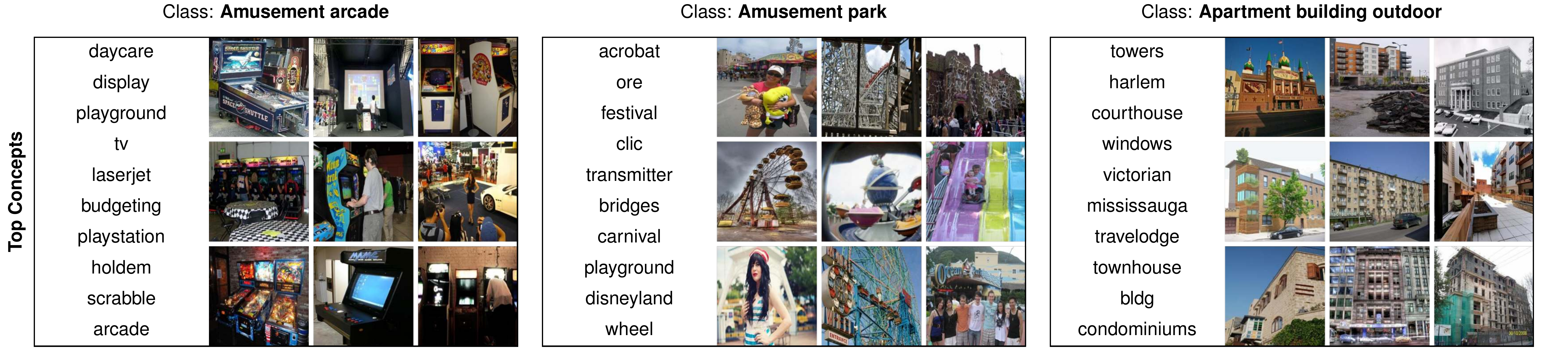}
    \end{subfigure}
    \caption{\textbf{Class-wise explanations of CBMs.} We show examples of classes from the \places dataset with the top contributing concepts. For each class, we show random examples of images belonging to that class and select concepts with the highest average contribution across all images from the class in the validation set. We find that our approach yields classifiers that use concepts highly semantically related to each class.}
    \label{fig:globalexp}
\end{figure}

%% file: tables/table_waterbirds.tex
\begin{table}
    \centering
    \caption{
    \textbf{Performance before and after intervening on the concept bottleneck model trained for the Waterbirds-100 dataset.} We report the classification accuracy (\%) on the full test set (`Overall') and each of the four groups (\eg Landbird on Water, shown as `L.Bird@W') before and after applying interventions. We find that intervening to only keep bird relevant concepts increases the overall and worst group \cite{sagawa2020distributionally} (Landbird on Water, Waterbird on Land) accuracy significantly, and conversely removing exactly these concepts leads to a large drop in accuracy, without adversely affecting performance on the groups in the training set (`Training Groups').
    }
    \begin{tabular}{l@{\hskip8pt}c@{\hskip8pt}c@{\hskip8pt}c@{\hskip8pt}c@{\hskip8pt}c@{\hskip8pt}}\toprule
    \multirow{2}{*}{\bf Model} & \multirow{2}{*}{\bf Overall} & \multicolumn{2}{c}{\bf Worst Groups} & \multicolumn{2}{c}{\bf Training Groups} \\
    & & \scriptsize\bf L.Bird@W & \scriptsize\bf W.Bird@L & \scriptsize\bf L.Bird@L & \scriptsize\bf W.Bird@W \\
    \midrule
    \scriptsize Before Intervention & 82.8 & 71.3 & 57.5 & \bf 98.6 & \bf 93.3 \\
    \scriptsize Only Bird Concepts & \bf 89.4 \scriptsize({\color{forestgreen}{\Plus6.6}}) & \bf 86.6 \scriptsize({\color{forestgreen}{\Plus15.3}}) & \bf 71.3 \scriptsize({\color{forestgreen}{\Plus13.8}}) & 96.8 \scriptsize({\color{red}{\Minus1.8}}) & 91.4 \scriptsize({\color{red}{\Minus1.9}}) \\
    \scriptsize No Bird Concepts & 60.8 \scriptsize({\color{red}{\Minus22.0}}) & 28.5 \scriptsize({\color{red}{\Minus42.8}}) & 28.8 \scriptsize({\color{red}{\Minus28.7}}) & 95.0 \scriptsize({\color{red}{\Minus3.6}}) & 85.8 \scriptsize({\color{red}{\Minus7.5}}) \\
    \bottomrule
    \end{tabular}
\label{tab:waterbirds}
\end{table}

%% file: 6_conclusion.tex
\section{Conclusion}
\label{sec:conclusion}

In this work, we proposed Discover-then-Name CBM (\dncbm), a novel CBM approach that uses sparse autoencoders to discover and automatically name concepts learnt by CLIP, and then use the learnt concept representations as a concept bottleneck and train linear layers for classification. We find that this simple approach is surprisingly effective at yielding semantically consistent concepts with appropriate names. Further, we find despite being task-agnostic, \ie only extracting and naming concepts once, our approach can yield performant and interpretable CBMs across a variety of downstream datasets. Our results further corroborate the promise of sparse autoencoders for concept discovery. Training a more `foundational' sparse autoencoder with a much larger dataset (\eg at CLIP scale) and concept space dimensionality (with hundreds of thousands or millions of concepts) to obtain even more general-purpose CBMs, particularly for fine-grained classification, would be a fruitful area for future research.

%% file: supplement/0_main.tex
\appendix 

\renewcommand\thesection{\Alph{section}}
\numberwithin{equation}{section}
\numberwithin{figure}{section}
\numberwithin{table}{section}
\renewcommand{\thefigure}{\thesection\arabic{figure}}
\renewcommand{\thetable}{\thesection\arabic{table}}
\crefname{appendix}{Sec.}{Secs.}

{\begin{center}
\Large\bf
\phantom{skip}\\[.25em]
{Discover-then-Name: Task-Agnostic Concept Bottlenecks via Automated Concept Discovery}\\[1em]
\large
Appendix
\end{center}
}

\newcommand{\additem}[2]{%
\item[\textbf{(\ref{#1})}] 
    \textbf{#2} \dotfill\makebox{\textbf{\pageref{#1}}}
}
\newcommand{\addsubitem}[2]{%
    \textbf{(\ref{#1})}\hspace{1em}
    #2\\[.1em] 
}

\newcommand{\adddescription}[1]{\newline
\begin{adjustwidth}{0cm}{0cm}
#1
\end{adjustwidth}
}

{\noindent\bf\large Table of Contents\\[1em]}
\vspace{-20pt}
\\[1em]

{
\begin{adjustwidth}{1cm}{1cm}
\begin{enumerate}[label={({\arabic*})}, topsep=1em, itemsep=.75em]
    \additem{supp:sec:limitations}{Limitations and Broader Impact}
    \adddescription{In this section, we discuss the limitations of our work and broader impact.}
    \additem{supp:sec:implementation}{Implementation Details}
    \adddescription{In this section, we provide training and implementation details for our evaluations.\\\\
        \addsubitem{supp:sec:implementation:training}{Training SAEs and DN-CBM}
        \addsubitem{supp:sec:implementation:userstudy}{User Study}
        \addsubitem{supp:sec:implementation:conceptaccuracy}{Concept Accuracy on SUNAttributes}
        \addsubitem{supp:sec:implementation:interventions}{Applying Interventions on DN-CBM}
        }
    \additem{supp:sec:quantitative}{Additional Quantitative Results}
    \adddescription{In this section, we provide additional quantitative results across backbones and datasets.\\\\
        \addsubitem{supp:sec:quantitative:classification}{DN-CBM Classification Performance}
        \addsubitem{supp:sec:quantitative:sparsity}{Sparsity of DN-CBM Explanations}
    }
    \additem{supp:sec:qualitative}{Additional Qualitative Results}
    \adddescription{In this section, we provide additional qualitative results across backbones and datasets.\\\\
        \addsubitem{supp:sec:qualitative:taskagnosticity}{Task Agnosticity of Concepts}
        \addsubitem{supp:sec:qualitative:clustering}{Clustering Concept Strength Vectors}
        \addsubitem{supp:sec:qualitative:local}{Local Explanations from \dncbm}
        \addsubitem{supp:sec:qualitative:global}{Global Explanation from \dncbm}
    }

\end{enumerate}
\end{adjustwidth}
}

\clearpage
\input{supplement/4_limitations_impact}
\clearpage
\input{supplement/3_implementation}

\clearpage
\input{supplement/2_quantitative}

\clearpage 
\input{supplement/1_qualitative}

\clearpage

%% file: supplement/4_limitations_impact.tex
\section{Limitations and Broader Impact}
\label{supp:sec:limitations}

In this work, we proposed \dncbm, as task-agnostic approach to discovering concepts from \clip\citeS{radford2021learningS} using sparse autoencoders (SAEs) and then using them to construct a concept bottleneck model across downstream classification tasks. We find that this simple approach shows promise that using a higher dimensional concept space and training SAEs on much larger datasets than \cctm \citeS{sharma2018conceptualS} would provide a richer and more diverse concept representation. We already find that concept discovery from our SAEs using a relatively small dataset like \cctm can lead to performant and interpretable CBMs for datasets such as \places \citeS{zhou2017placesS}, leveraging concepts related to objects, colours, shapes, locations, and associations to reach decisions in an interpretable manner. However, these constitute relatively coarse concepts, and given the nature and size of the \cctm dataset, we do not find highly fine-grained concepts (\eg bird body parts as used in the CUB dataset \citeS{wah2011thecaltechS}), and mitigating this by scaling up to a ``foundational'' SAE could be a promising direction for future research. Further, one could also explore using better vocabularies for concept naming, \eg that are more tailored to the type of dataset used for SAE training. Further, note that spurious correlations learnt by \clip would likely persist in our concept discovery, leading to concepts being activated when features correlated with the concept are present (\eg see concept `plane' activated for `Airport terminal' in the example in \cref{fig:localexp_places_misclas}, (c)). Mitigating this is orthogonal to this work and a fruitful direction for future research. Overall, we find that despite being simple, our approach is surprisingly effective in finding meaningful concepts, assigning them human interpretable names, and constructing performant classifiers in an efficient and task-agnostic manner, and can further the goal of building more interpretable models. 

%% file: supplement/3_implementation.tex
\section{Implementation Details}
\label{supp:sec:implementation}

In this section, we provide additional details for each of our evaluations in \cref{sec:discoverynaming} and \cref{sec:results}.

\subsection{Training SAEs and DN-CBM}
\label{supp:sec:implementation:training}

In this section, we describe the details for training the sparse autoencoders (SAEs) used for discovering and naming concepts (\cref{sec:method:extract,sec:method:name}), and then for constructing our concept bottleneck models (DN-CBM) (\cref{sec:method:cbm}). We implement our code for all our experiments using PyTorch \citeS{paszke2019pytorchS} and use Captum \citeS{kokhlikyan2020captumS} for visualization.

\myparagraph{Feature Extractors.} We use \clip \citeS{radford2021learningS} \resnetf \citeS{he2016deepS}, \vitb \citeS{dosovitskiy2021anS}, and \vitl \citeS{dosovitskiy2021anS} pre-trained feature extractors from the official repository\footnote{\href{https://github.com/openai/CLIP}{https://github.com/openai/CLIP}}. We use the output features (after pooling) for discovering concepts using sparse autoencoders.

\myparagraph{Datasets.} We train our sparse autoencoders on the \cctm dataset \citeS{sharma2018conceptualS}, and then train linear probes for classification on \imagenet\citeS{deng2009imagenetS}, \places\citeS{zhou2017placesS}, \cifart\citeS{krizhevsky2009learningS}, and \cifarh\citeS{krizhevsky2009learningS}. To speed up training, we pre-compute features and concept strengths respectively before training, and do not perform any augmentations. Performing such augmentations (\eg random cropping, flipping, \etc) would likely improve our classification performance further.

\begin{figure}[t!]
    \centering
    \includegraphics[width=0.4\textwidth]{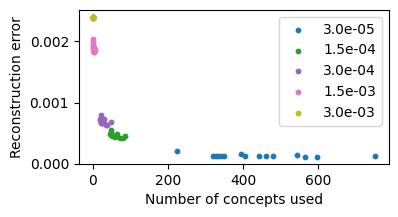}
\caption{\textbf{Impact of sparsity parameter $\lambda_1$ (\cref{eq:sae_loss}) on reconstruction.} We plot the reconstruction error versues the number of concepts used for various values of the $L_1$ sparsity constraint parameter $\lambda_1$. We find that higher $\lambda_1$ leads to sparser concept representations, but at the cost of a higher representation error.}
\label{fig:lambda}
\end{figure}

\myparagraph{Training Sparse Autoencoders (SAEs).} We train sparse autoencoders following the setup of \citeS{bricken2023monosemanticityS}, using the implementation of \citeS{cooney2023sparseS}\footnote{\href{https://github.com/ai-safety-foundation/sparse_autoencoder/}{https://github.com/ai-safety-foundation/sparse\_autoencoder}} (v1.3.0). We train for 200 epochs, and resample every 10 epochs. We perform hyperparameter sweeps using a heldout set over the learning rate $\{1\times10^{-5}, 5\times10^{-5}, 1\times10^{-4}, 5\times10^{-4}, 1\times10^{ -3}\}$, $L_1$ sparsity coefficient ($\lambda_1$) $\{3\times10^{-5}, 1.5\times10^{-4}, 3\times10^{-4}, 1.5\times10^{-3}, 3\times10^{-3}\}$, and expansion factors $\{2, 4, 8\}$. For the \clip \resnetf model, we choose the SAE with learning rate $5\times10^{-4}$, $L_1$ sparsity   $3\times10^{-5}$, expansion factor $8$ based on \imagenet zeroshot performance on reconstructions. See also \cref{fig:lambda} for an evaluation of the impact of $\lambda_1$ on reconstruction error and sparsity.

\myparagraph{Training Linear Probes.} We train linear probes without bias on the learned concept representations using the Adam optimizer \citeS{kingma2014adamS}. In addition to using the cross entropy loss for classification, we apply a $L_1$ sparsity constraint on the weights, and train for 200 epochs. We perform hyperparameter sweeps using a heldout set over the learning rate $\{1\times10^{-4}, 1\times10^{-3}, 1\times10^{-2}\}$, $L_1$ sparsity coefficient ($\lambda_2$) $\{0, 0.1, 1\}$. We train such probes over all trained SAEs and pick the probes with the best top-1 validation accuracy for each dataset.

\subsection{Concept Accuracy on SUNAttributes}
\label{supp:sec:implementation:conceptaccuracy}
In this section, we describe the process for quantitatively evaluating concept accuracy on the SUNAttributes dataset \citeS{patterson2014sunS}, as discussed in \cref{sec:discoverynaming:taskagnosticity}.
Following \cite{panousis2023hierarchical}, to quantify the agreement, we estimate the Jaccard index between the ground truth (GT) concepts and the predicted concepts per image, which is then averaged over all the images. The jaccard index is given by 

$$\text{Jaccard(GT, Predicted)} = \frac{M_{11}}{M_{11}+M_{10}+M_{01}}$$

where, $M_{11}$= true positives ; $M_{10}$= false negatives; $M_{01}$= false positives.

To evaluate the concept accuracy, we use the ground truth concepts of the SUNAttributes dataset. There are 102 GT concepts for SUNAttributes, which is of much smaller dimension than our SAE latent space.
To bring the SAE latent dimension down to the dimension of GT concepts, we cluster the dictionary vectors with the same concept name and treat it as a `compound' node. From each cluster, we filter the unactivated nodes and nodes with poor alignment between their dictionary vector and the associated text embedding. The compound node is then associated with the maximum concept strength of all the constituent individual nodes. We then use a heldout set to dynamically find a per concept threshold to binarize the concept strengths. For the CLIP retrieval baseline, we follow the same procedure on the cosine similarities of the nodes with the text embeddings.

\subsection{User Study}
\label{supp:sec:implementation:userstudy}

In this section, we describe the details of our user study to quantitatively measure the consistency and accuracy of our discovered and named concepts, as discussed in \cref{sec:discoverynaming:taskagnosticity}.

\myparagraph{Methods.} We evaluate on nodes from our SAE (\cref{supp:sec:implementation:training}) for the \clip \resnetf model, assigned with names from our naming scheme (\cref{sec:method:name}). As a baseline, we use nodes from the image feature vector of the same \clip \resnetf model, assigned names using CLIP-Dissect \citeS{oikarinen2022clipS} with the same vocabulary.

\myparagraph{Selecting Nodes.} To obtain a holistic view of the consistency and name accuracy of the node concepts, we sort nodes based on the alignment with the text embedding vector of the name assigned to them. Specifically, for the SAEs, we sort based on the cosine similarity between the dictionary vector and the text embedding, and for the CLIP features, we sort based on the CLIP-Dissect similarity. We then uniformly at random sample nodes from three bins, where the alignment is the highest, intermediate, and lowest. Specifically, for our SAE with 8192 nodes, we sample five nodes each from the top 2000, intermediate 4192, and bottom 2000 nodes. For the CLIP feature vector with 1024 nodes, we scale down and sample three nodes each from the top 250, intermediate 724, and bottom 250 nodes.

\myparagraph{Question Structure.} For each node, we extract the top four activating images from three diverse datasets -- \imagenet \citeS{deng2009imagenetS}, \places \citeS{zhou2017placesS}, and \cctm \citeS{sharma2018conceptualS} -- and create a grid of twelve images. Together with the image, we provide the word that corresponds to the name assigned to the node. We then ask two questions to evaluate: (1) if there is a semantically consistent and human interpretable concept common among the top activating images, and (2) if the word assigned accurately depicts such a concept, if any. For each question, participants are invited to rate on a five point scale, from 1 (``Strongly Disagree'') to 5 (``Strongly Agree''). For the second question, a ``Not Applicable'' option is also provided to account for the case that no common concept may be found by the participant. \cref{fig:supp:userstudy_example} shows an exmample of a question.

\myparagraph{Survey Structure.} We randomly order the 24 nodes (15 from our SAE, and 9 from CLIP-Dissect) in the survey. To help participants, we also provide examples at the beginning of the survey, as shown in \cref{fig:supp:survey_example} We published the survey internally, and received 22 responses.

\input{supplement/figures_tex/survey_example}

\input{supplement/figures_tex/userstudy_example}

\clearpage
\subsection{Applying Interventions on DN-CBM}
\label{supp:sec:implementation:interventions}

In this section, we describe the details of the evaluation by intervening on the concept bottleneck using the Waterbirds-100 \citeS{sagawa2020distributionallyS,petryk2022galsS} dataset on our DN-CBM, as discussed in \cref{sec:results:interventions}.

\myparagraph{Training Setup.} We use our SAE for the \clip \resnetf model, and train a linear probe for the binary classification task of the Waterbirds-100 dataset (\ie Landbird versus Waterbird) to obtain a concept bottleneck model. Specifically, we use a learning rate of $0.1$, $L_1$ sparsity coefficient of $10$, and train for 200 epochs. To further improve sparsity, we prune the weights for each class, leaving only the five largest weights and replacing the rest with zeroes. This model obtains an accuracy of 82.8\% (see \cref{tab:waterbirds}, `Before Interventions').

\myparagraph{Group-wise Evaluation.} Following \citeS{sagawa2020distributionallyS,petryk2022galsS,rao2023studyingS}, we also report the performance at a group-wise level, \ie both for the groups found in training (`Landbird on Land', `Waterbird on Water') and new groups only found in the test set (also known as the `Worst Groups', \ie `Landbird on Water', `Waterbird on Land').

\myparagraph{Interventions.} For each class, we look at the assigned names of the five concepts and manually classify them as `Bird Concept' and `Not Bird Concept'; for the full list, see \cref{tab:supp:concepts_waterbirds}. We find that the bird concepts typically correspond to examples of the type of bird (\eg `sparrow' for `Landbird', `gull' for `Waterbird'), which we attribute to the granularity of examples in the \cctm dataset used for the training the SAE and the granularity of the vocabulary. We perform two sets of interventions: (1) keeping only the bird concepts, and (2) removing only the bird concepts. For a full discussion on the results, see \cref{sec:results:interventions}.

\input{supplement/tables/table_concepts_waterbirds}

%% file: supplement/figures_tex/survey_example.tex
\begin{figure}[t!]
    \centering
    \begin{subfigure}[c]{.45\textwidth}
    \includegraphics[width=\linewidth]{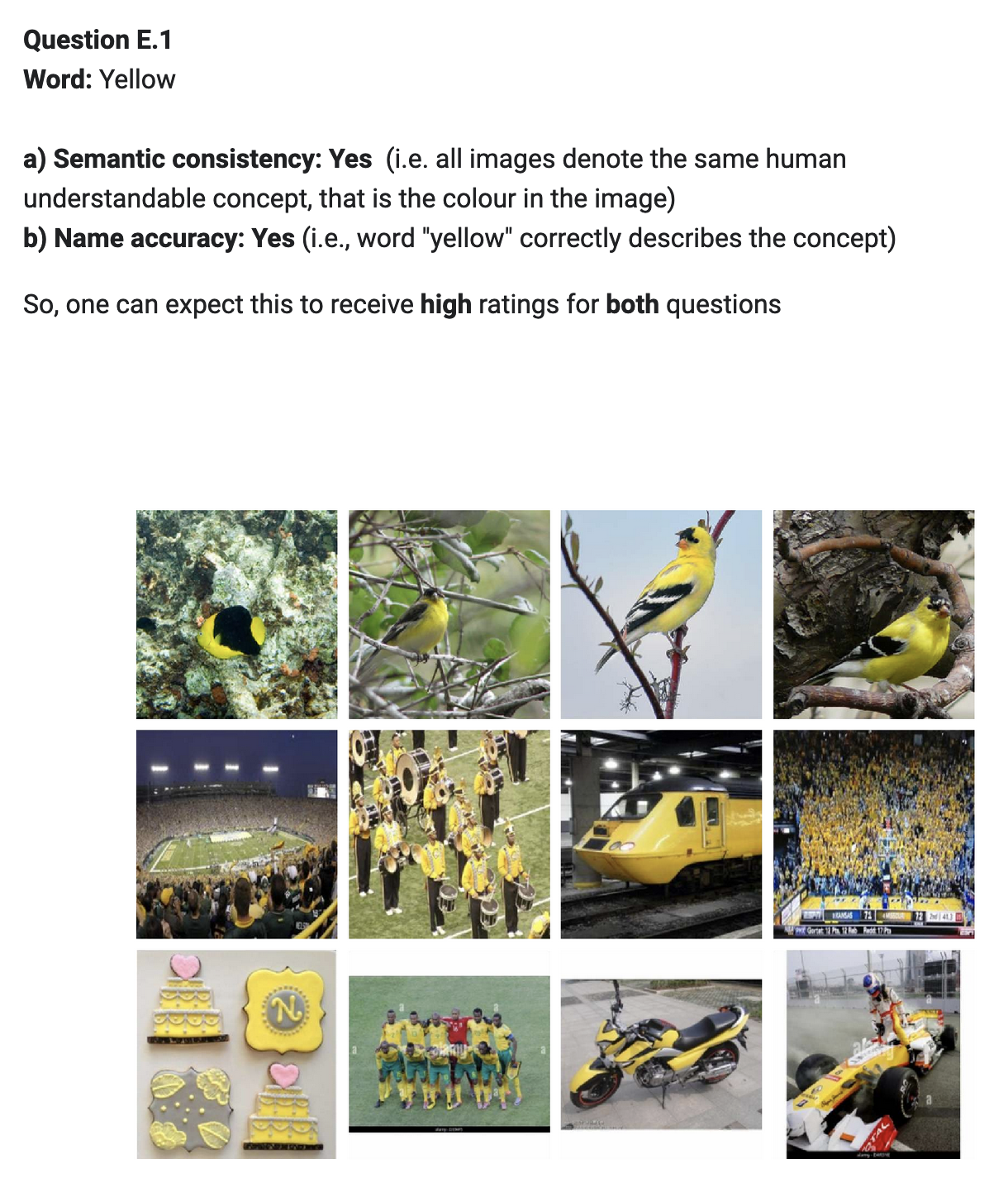}
    \end{subfigure}
    \begin{subfigure}[c]{.45\textwidth}
    \includegraphics[width=\linewidth]{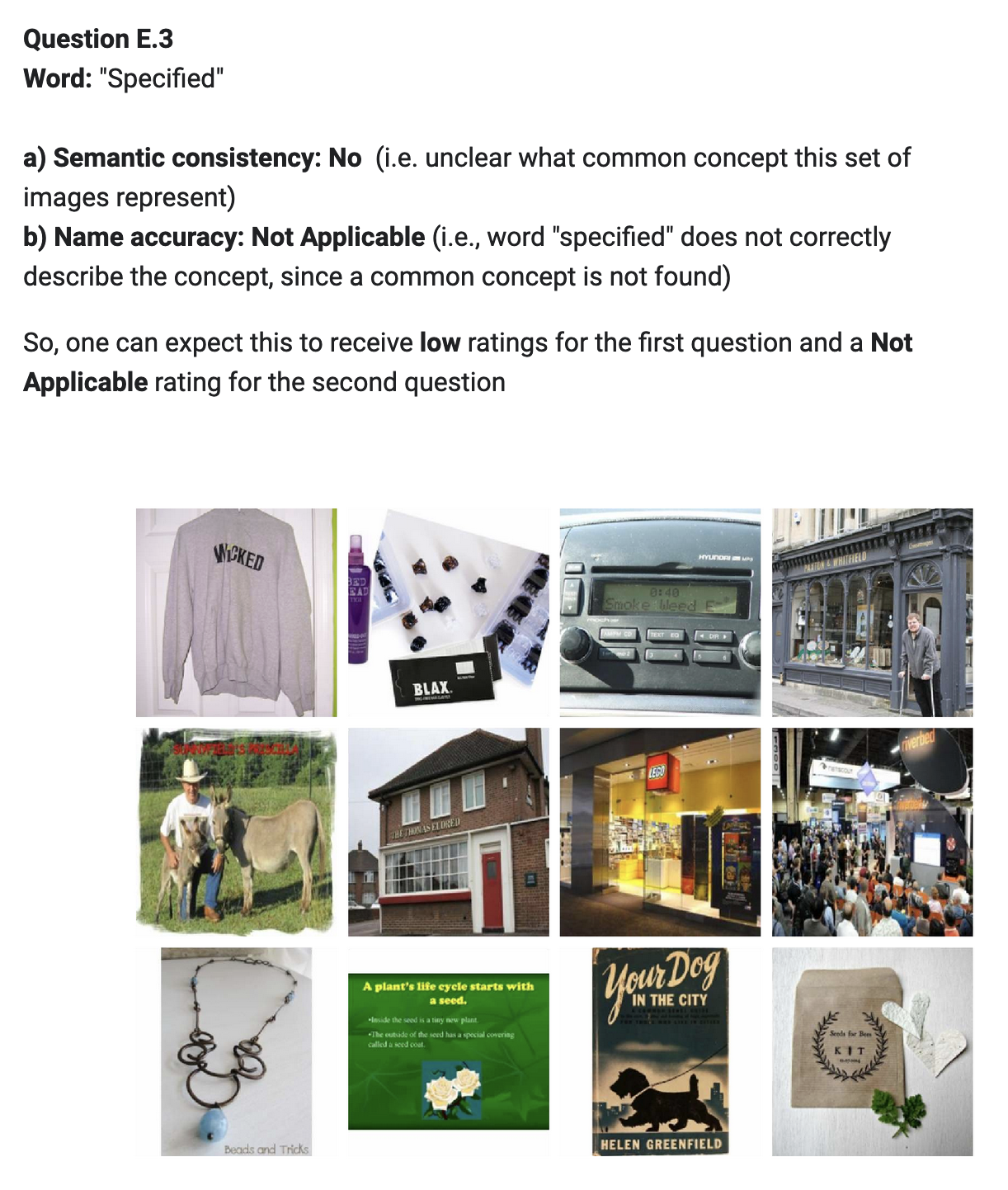}
    \end{subfigure}
    \caption{\textbf{Examples provided at the beginning of the user study.} Three examples (two shown here) were provided at the beginning of the survey to help partipants familiarize themselves with the task.
    }
    \label{fig:supp:survey_example}
    
\end{figure}

%% file: supplement/figures_tex/userstudy_example.tex
\begin{figure}[t!]
    \centering
    \begin{subfigure}[c]{.65\textwidth}
    \includegraphics[width=\linewidth]{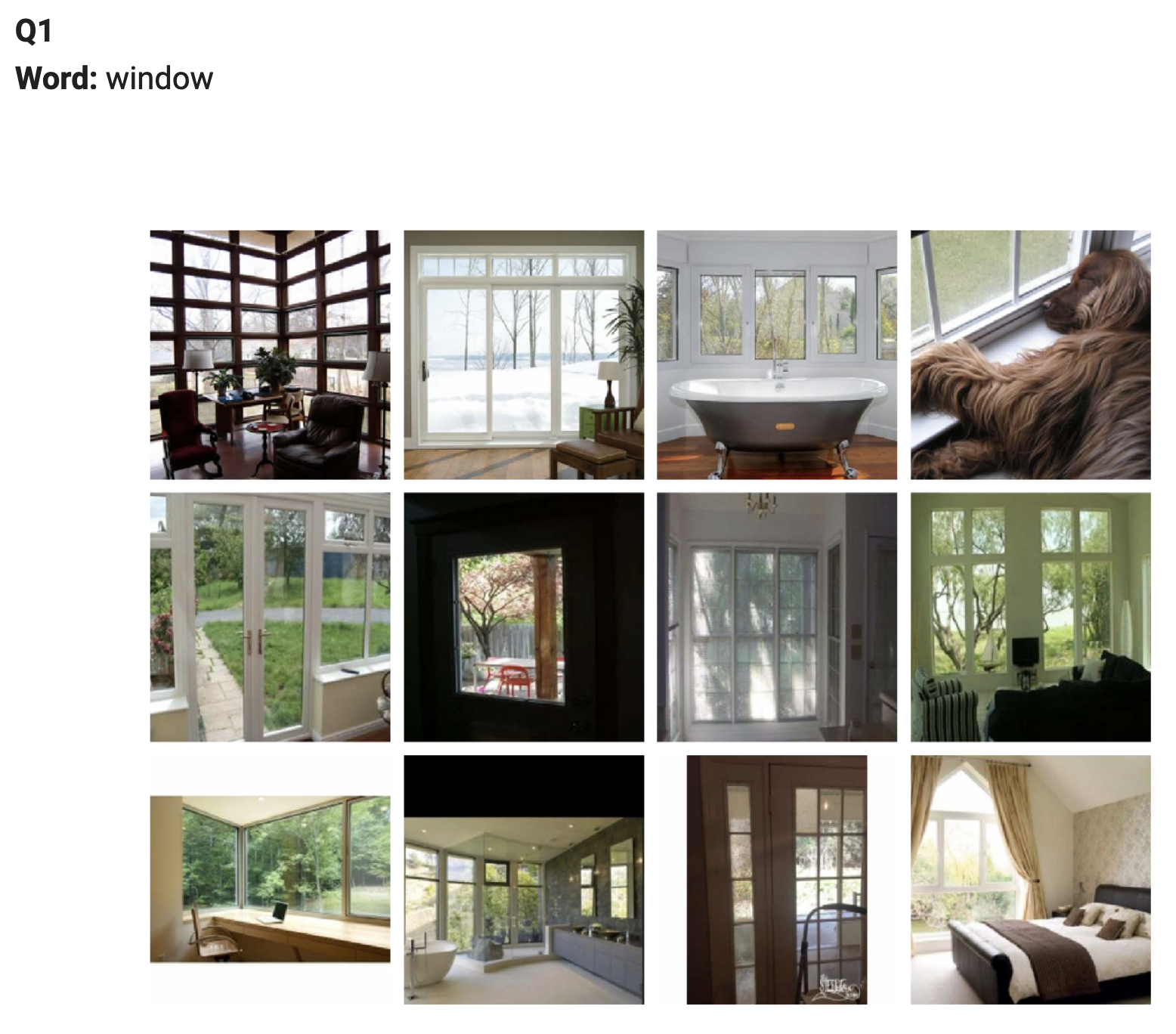}
    \end{subfigure}
    \begin{subfigure}[c]{.65\textwidth}
    \includegraphics[width=\linewidth]{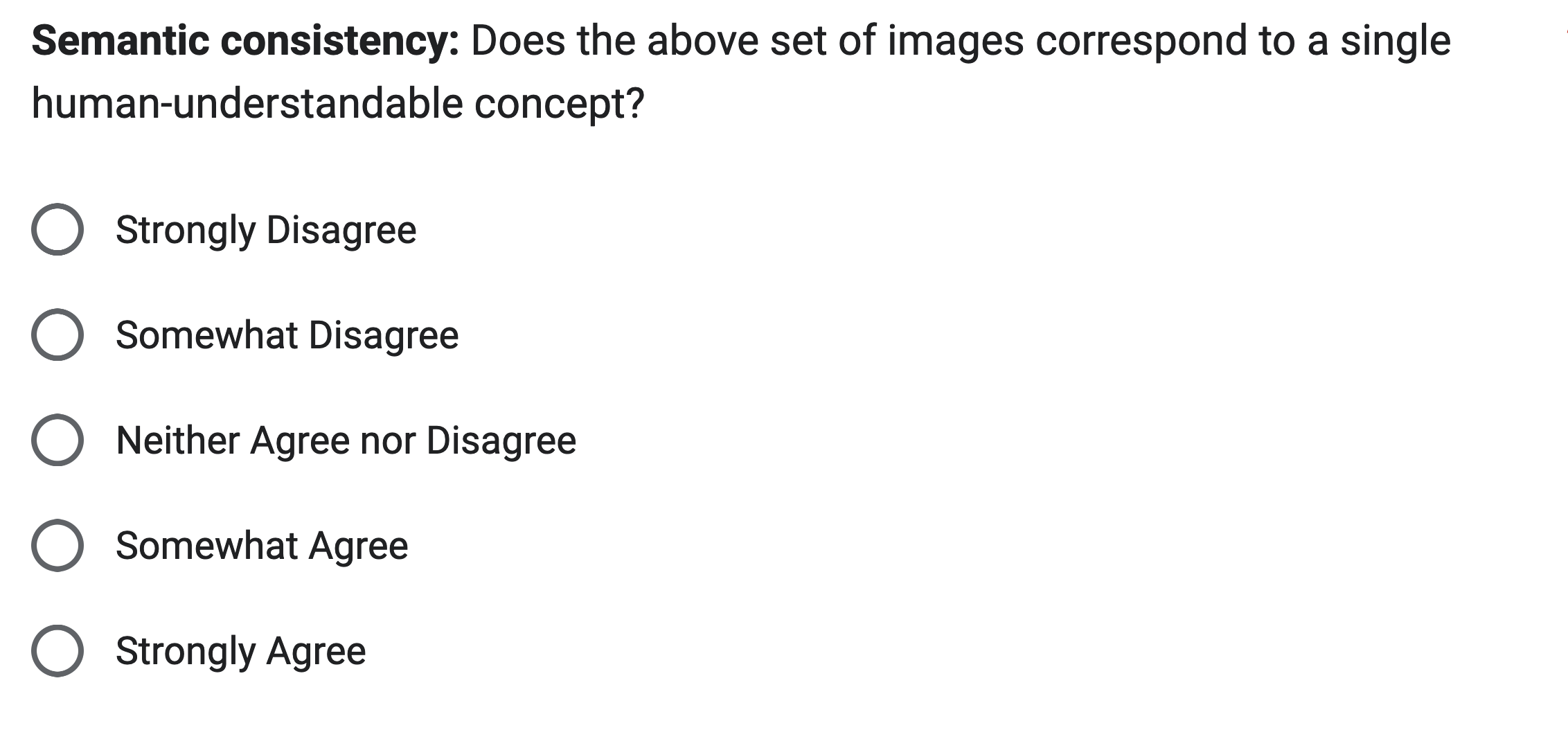}
    \end{subfigure}
    \begin{subfigure}[c]{.65\textwidth}
    \includegraphics[width=\linewidth]{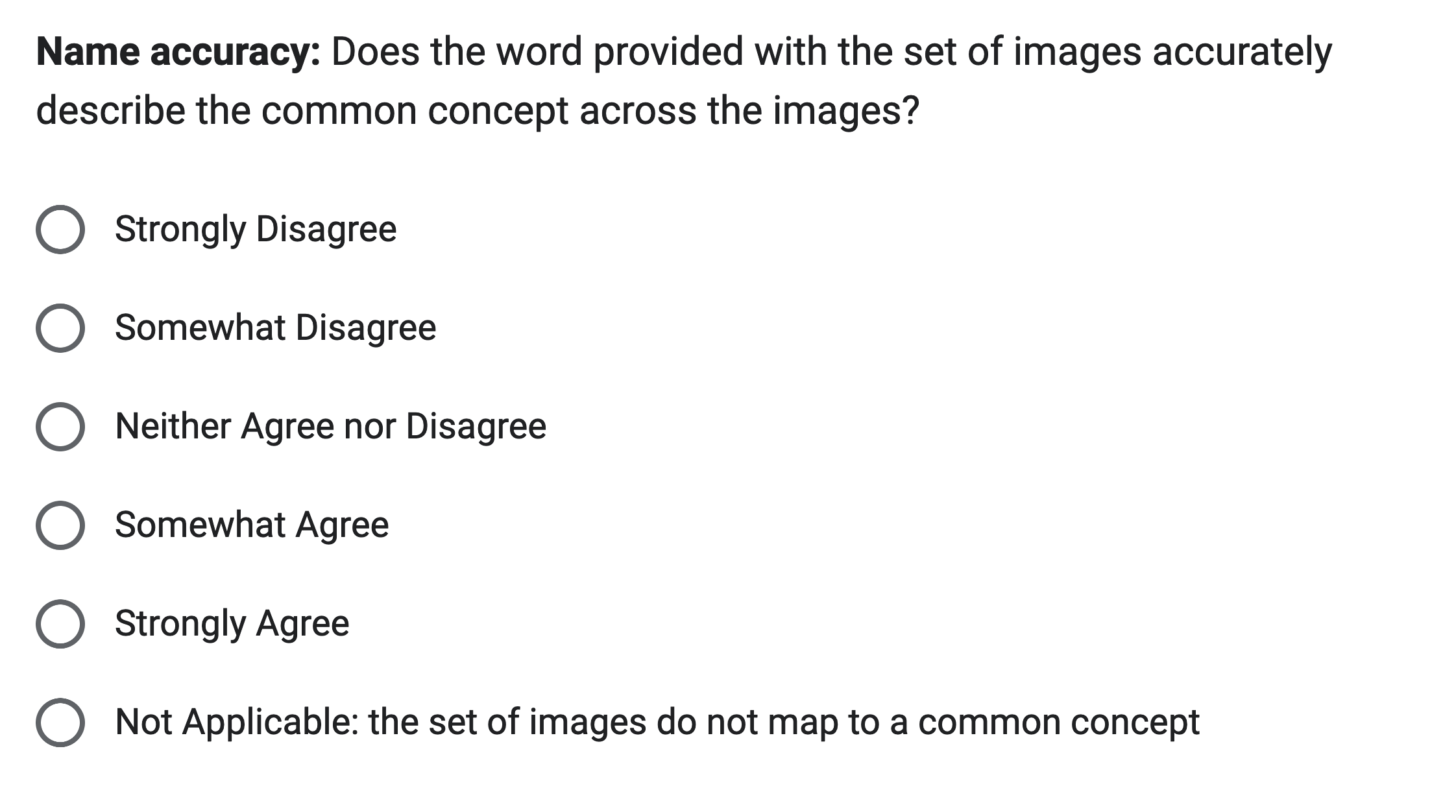}
    \end{subfigure}
    \caption{\textbf{An example of a question in the user study.} For each node, we provide top twelve activating images across three datasets and the name assigned to the node, and ask participants to rate for semantic consistency and name accuracy.
    }
    \label{fig:supp:userstudy_example}
    
\end{figure}

%% file: supplement/tables/table_concepts_waterbirds.tex
\begin{table}
    \centering
    \caption{\textbf{Set of concepts in our evaluation for performing interventions on DN-CBMs.} For each class, we manually classify each of the five concepts as being a bird or non-bird concept, for applying appropriate interventions.}
    \begin{tabular}{l@{\hskip8pt}c@{\hskip8pt}c@{\hskip8pt}}\toprule
    \bf Class & \bf Bird Concepts & \bf Non-Bird Concepts \\
    \midrule
    Landbird & sparrow, parrot, crow & forest, clic \\
    Waterbird & gull, ducks & landing, beach, canoeing \\
    \bottomrule
    \end{tabular}
\label{tab:supp:concepts_waterbirds}
\end{table}

%% file: supplement/2_quantitative.tex
\section{Additional Quantitative Results}
\label{supp:sec:quantitative}

In this section, we provide additional quantitative results. In \cref{supp:sec:quantitative:classification} we provide results for the performance of DN-CBM on additional backbones. In \cref{supp:sec:quantitative:sparsity}, we discuss the tradeoff between accuracy and sparsity of DN-CBM explanations.

\subsection{DN-CBM Classification Performance}
\label{supp:sec:quantitative:classification}

In this section, we provide full quantitative results across the three backbones, \ie \clip\citeS{radford2021learningS} \resnetf\citeS{he2016deepS}, \clip\vitb\citeS{dosovitskiy2021anS}, and \clip\vitl\citeS{dosovitskiy2021anS}. The classification accuracies across the four datasets, \ie, \imagenet\citeS{deng2009imagenetS}, \places\citeS{zhou2017placesS}, \cifart\citeS{krizhevsky2009learningS}, and \cifarh\citeS{krizhevsky2009learningS}, for our \dncbm and the baseline methods for each of these backbones can be found in \cref{supp:tab:accuracy:resnet50,supp:tab:accuracy:vitb,supp:tab:accuracy:vitl}.

Broadly, we find that our \dncbm outperforms all the baselines and almost completely bridges the gap in accuracy with linear probes

For our \dncbm, in addition to the proposed task-agnostic setting, we also additionally report accuracies by using the best configuration across SAE configurations, based on classification performance on a held out validation set, and call this \dncbm$_T$. In other words, the performance reported under \dncbm$_T$ constitutes the setting when our concept discovery is not task-agnostic, \ie, when a separate SAE is selected for each dataset. We find, as expected, that \dncbm$_T$ slightly outperforms the task-agnostic \dncbm setting. However, interestingly, the performance difference is very small, showing that our task-agnostic approach, while being more general, also yields highly performant classifiers.

\input{supplement/tables/table_accuracy_taskagnostic}

\subsection{Sparsity of DN-CBM Explanations}
\label{supp:sec:quantitative:sparsity}

\input{supplement/figures_tex/accuracy_sparsity_tradeoff}
In addition to being accurate and general, one also desires that the explanations are \emph{sparse}, since explanations with a large number of concepts are not very human interpretable \citeS{ramaswamy2023overlookedS}.
We explore the relationship between sparsity and accuracy of \dncbm in \cref{fig:accspar} by evaluating models across SAE and classifier hyperparameters on both metrics. To measure sparsity, we count the average number of concepts required to reach 0.9 fraction of the original logit value for each image across the dataset and find, quite naturally, that there exists a trade-off between the two metrics across datasets. However, interestingly, we find that one can obtain models that require very few concepts per decision on average by sacrificing only a small amount of accuracy.

%% file: supplement/tables/table_accuracy_taskagnostic.tex
\begin{table}
    \centering
    \caption{\textbf{Classification accuracy ($\%$) on \clip \resnetf.} `*' indicates results reported for the respective baselines, and linear probe and 
    zero-shot performance is as reported by \protect\citeS{radford2021learningS}. `-' indicates configurations where results were not reported or which could not be rerun since concept sets were not generated by the baseline methods (\eg \labo with \places). \dncbm$_T$ refers to our method, where in we choose the SAE which achieves best validation accuracy depending on the downstream dataset.}
    \begin{tabular}{l@{\hskip8pt}c@{\hskip16pt}c@{\hskip16pt}c@{\hskip16pt}c@{\hskip16pt}}\toprule
     \multirow{2}{*}{\footnotesize
     Model} &
    \multicolumn{4}{c}{\bf \clip\resnetf\phantom{shift}}\\[.5em]
    & \imagenet & \places & \cifart & \cifarh \\
    \midrule
    Linear Probe & 73.3* & 53.4 & 88.7* & 70.3* \\
    Zero Shot & 59.6* & 38.7 & 75.6* & 41.6*  \\
    \midrule 
    \lfcbm\citeS{oikarinen2023labelS} & 67.5\phantom{*} & 49.0\phantom{*} & 86.4* & 65.1*  \\
    
    \labo\citeS{yang2023languageS} & 68.9\phantom{*} & - & 87.9* & 69.1*  \\
    
    \cdm\citeS{panousis2023sparseS} & 72.2* & 52.7* & 86.5* & 67.6*  \\
    
    \dclip\citeS{menon2022visualS} & 59.6\phantom{*} & 37.9\phantom{*} & - & -  \\
    
    \midrule
    \dncbm (Ours) & 72.9\phantom{*} & 53.5\phantom{*} & 87.6\phantom{*}  & 67.5\phantom{*} \\
    \dncbm$_T$ (Ours) & \textbf{73.2}\phantom{*} & \textbf{53.9}\phantom{*} & \textbf{88.6}\phantom{*} & \textbf{69.2}\phantom{*} \\
    \midrule
    \end{tabular}
\label{supp:tab:accuracy:resnet50}
\end{table}

\begin{table}
    \centering
    \caption{\textbf{Classification accuracy ($\%$) on \clip \vitb.} `*' indicates results reported for the respective baselines, and linear probe and zero-shot performance is as reported by \protect\citeS{radford2021learningS}. `-' indicates configurations where results were not reported or which could not be rerun since concept sets were not generated by the baseline methods (\eg \labo with \places). \dncbm$_T$ refers to our method, where in we choose the SAE which achieves best validation accuracy depending on the downstream dataset.}
    \begin{tabular}{l@{\hskip8pt}c@{\hskip16pt}c@{\hskip16pt}c@{\hskip16pt}c@{\hskip16pt}}\toprule
     \multirow{2}{*}{\footnotesize
     Model} &
    \multicolumn{4}{c}{\bf \clip\vitb\phantom{shift}}\\[.5em]
    & \imagenet & \places & \cifart & \cifarh \\
    \midrule
    Linear Probe & 80.2* & 55.1 & 96.2* & 83.1* \\
    Zero Shot & 68.6* & 41.2 & 91.6* & 68.7* \\
    \midrule 
    \lfcbm\citeS{oikarinen2023labelS} & 75.4\phantom{*} & 50.6\phantom{*} & 94.6\phantom{*} &  77.4\phantom{*} \\
    
    \labo\citeS{yang2023languageS} & 78.9\phantom{*} & - & 95.7\phantom{*} & 81.2\phantom{*} \\
    
    \cdm\citeS{panousis2023sparseS} & 79.3* & 52.6* & 95.3* & 80.5* \\
    
    \dclip\citeS{menon2022visualS} & 68.0* & 40.3* & - & -  \\
    
    \midrule
    \dncbm (Ours) &  \textbf{79.5}\phantom{*} & \textbf{55.1}\phantom{*} & \textbf{96.0}\phantom{*} & \textbf{82.1}\phantom{*} \\
    \dncbm$_T$ (Ours) & 79.5\phantom{*} & 55.1\phantom{*} & 95.7\phantom{*} & 82.1\phantom{*} \\
    \midrule
    \end{tabular}
\label{supp:tab:accuracy:vitb}
\end{table}

\begin{table}
    \centering
    \caption{\textbf{Classification accuracy ($\%$) on \clip \vitl.} `*' indicates results reported for the respective baselines. \lfcbm is not reported here as they take \vitb as the teacher model and hence we do not use it with a \vitl student. Linear probe and 
    zero-shot performance is as reported by \protect\citeS{radford2021learningS}. `-' indicates configurations where results were not reported or which could not be rerun since concept sets were not generated by the baseline methods (\eg \labo with \places). \dncbm$_T$ refers to our method, where in we choose the SAE which achieves best validation accuracy depending on the downstream dataset.}
    \begin{tabular}{l@{\hskip8pt}c@{\hskip16pt}c@{\hskip16pt}c@{\hskip16pt}c@{\hskip16pt}}\toprule
     \multirow{2}{*}{\footnotesize
     Model} &
    \multicolumn{4}{c}{\bf \clip\vitl\phantom{shift}}\\[.5em]
    & \imagenet & \places & \cifart & \cifarh \\
    \midrule
    Linear Probe & 83.9* & 55.6 & 98.0* & 87.5* \\
    Zero Shot & 75.3* & 41.4 & 96.2* & 77.9* \\
    \midrule 
    
    \labo\citeS{yang2023languageS} & \textbf{84.0}* & - & 97.8* & 86.0* \\
    
    \cdm\citeS{panousis2023sparseS} & 83.4\phantom{*} & 55.2\phantom{*} & 98.0\phantom{*} & 86.4\phantom{*} \\
    
    \dclip\citeS{menon2022visualS} & 75.0* & 40.6* & - & -  \\
    
    \midrule
    \dncbm (Ours) &  83.6\phantom{*} & \textbf{55.6}\phantom{*} & \textbf{98.1}\phantom{*} & 86.0\phantom{*} \\
    \dncbm$_T$ (Ours) & 83.6\phantom{*} & 55.6\phantom{*} & 97.9\phantom{*} & \textbf{87.4}\phantom{*} \\
    \midrule
    \end{tabular}
\label{supp:tab:accuracy:vitl}
\end{table}

%% file: supplement/figures_tex/accuracy_sparsity_tradeoff.tex
\begin{figure}[t!]
\centering
    \includegraphics[width=0.65\linewidth]{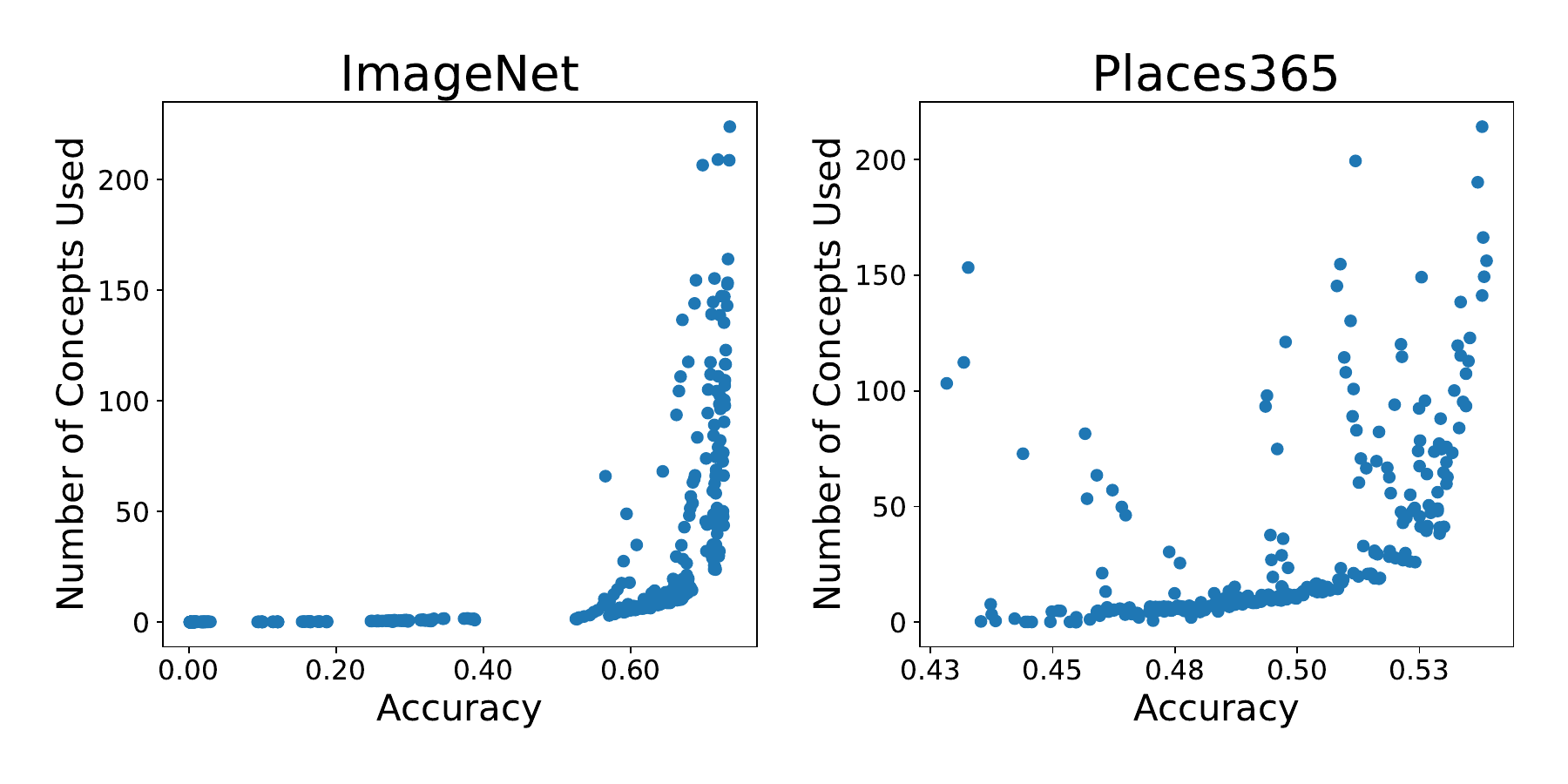}
    \caption{\textbf{Accuracy vs. Sparsity Trade-off.} For each dataset, we show the model accuracy (x-axis) versus the average number of concepts needed to reach 0.9 of the total logit value (y-axis). We find that while there exists a trade-off between accuracy and sparsity, highly sparse decisions can be achieved by a very small drop in accuracy. 
    }
    \label{fig:accspar}
\end{figure}

%% file: supplement/1_qualitative.tex
\section{Additional Qualitative Results}
\label{supp:sec:qualitative}

In this section, we provide additional qualitative results. In \cref{supp:sec:qualitative:taskagnosticity}, we provide additional qualitative examples of task agnosticity of our discovered concepts. In \cref{supp:sec:qualitative:clustering}, we provide additional examples of clustering concept strength vectors. In \cref{supp:sec:qualitative:local} and \cref{supp:sec:qualitative:global}, we provide examples of local and global explanations from our \dncbm and discuss our findings.

\subsection{Task Agnosticity of Concepts}
\label{supp:sec:qualitative:taskagnosticity}

In this section, we provide further qualitative evidence of task agnosticity of our discovered concepts by showing examples of discovered concepts and top activating images for each concept from \imagenet\citeS{deng2009imagenetS}, \places\citeS{zhou2017placesS}, \cifart\citeS{krizhevsky2009learningS} and \cifarh\citeS{krizhevsky2009learningS}, using sparse autoencoders trained on each of the three vision backbones, \ie \clip\citeS{radford2021learningS} \resnetf\citeS{he2016deepS} (\cref{fig:taskgnostic_rn50}), \clip\vitb\citeS{dosovitskiy2021anS} (\cref{fig:taskgnostic_vitb}), and \clip\vitl\cite{dosovitskiy2021anS} (\cref{fig:taskgnostic_vitl}).

Overall (\cref{fig:taskgnostic_rn50,fig:taskgnostic_vitb,fig:taskgnostic_vitl}), we find that across backbones, the discovered concepts are highly semantically consistent and map well to the name that is automatically assigned to them. The discovered concepts greatly vary in complexity, from simple concepts such as colours (\eg `maroon', \cref{fig:taskgnostic_vitb}) to complex concepts such as `conversation' (\cref{fig:taskgnostic_vitl}). The images activating on each concept are highly visually diverse, while still being semantically related to the concept. For a more detailed discussion, please refer to the captions of \cref{fig:taskgnostic_rn50,fig:taskgnostic_vitb,fig:taskgnostic_vitl}.

\input{supplement/figures_tex/task_agnosticity}

\subsection{Clustering Concept Strength Vectors}
\label{supp:sec:qualitative:clustering}
In this section, we provide additional qualitative evidence that our sparse autoencoder (SAE) is able to learn concepts
 which are semantically consistent, for which we cluster the concept strengths in the SAE latent space and find that the concept strengths are able segregate images well formed groups. Overall (\cref{fig:metaclusters_imgnet}), we find that across backbones, the discovered clusters contain images which are highly similar to each other visually. (\cref{fig:metaclusters_imgnet}) shows the clusters formed for the \imagenet dataset. For a more detailed discussion, please refer to the captions of (\cref{fig:metaclusters_imgnet}).
\input{supplement/figures_tex/meta_clusters.tex}
\clearpage

\subsection{Local Explanations from \dncbm}
\label{supp:sec:qualitative:local}
In this section, we provide additional examples of local explanations of our \dncbm on the \places and \imagenet datasets, using all three vision backbones (\clip\resnetf, \clip\vitb, and \clip\vitl). In \cref{fig:localexp_places_misclas}, we show examples of \places images misclassified by the model, and analyze and discuss the misclassifications based on the provided explanations. In \cref{fig:localexp_places_correct}, we provide additional examples of correct classifications on \places. Finally, in \cref{fig:localexp_imagenet}, we provide examples of local explanations from \imagenet, including a misclassified example.

Overall, we find that the provided explanations typically describe the input image well, are highly diverse, and can even help better understand misclassified decisions by the model. For full details, please refer to the captions of \cref{fig:localexp_places_misclas,fig:localexp_places_correct,fig:localexp_imagenet}.

\input{supplement/figures_tex/local_explanations.tex}

\subsection{Global Explanations from \dncbm}
\label{supp:sec:qualitative:global}
In this section, we provide additional qualitative examples of global explanations on the \places dataset from our \dncbm. This figure contains explanations as concept names which contribute the most to the class. Our method is able to explain the class with concepts which are highly relevant (\cref{fig:globalexp_supp}) to it. Overall, we find that across backbones, the classes are explained well with the top-contributing concepts. For detailed discussion, please refer to the captions of (\cref{fig:globalexp_supp}).

\input{supplement/figures_tex/global_explanations}

%% file: supplement/figures_tex/task_agnosticity.tex
\begin{figure}
    \centering
    \begin{subfigure}[c]{\textwidth}
    \includegraphics[width=\linewidth]{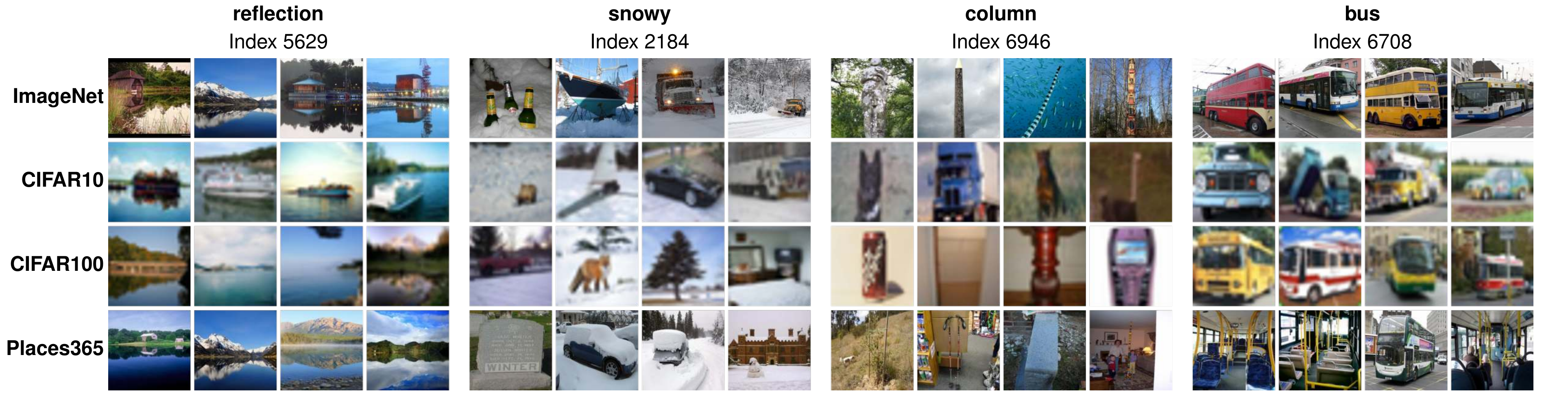}
    \end{subfigure}\\
    \begin{subfigure}[c]{\textwidth}
    \includegraphics[width=\linewidth]{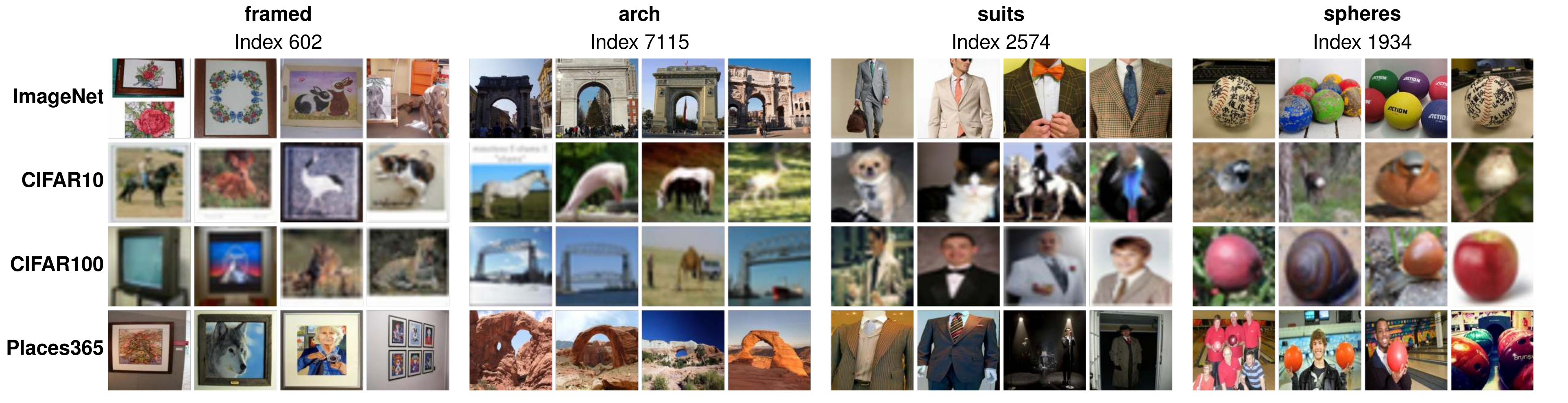}
    \end{subfigure}
    \caption{\textbf{Task-agnosticity of concept extraction using \clip \resnetf.} We show examples of named concepts (blocks) and top images activating them from four datasets (rows).
    \\\textbf{(1)} We find overall that the images activating the concept are highly consistent with the concept name across datasets and are from diverse scenarios. For example, the concept `column' (top row, middle right) is activated by images from a diverse set of classes, while visually sharing the common feature of being a long and narrow object, including an underwater animal (row 1, col 3 in the block), an animal head (row 2, col 4), a rocket (row 1, col 2), and an elongated version of a truck (row 2, col 2). \\\textbf{(2)} Images activating a concept are highly visually diverse. For example, the concept `bus' (top right) contains images depicting both exterior as well as interior views of buses. Similarly, the concept `arch' (bottom row, middle left), is activated by images from different classes such as animals, gates, bridges, rocks, all similar in the aspect that their shape resembles an arch; and the concept `spheres' (bottom right) is activated by images of apples, snails, as well as the birds which look round in shape.
    The concept `suits' (bottom row, middle right), interestingly and aptly, is also activated by an image of a dog wearing a tie.
    }
    \label{fig:taskgnostic_rn50}
\end{figure}

\begin{figure}
    \centering
    \begin{subfigure}[c]{\textwidth}
    \includegraphics[width=\linewidth]{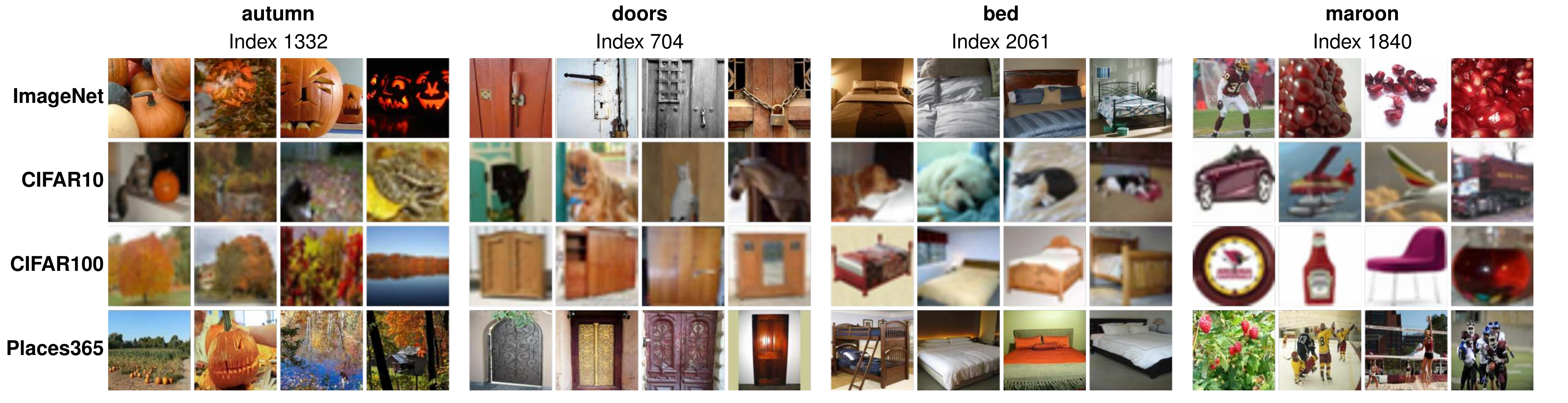}
    \end{subfigure}\\
    \begin{subfigure}[c]{\textwidth}
    \includegraphics[width=\linewidth]{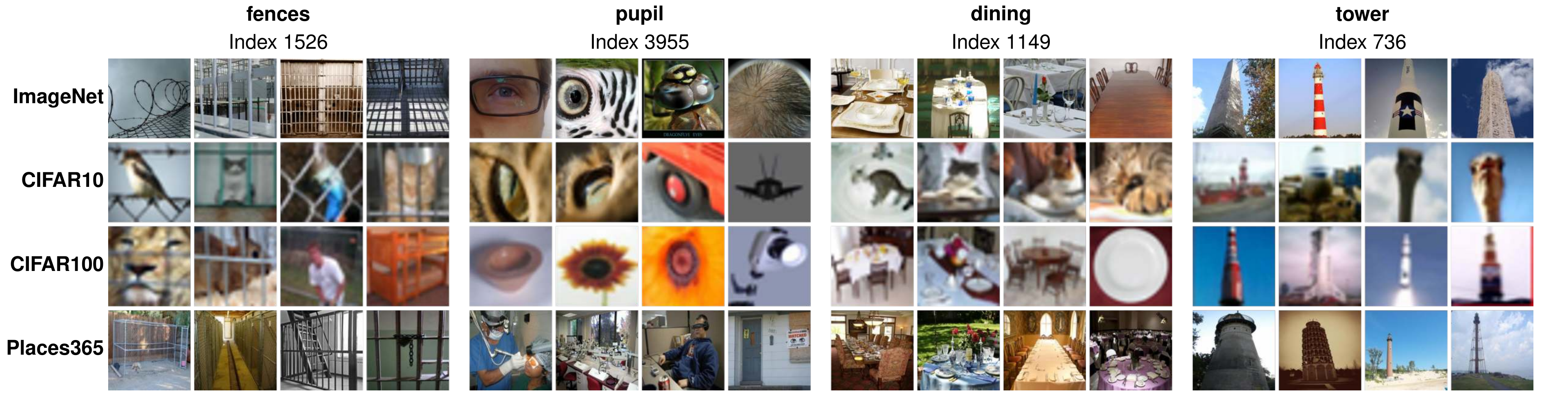}
    \end{subfigure}
    \caption{\textbf{Task-agnosticity of concept extraction using \clip \vitb.} We show examples of named concepts (blocks) and top images activating them from four datasets (rows). \\\textbf{(1)} We find overall that the images activating the concept are highly consistent with the concept name across datasets and are from diverse scenarios. For example, the concept `autumn' (top left) includes images of trees with autumn colours, fallen leaves, and pumpkins which are associated with autumn and Halloween. \\\textbf{(2)} Concepts range from simple concepts (\eg colours, such as `marooon' in the top right) to complex ones (\eg `dining', bottom row, middle right). \\\textbf{(3)} Images activating a concept are highly visually diverse. For example, the concepts `door' (top row, middle left) and `dining' (bottom row, middle right) include images at of doors and dining tables respectively at different scales, view points, of diverse styles and colours, and in diverse scenarios (\eg indoor as well as outdoor). Similarly, the concept `tower' (bottom right) includes a diverse set of images that look like towers, such as lighthouses, rockets, skyscrapers, and water towers; and the concept `pupil' (bottom row, middle left) contains examples of eyes of a diverse set of species, including humans, mammals, and insects, and also associations, such as images of an ophthalmologist's office. \\\textbf{(4)} When images matching a particular concept are scarce in a dataset (\eg \cifart, which has only ten classes and likely a far smaller set of visible concepts), the top concepts still appear to be good approximations of and are reasonably semantically aligned to the named concept. For example, the concept `tower' (bottom right) includes examples of front views of an ostrich head and an airplane, both of which visually resemble towers. Similarly, the concept `pupil' (bottom row, middle left) includes examples of sunflowers, the top view of a flower pot, and a wheel, which have a similar shape and appearance as eye pupils. This provides further evidence that the discovered concepts activate for semantically meaningful and similar visual features.
    }
    \label{fig:taskgnostic_vitb}
\end{figure}

\begin{figure}
    \centering
    \begin{subfigure}[c]{\textwidth}
    \includegraphics[width=\linewidth]{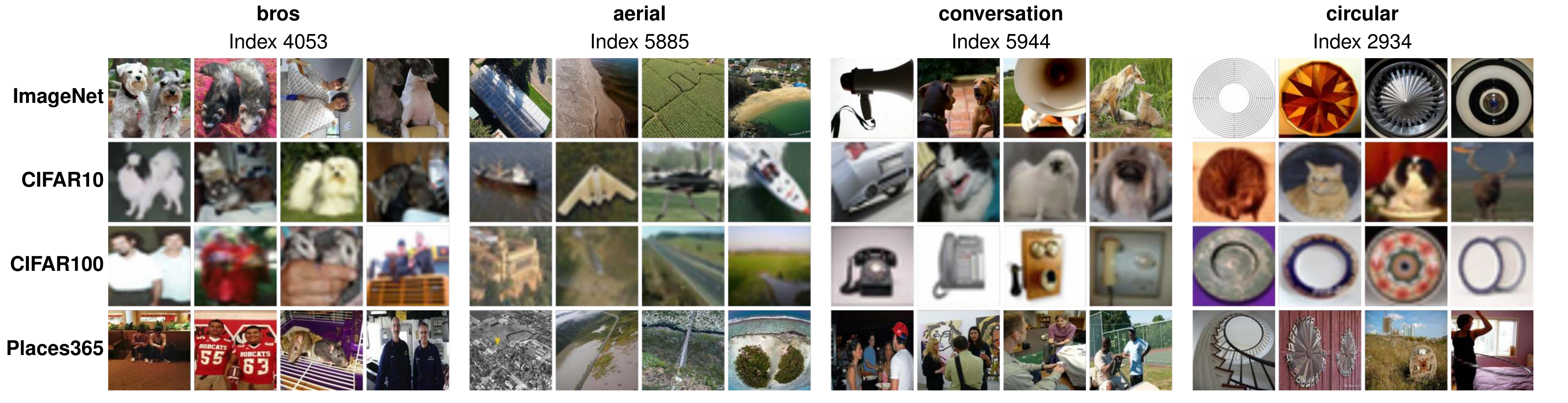}
    \end{subfigure}\\
    \begin{subfigure}[c]{\textwidth}
    \includegraphics[width=\linewidth]{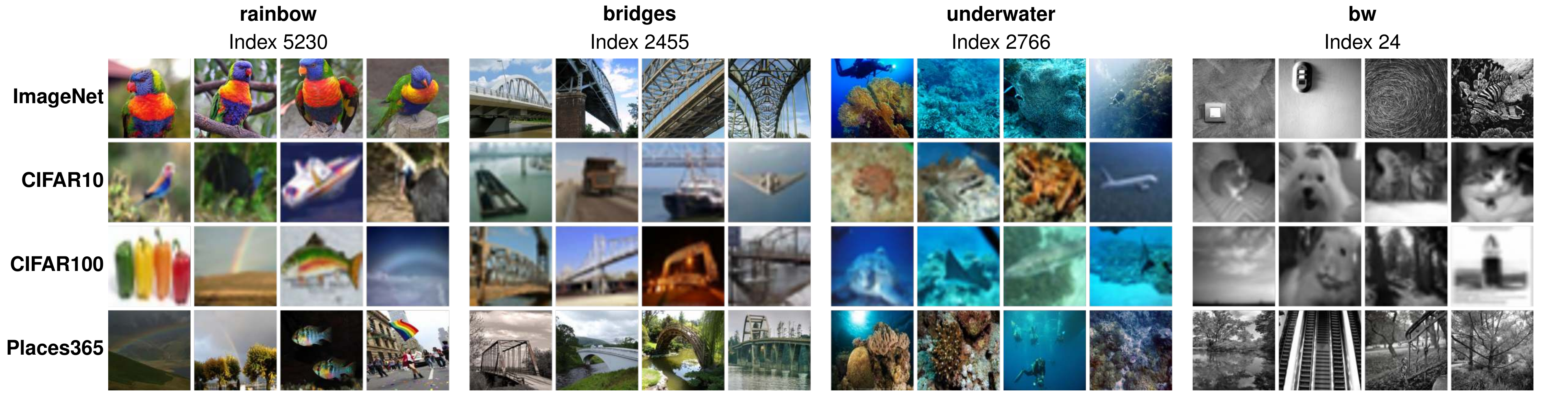}
    \end{subfigure}
    \caption{\textbf{Task-agnosticity of concept extraction using \clip \vitl.} We show examples of named concepts (blocks) and top images activating them from four datasets (rows). We find overall that the images activating the concept are highly consistent with the concept name across datasets and are from diverse scenarios. \\\textbf{(1)} The concept `bros' (top left) is activated by the presence of two entities in the image, from humans to to dogs, rats, and cats. \\\textbf{(2)} The concept `rainbow' (bottom left) is activated by highly colourful birds, and an image of bell peppers which are of different colors that are present in a rainbow. It is also activated by a rainbow-coloured flag, by bio-luminescent fish, and drawings of fish. \\\textbf{(3)} Concepts representing viewpoints such as `aerial' (top row, middle left) is also discovered by the model, which shows farming lands, highways, water bodies, and landmass. \\\textbf{(4)} The concept `conversation' (top row, middle right), is activated by people present in a crowd facing each other, which are typical images of people engaged in conversations. It also is activated by phones and megaphones which are tools for having conversations, and objects which look visually similar when good matches may not be present in the dataset, such as an image of an exhaust pipe from \cifart.
    }
    \label{fig:taskgnostic_vitl}
\end{figure}

%% file: supplement/figures_tex/meta_clusters.tex
\begin{figure}
    \centering
    \begin{subfigure}[c]{.95\textwidth}
    \includegraphics[trim=0 0 0 1.25cm, clip, width=\linewidth]{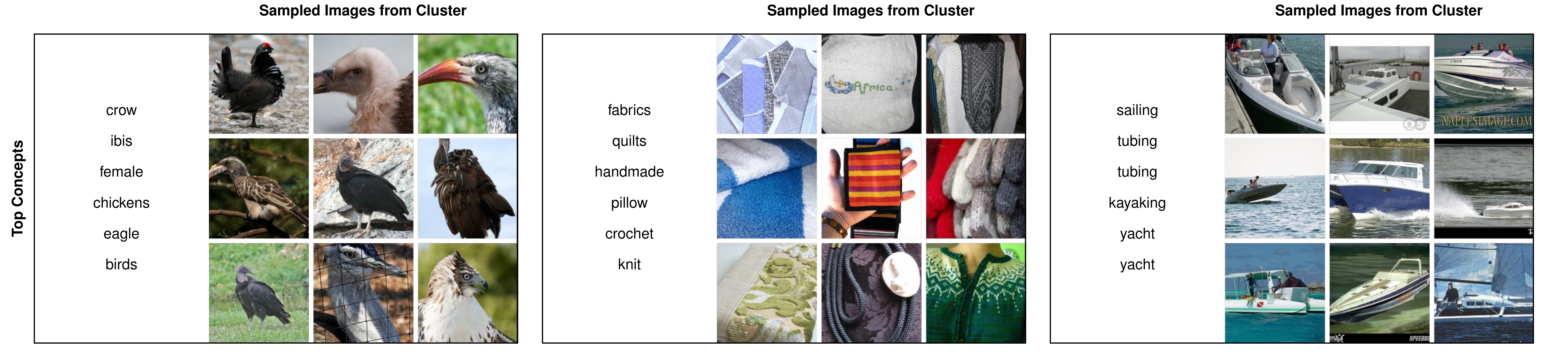}
    \caption{ \clip \resnetf}
    \end{subfigure}    
    \begin{subfigure}[c]{.95\textwidth}
    \includegraphics[width=\linewidth]{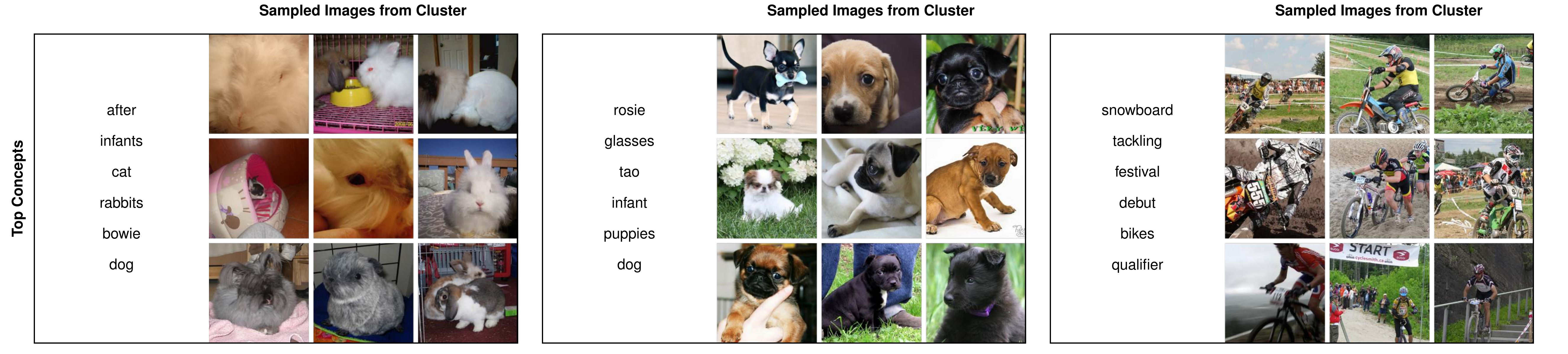}
    \caption{\clip \vitb}
    \end{subfigure}
    \begin{subfigure}[c]{.95\textwidth}
    \includegraphics[width=\linewidth]{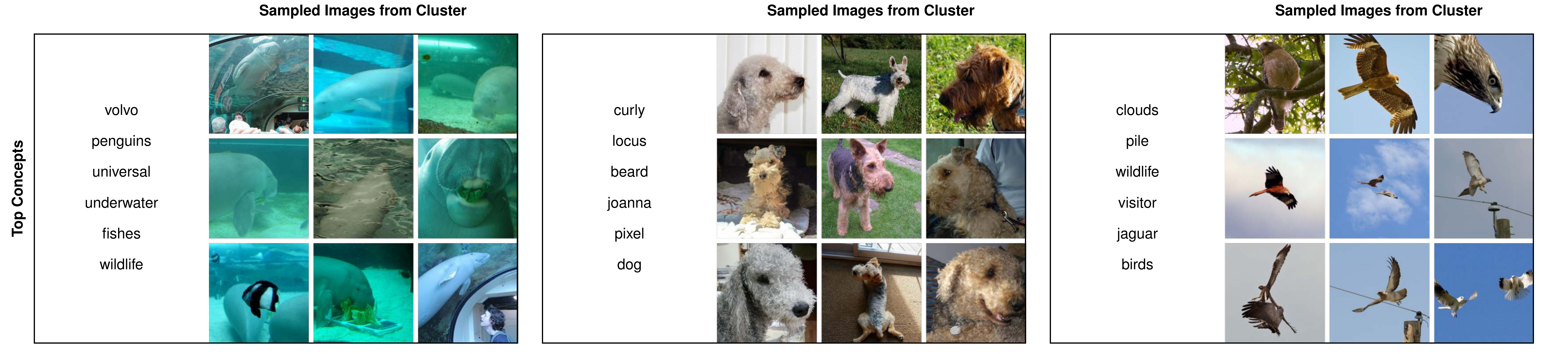}
    \caption{\clip \vitl}
    \end{subfigure}
    \caption{\textbf{Extracting meaningful clusters from concept strength vectors on the \imagenet dataset.} We perform K-Means clustering over concept activation vectors on the \imagenet dataset to evaluate the semantic consistency of these latent representations. We show a random subset of clusters: each block represents a cluster, and we show top concepts from the cluster centroid and randomly selected images assigned to the cluster. We find that highly semantically consistent clusters of concepts emerge.
    \\\textbf{(a)} The top row, shows clusters which are well segregated into different semantics, the first cluster shows close-up pictures of eagles, the second one shows knitted garments and the third cluster represents ships sailing in the ocean. 
    \\\textbf{(b)} Similarly, in the second row, the first cluster represents infant rabbits, the second cluster corresponds to infant dogs and the third one represents a person biking. 
    \\\textbf{(c)} The third row represents clusters corresponding to sea animal  underwater, dogs with a specific texture, and birds flying in the sky respectively. 
    }
    \label{fig:metaclusters_imgnet}
\end{figure}

%% file: supplement/figures_tex/local_explanations.tex
\begin{figure}[t]
    \centering
    \begin{subfigure}[c]{.45\textwidth}    \includegraphics[width=\linewidth]{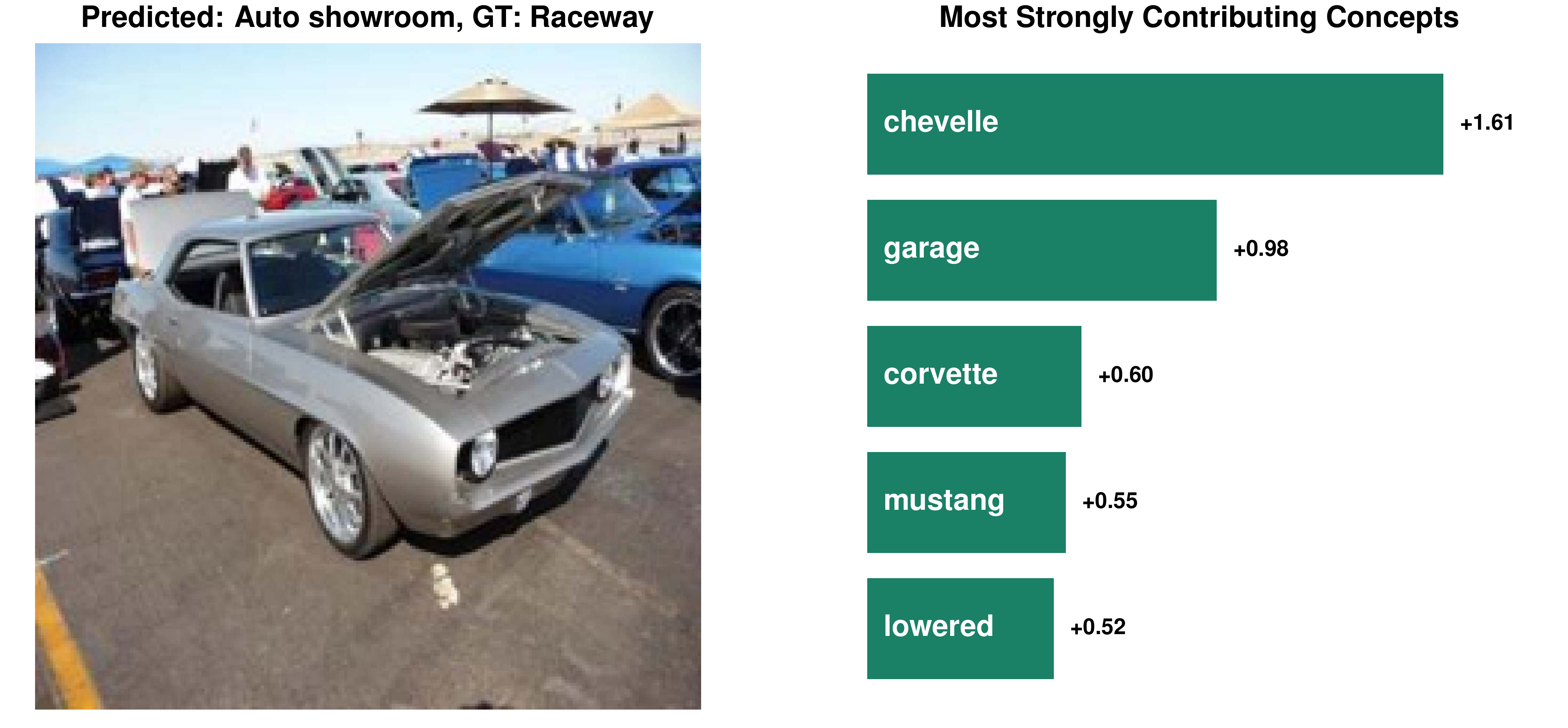}
    \caption{}
    \end{subfigure}\hfill
    \begin{subfigure}[c]{.45\textwidth}    \includegraphics[width=\linewidth]{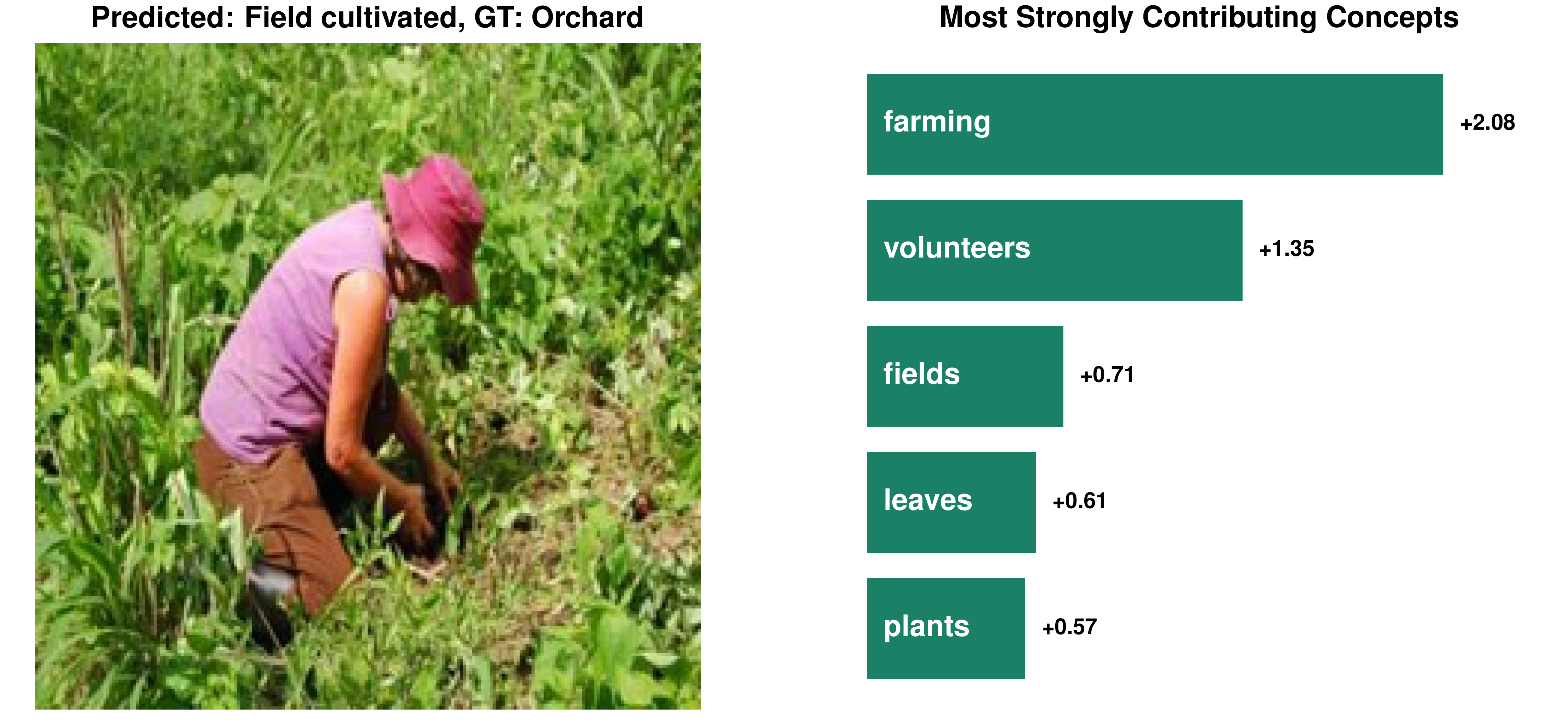}
    \caption{}
    \end{subfigure}\\
    \begin{subfigure}[c]{.45\textwidth}    \includegraphics[width=\linewidth]{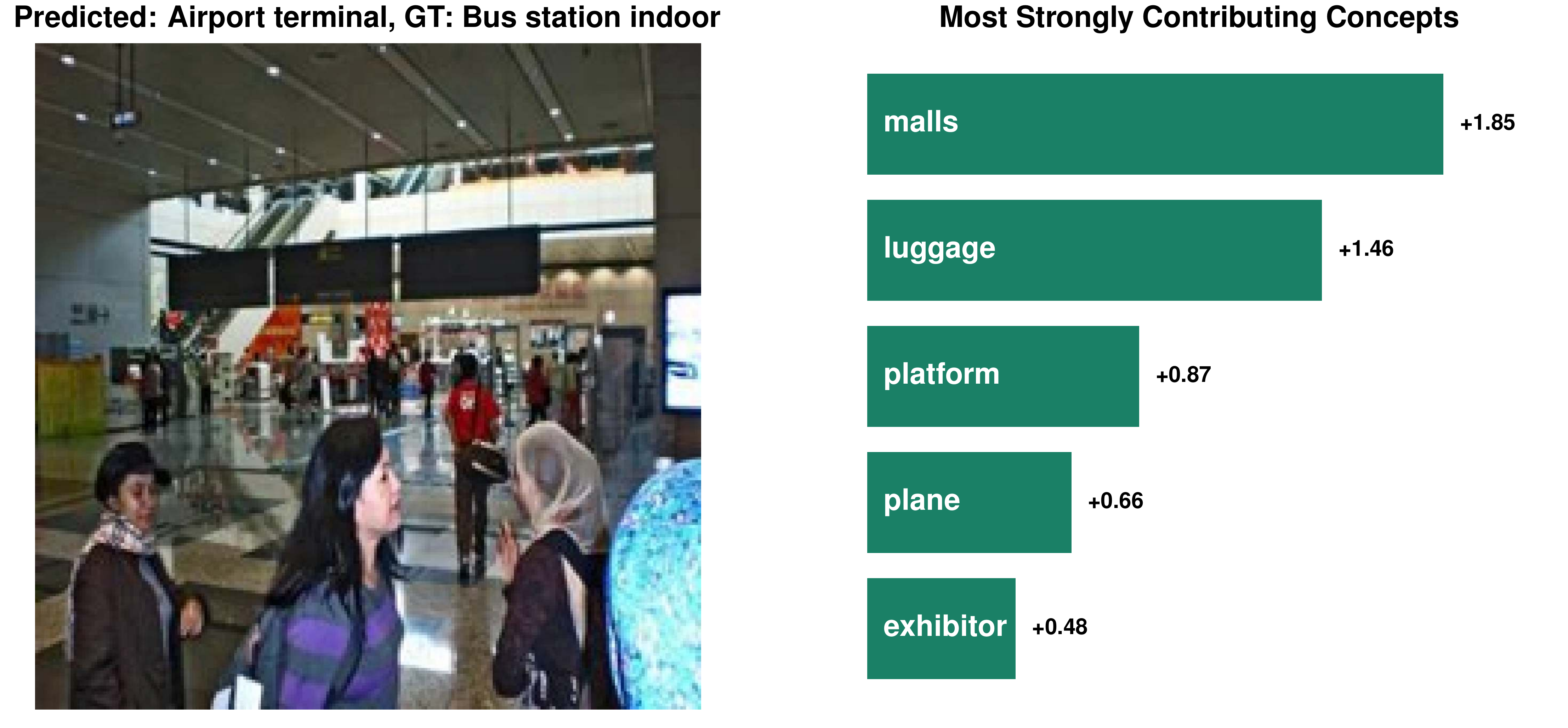}
    \caption{}
    \end{subfigure}\hfill
    \begin{subfigure}[c]{.45\textwidth}    \includegraphics[width=\linewidth]{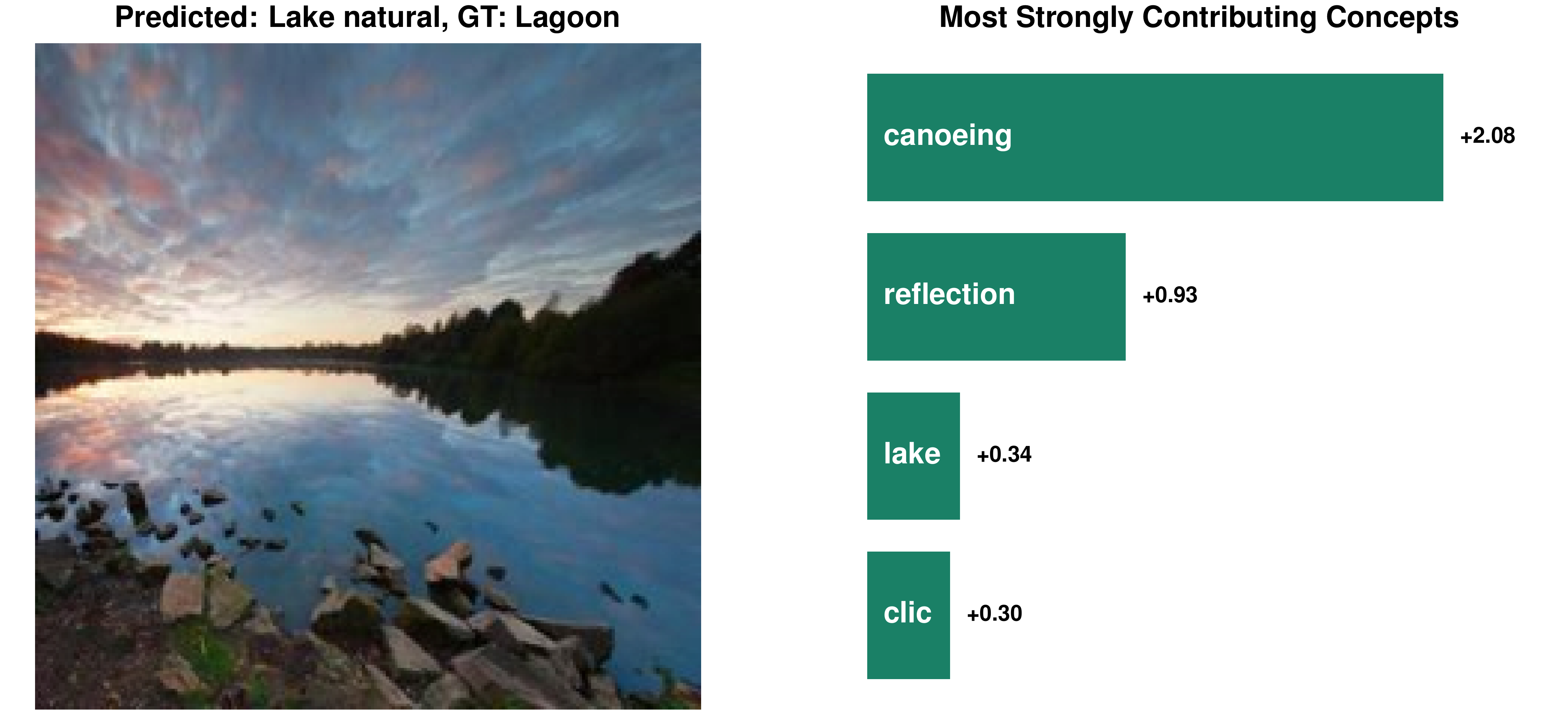}
    \caption{}
    \end{subfigure}\\
    \caption{\textbf{Examples of misclassifications on \places by \dncbm.} We show examples of misclassified images from the \places dataset using a \clip\resnetf backbone along with the top concepts contributing to the decision. \\\textbf{(a)} An image of class `Raceway' is misclassified as `Auto showroom', classes which both likely share similar car related concepts. The top five concepts in the decision include three car model names (`chevelle', `corvette', and `mustang') which might be correlated to cars found in images of auto showrooms. The image also prominently shows a car with an open hood, likely causing the concepts of `garage' and `lowered' to be activated and causing the misclassification. \\\textbf{(b)} An image of class `Orchard' is misclassified as `Field cultivated'. In this example, the image does appear to be visually more similar to the predicted class, with the top concepts for the decision all being visually present in the image. \\\textbf{(c)} An image of class `Bus station indoor' is misclassified as `Airport terminal'. Again, this example visually appears highly similar to the predicted class. However, the top concepts also include those that are not present in the image (\eg `plane'), suggesting that concepts may also be spuriously correlated with one another. This might even be a consequence of \clip having learnt such correlations, and investigating this further and finding ways to mitigate them would be an interesting direction for future research. \\\textbf{(d)} An image of class `Lagoon' misclassified as `Lake natural'. This image also visually appears better aligned with the predicted class. It also serves as another example where concepts appear to be detected due to spurious correlations---the top detected concept is `canoeing', despite it not being present in the image. It is probable that instances of `canoeing' are typically found in images of lakes, leading to this correlation being learnt by \clip or our \dncbm.
    }
    \label{fig:localexp_places_misclas}
\end{figure}

\begin{figure}[t]
    \centering
    \begin{subfigure}[c]{.45\textwidth}    \includegraphics[width=\linewidth]{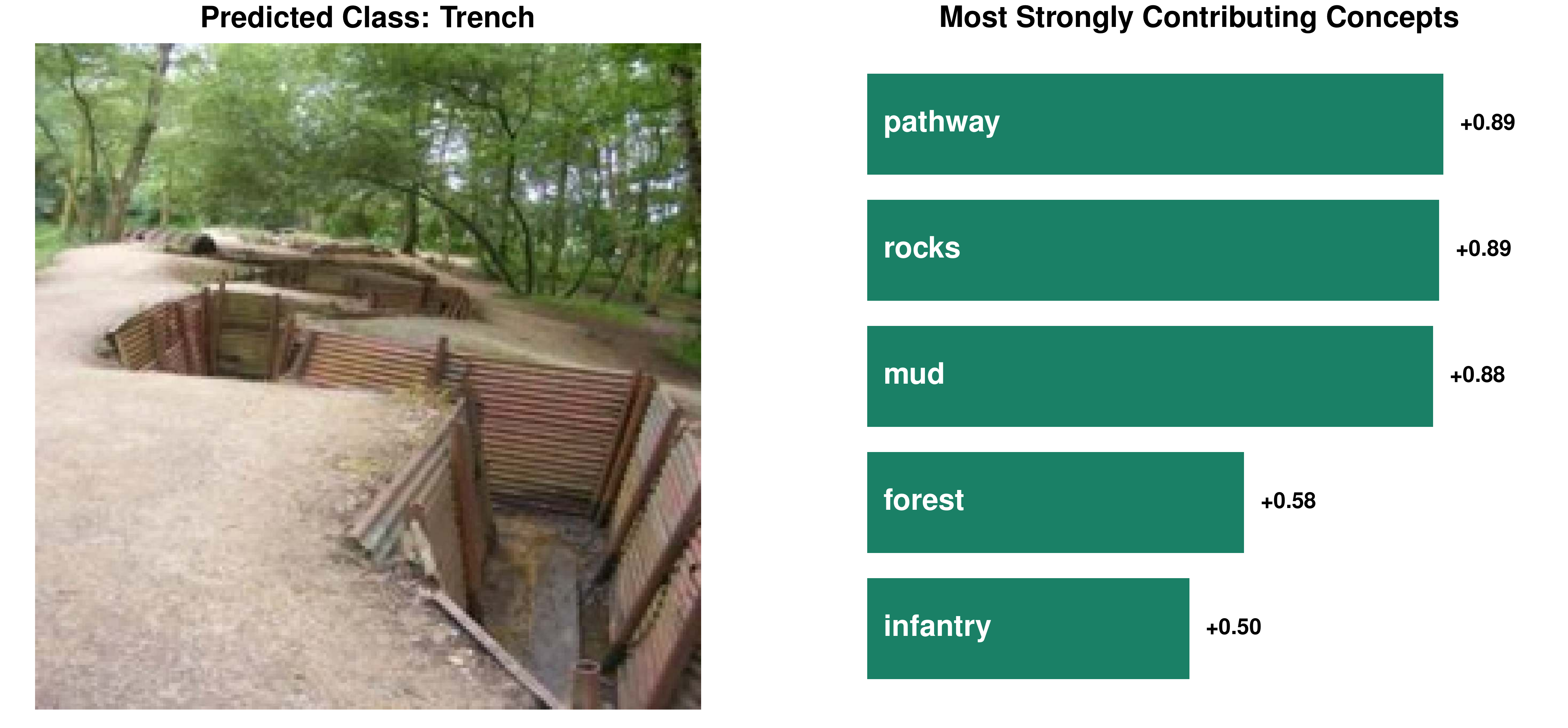}
    \caption{}
    \end{subfigure}\hfill
   \begin{subfigure}[c]{.45\textwidth}    \includegraphics[width=\linewidth]{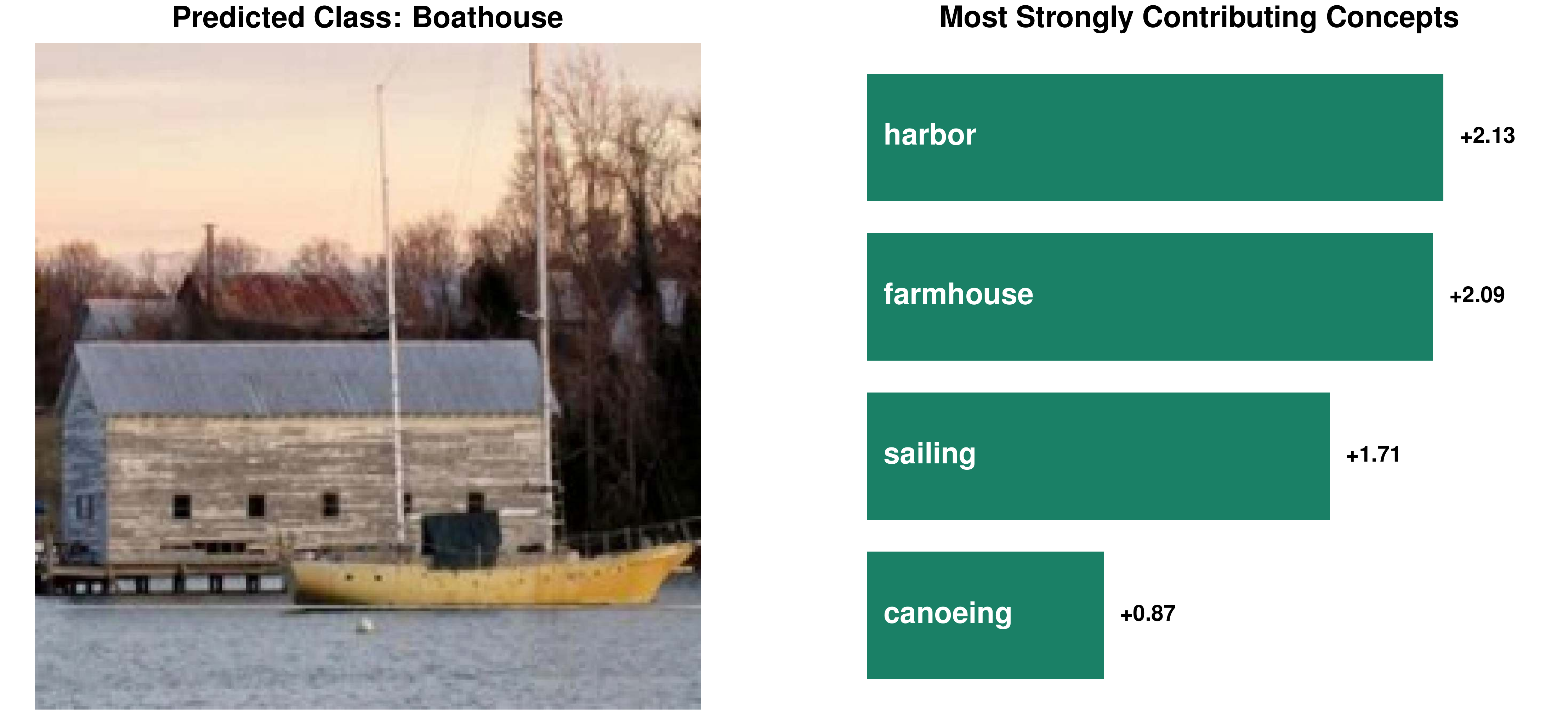}
   \caption{}
    \end{subfigure}\\
     \begin{subfigure}[c]{.45\textwidth}    \includegraphics[width=\linewidth]{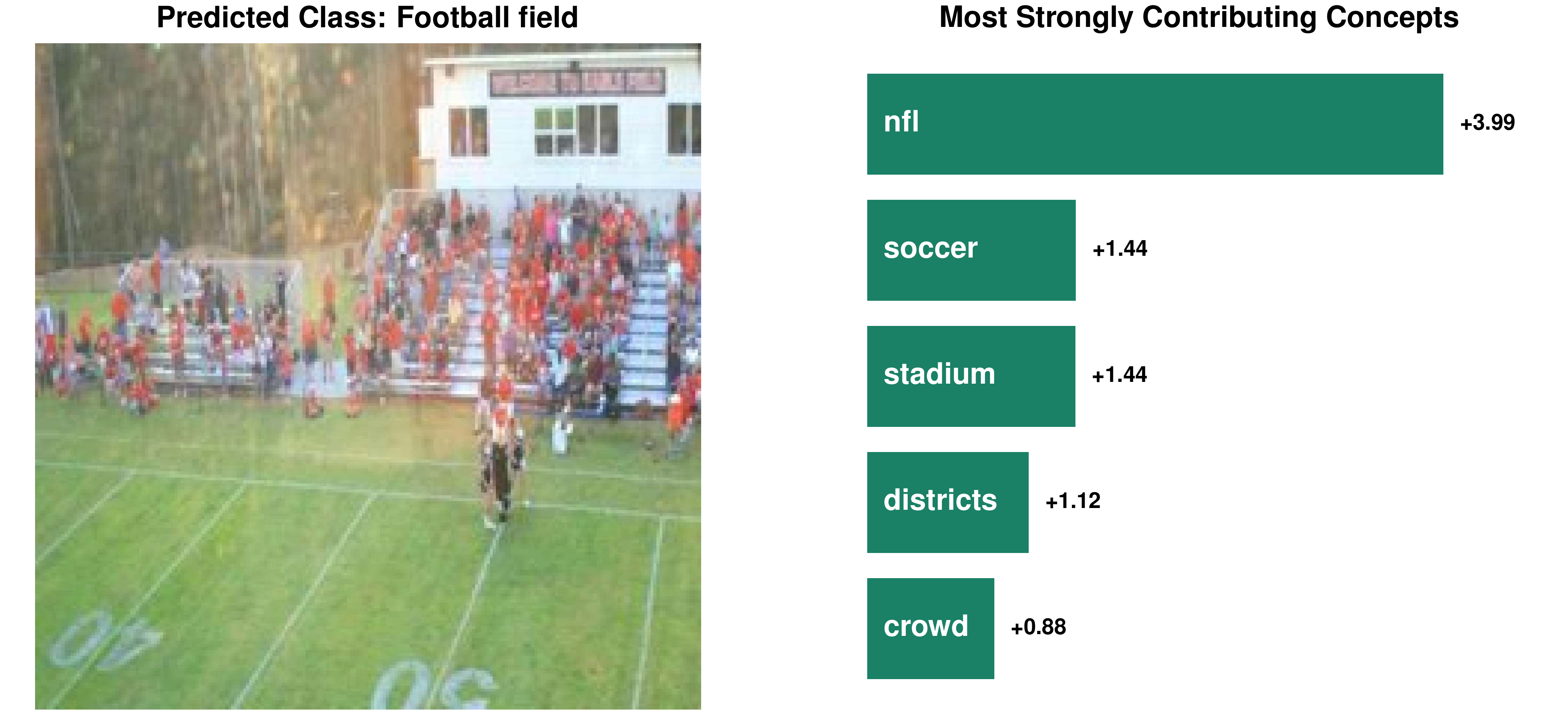}
     \caption{}
    \end{subfigure}\hfill
    \begin{subfigure}[c]{.45\textwidth}    \includegraphics[width=\linewidth]{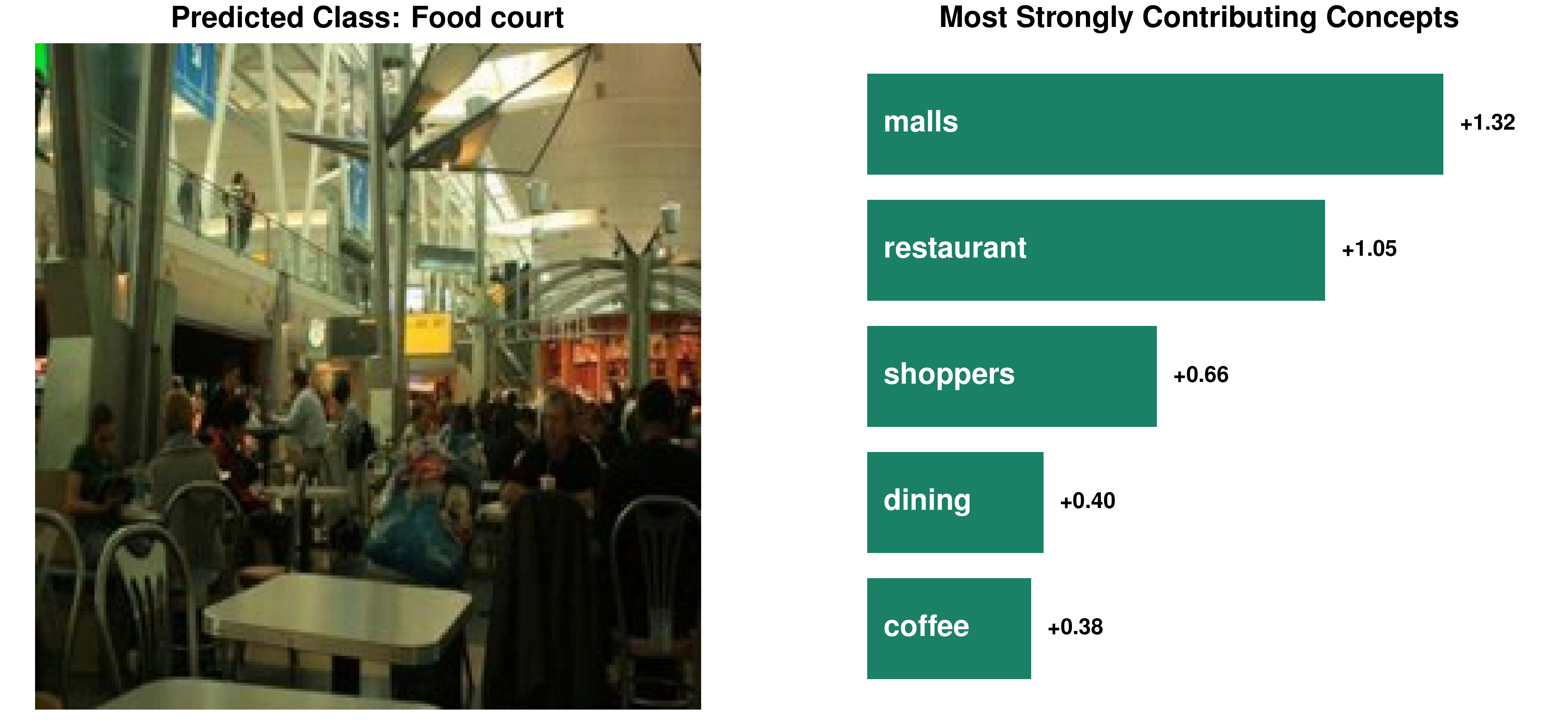}
    \caption{}
    \end{subfigure}\\
    \caption{\textbf{Examples of local explanations on \places.} We show examples of correctly classified images from the \places dataset along with the top concepts contributing to the decision using \clip\vitb (top) and \clip\vitl (bottom) vision backbones.
    We find that our \dncbm classifies based on a diverse set of concepts present in the image, including objects, similar features, higher level concepts, and things associated with the class (e.g. similar locations), thus aiding interpretability.
    \\\textbf{(a)} An example of class `Trench'. The top concepts in the decision include elements that are visually present in the image (\eg `mud', `forest'), elements that look visually similar to features in the image (\eg `rocks', `pathway'), and concepts associated with the class (\eg `infantry').
    \\\textbf{(b)} An example of class `Boathouse'. The top two concepts include elements similar to those visually in the image, \ie `harbor', which is depicted via the visible pier, and `farmhouse', which is visually similar to the depicted boathouse. The other top concepts include actions depicted in the image, \ie `sailing' and `canoeing'.
    \\\textbf{(c)} An example of class `Football field'. The top concepts are generally relevant to the class being predicted, and include features visibly present in the image, such as `stadium' and `crowd'. The top concept, `nfl', also accurately describes the sport. Interestingly, the concept `soccer' is also activated, which could be due to (potentially spurious) associations learnt by the model between the sports of American football and soccer, which is also commonly referred to as football. The concept `districts', however seems to have been activated incorrectly and does not appear to relate to any feature have any association with the image.
    \\\textbf{(d)} An example of class `Food court'. The top concepts include relevant locations (\ie `malls'), associated locations that are visually similar (\ie `restaurant'), actions (\ie `dining'), and visible elements (\ie `shoppers' and `coffee').
    }
    \label{fig:localexp_places_correct}
\end{figure}

\begin{figure}[t]
    \centering
    \begin{subfigure}[c]{.45\textwidth}    \includegraphics[width=\linewidth]{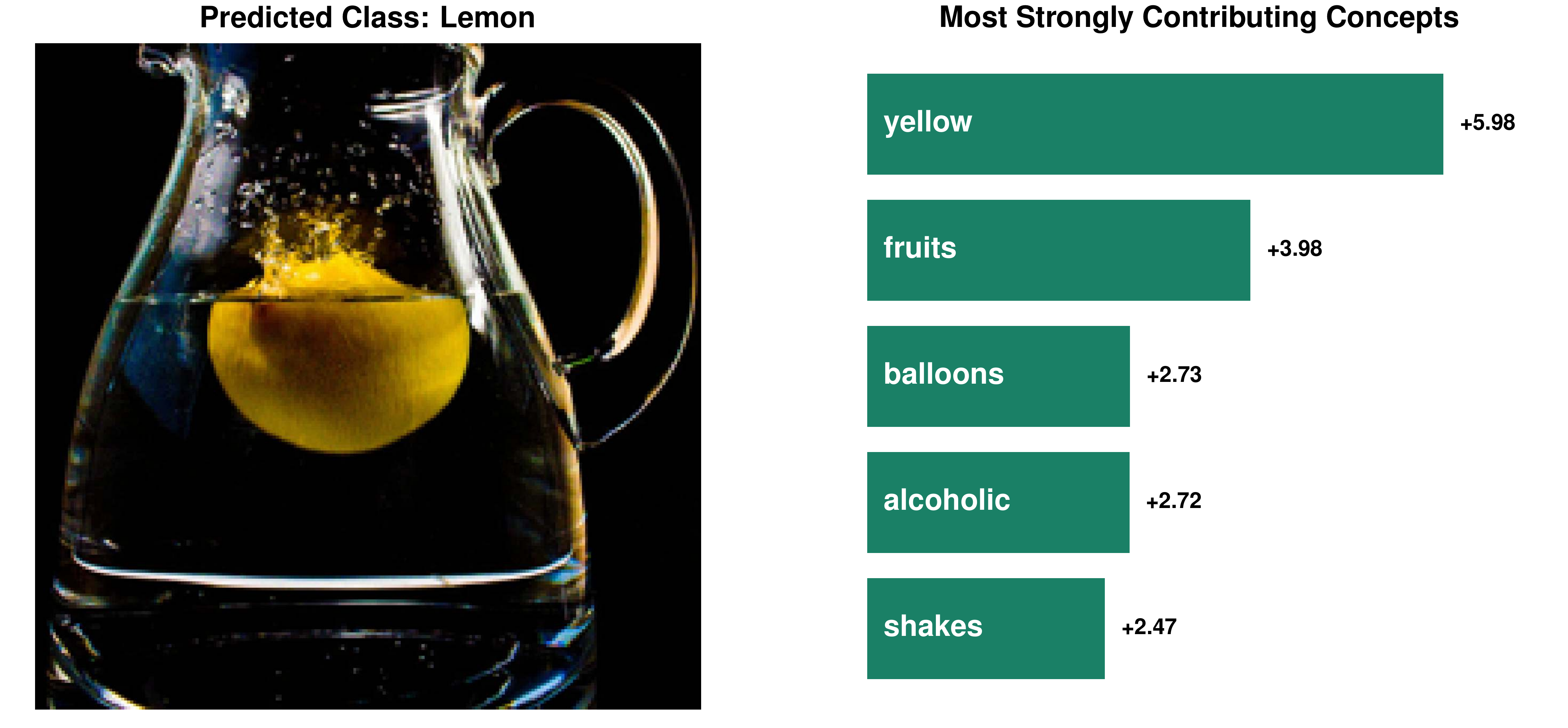}
    \caption{}
    \end{subfigure}\hfill
    \begin{subfigure}[c]{.45\textwidth}    \includegraphics[width=\linewidth]{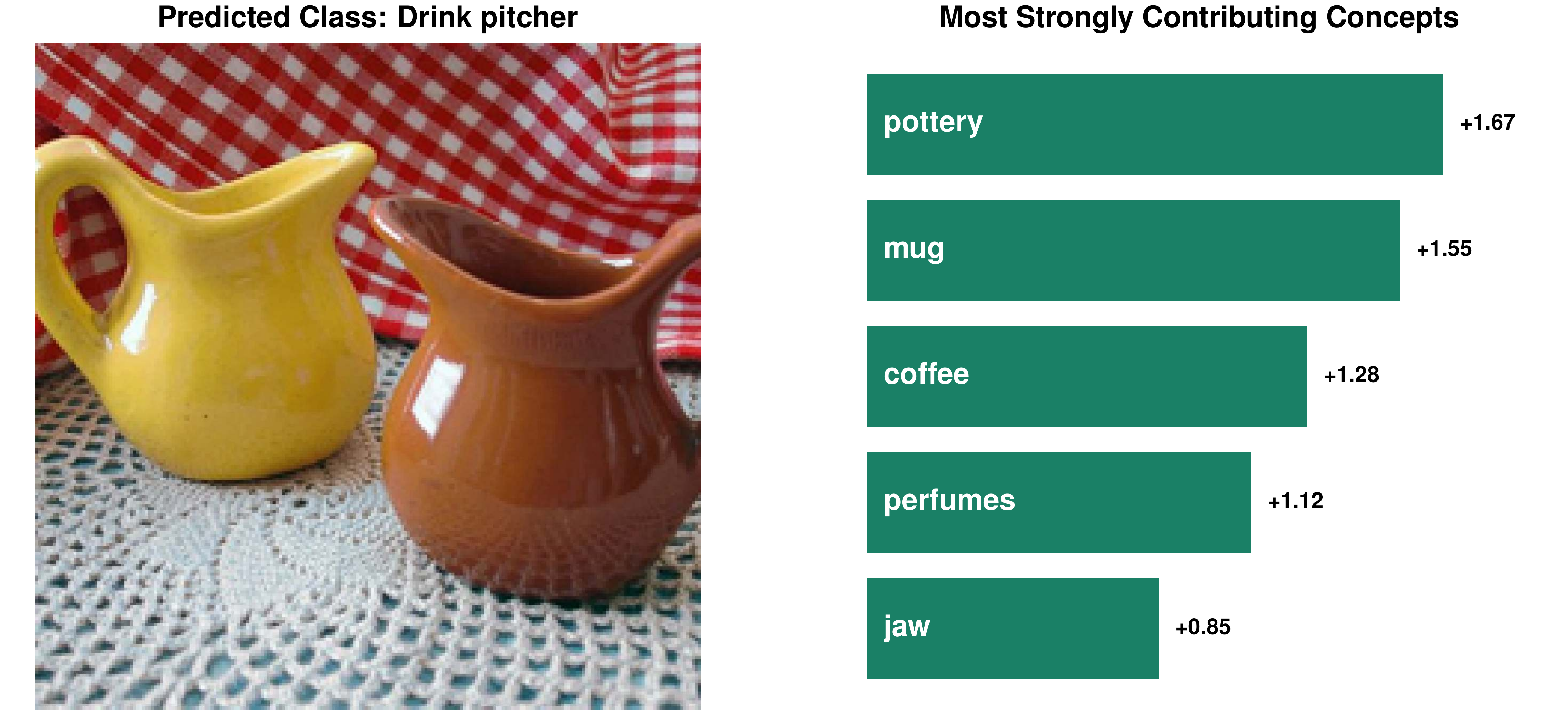}
    \caption{}
    \end{subfigure}\\
    \begin{subfigure}[c]{.45\textwidth}    \includegraphics[width=\linewidth]{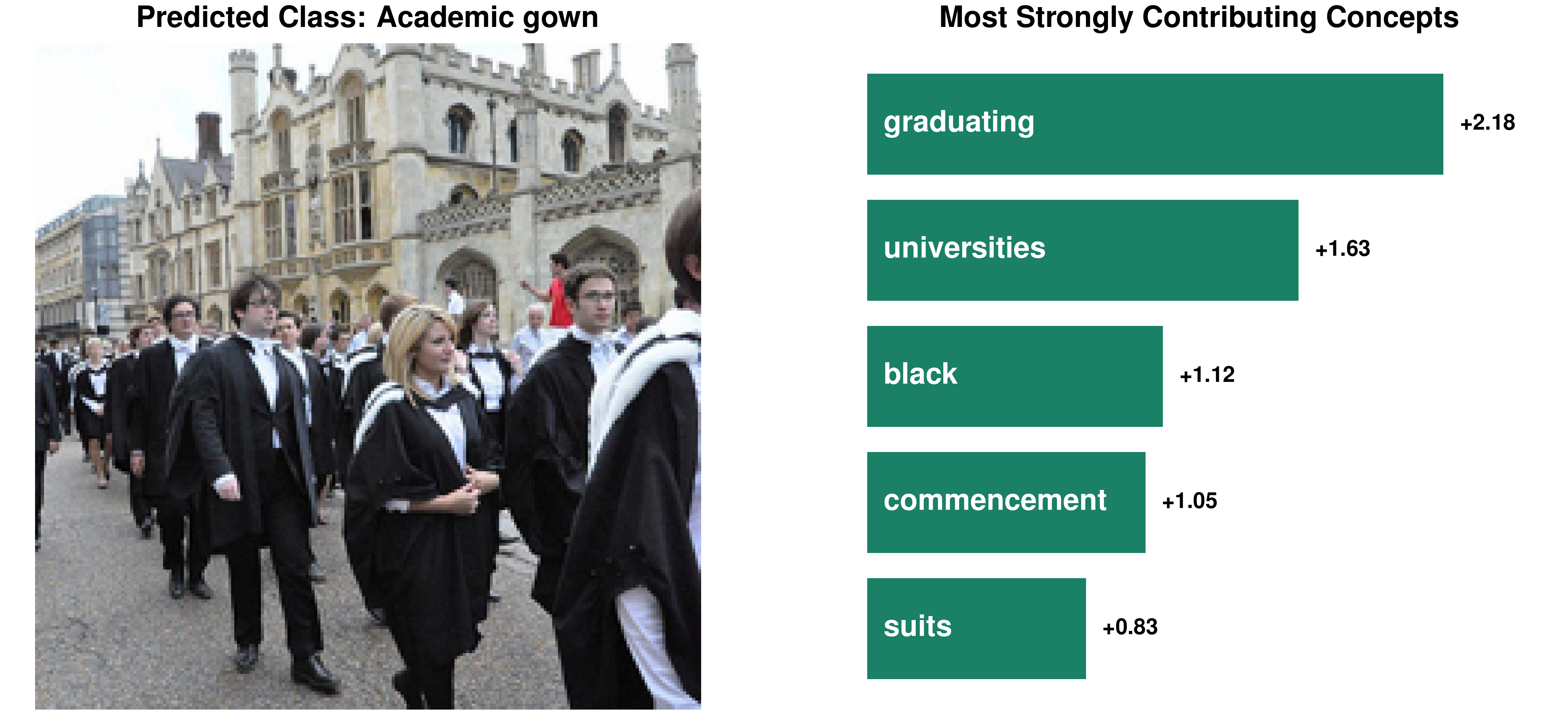}
    \caption{}
    \end{subfigure}\hfill
    \begin{subfigure}[c]{.45\textwidth}    \includegraphics[width=\linewidth]{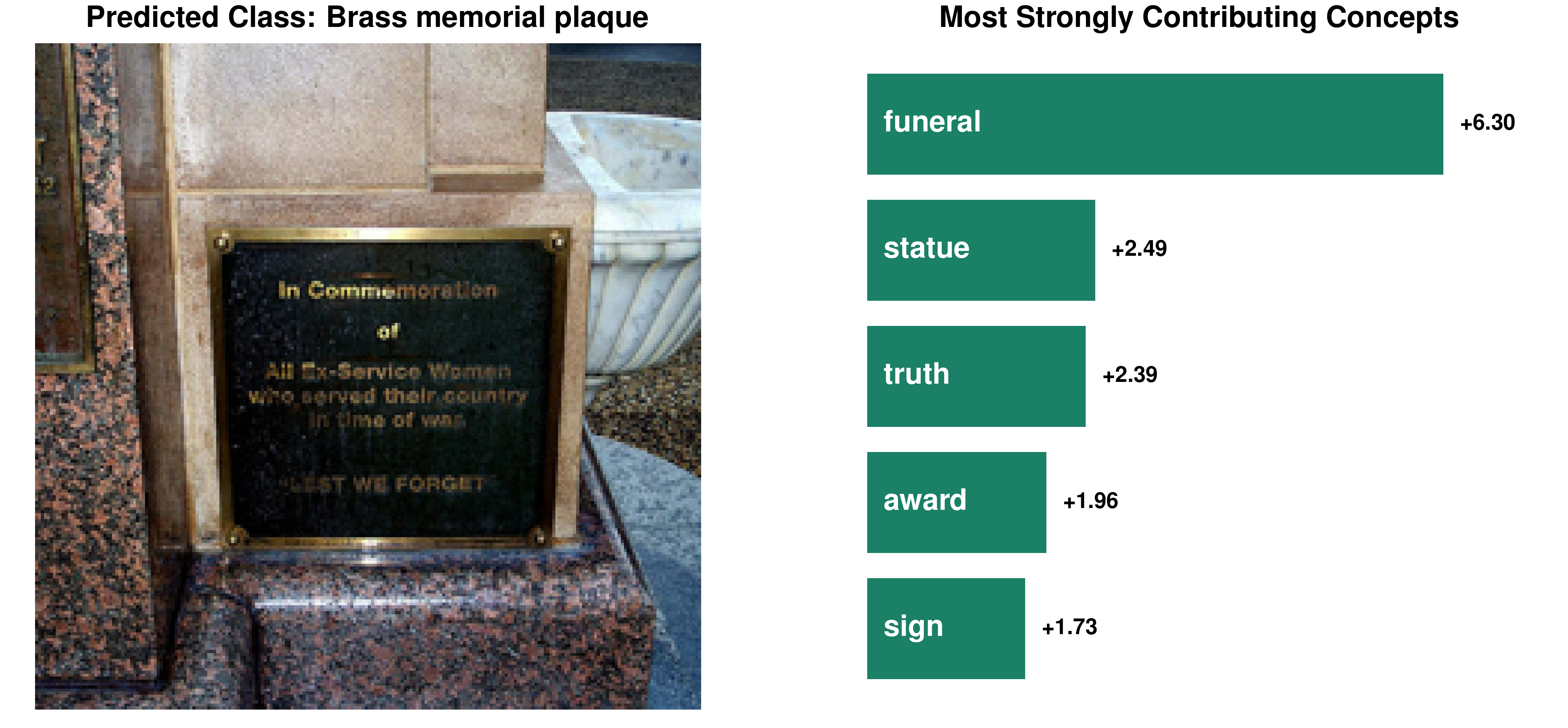}
    \caption{}
    \end{subfigure}\\ 
    \caption{\textbf{Examples of local explanations on \imagenet.} We show examples of correctly classified images from the \imagenet dataset along with the top concepts contributing to the decision using the three vision backbones.
    We find that our \dncbm classifies based on a diverse set of concepts present in the image, including objects, similar features, higher level concepts, and things associated with the class (e.g. similar locations), thus aiding interpretability.
    \\\textbf{(a)} An example of class `Water jug' misclassified as class `Lemon'. The image and the top concepts appear to clearly explain the misclassification: the lemon, being highly salient in the image, activates `yellow' (colour), `fruits' (category), and balloons (similar shape) as the top concepts. The other concepts (\ie `alcoholic', `shakes') might be associated to drinks typically found in such jugs, but are outweighed by the presence of the top three concepts. This shows that the explanations could be used to better understand misclassifications.
    \\\textbf{(b)} An example of class `Drink pitcher'. The top concepts include visually similar elements, such as `pottery' and `mug'. Interestingly, one of the top concepts is also `jaw', which is visually similar to the openings of the pitchers shown. Concepts involving contents that may be present in such pitchers are also activated, such as `coffee' and `perfumes'.
    \\\textbf{(c)} An example of class `Academic gown'. The top concepts are highly diverse, including the colour of the gowns in the image (\ie `black'), what the gowns look visually similar to (\ie `suits'), the location in the background, that is typical for such ceremonial gowns (\ie `universities'), and concepts related to the event where such gowns are worn (\ie `graduating' and `commencement').
    \\\textbf{(d)} An example of class `Brass memorial plaque'. The top concepts include what plaques represent (\ie `sign') and locations where plaques are typically found (\ie `statue', `award'). The highest activating concept, `funeral', could be associated with the semantic closeness specifically of memorial plaques to deceased individuals. However, the concept `truth' appears to have been detected incorrectly.
    }
    \label{fig:localexp_imagenet}
\end{figure}

%% file: supplement/figures_tex/global_explanations.tex
\begin{figure}[t!]
    \centering
    \begin{subfigure}[c]{.95\textwidth}
    \includegraphics[width=\linewidth]{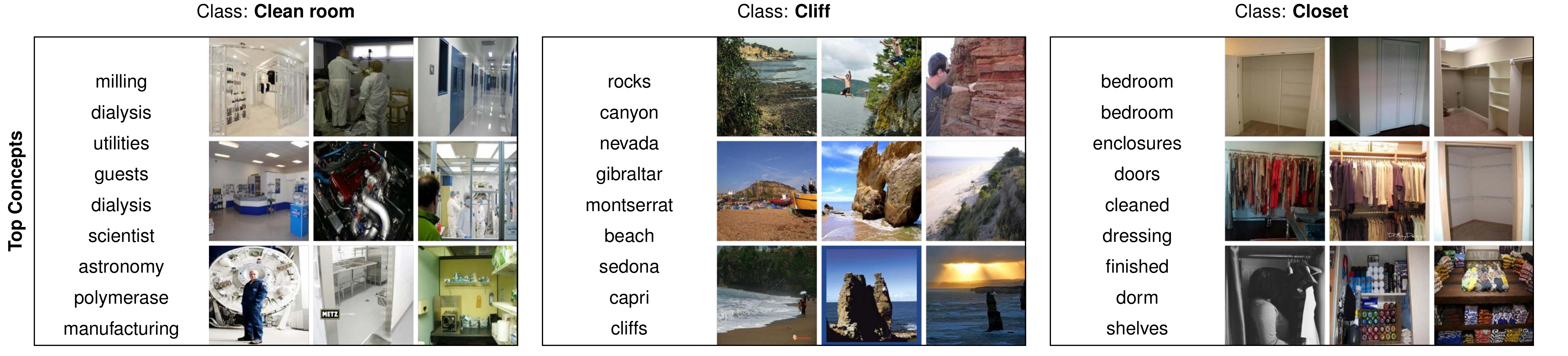}
    \caption{\clip\resnetf}
    \end{subfigure}
    \begin{subfigure}[c]{.95\textwidth}
    \includegraphics[width=\linewidth]{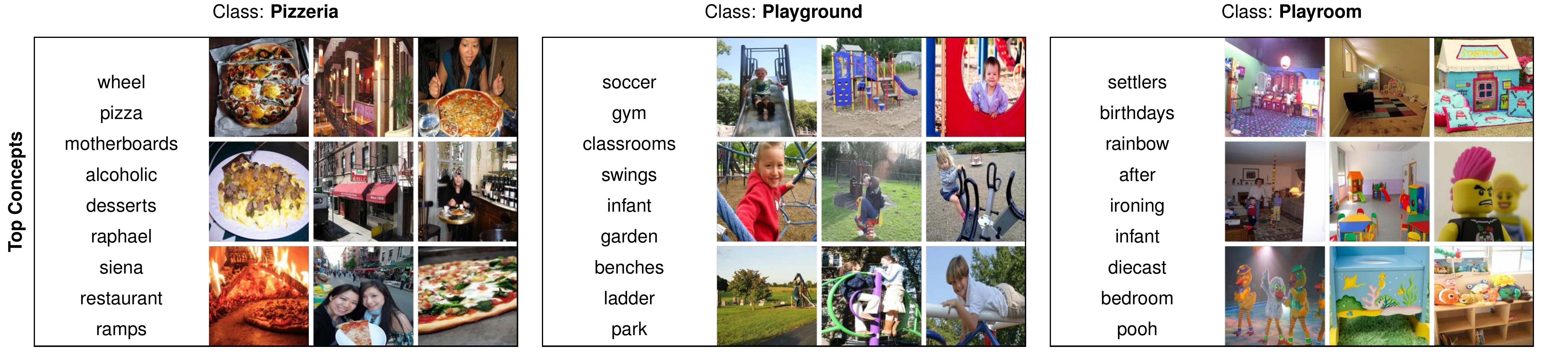}
    \caption{\clip\vitb}
    \end{subfigure}
    \begin{subfigure}[c]{.95\textwidth}
    \includegraphics[width=\linewidth]{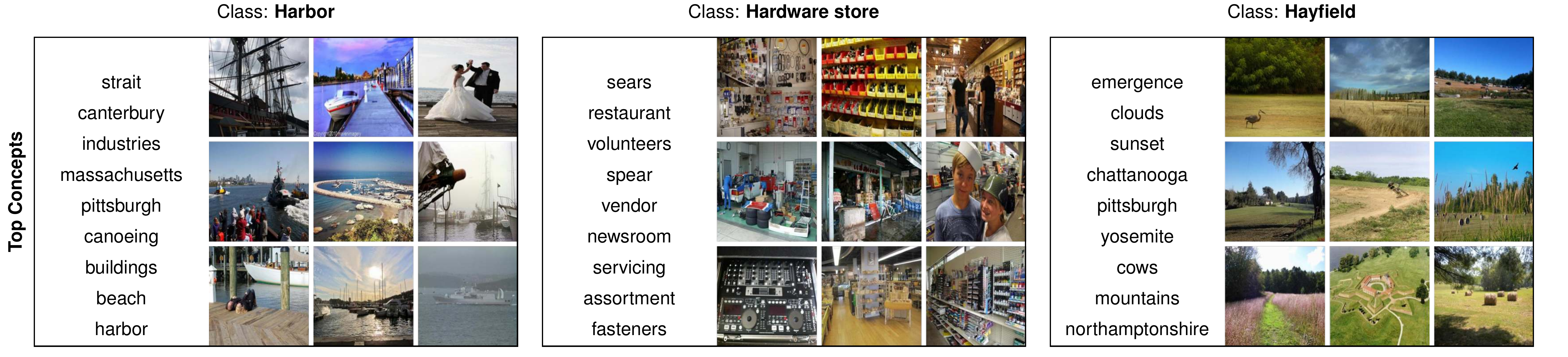}
    \caption{\clip\vitl}
    \end{subfigure}
    \caption{\textbf{Class-wise explanations using our \dncbm.} We show examples of classes from the \places dataset with the top contributing concepts. For each class (block), we show random examples of images belonging to that class and select concepts with the highest average contribution across all images from the class in the validation set.
    \\\textbf{(1)} We find that our approach learns classifiers that use concepts highly semantically related to each class. For e.g., the class `cliff' is associated with concepts such as rocks, canyon, beach and also with places like gibraltar, montserrat which could due to the association the \clip model has learnt from the training data. 
    \\\textbf{(2)} Note that the class name itself often appears in the list of top concepts (\eg concept `cliffs' for class `cliff', and concept `harbor' for class `harbor') as we use a generic vocabulary  which consists of the most used English words. As we assume for our CBM method that we do not have access to the class labels a priori, we do not filter out these words, so as we are working with the last layer representation, the model sometimes fires the concept corresponding to the class name to explain the same class.
    \\\textbf{(3)} The `Playroom' class is explained by concepts such as `rainbow' which is usually associated with colors, and the concept `pooh', 'birthdays' and  `infant' which are closely associated with a child's `Playroom'. The class `Playground' is well described by concepts such as `ladder', `benches', `swings', `infant', `soccer' which are usually present in a child's playground, as well as concepts like  `park', `garden' are similar looking places and the concept `gym' is activated as it contains gym equipments which are similar to the exercising equipments present in a playground or park.   
    }
    \label{fig:globalexp_supp}
\end{figure}